\documentclass[11pt]{article}

\usepackage{enumitem}
\usepackage{acl}

\usepackage{makecell} 
\usepackage{comment} 
\usepackage{dblfloatfix}
\usepackage{cuted}   
\usepackage{capt-of} 
\usepackage{tabularx}

\usepackage{times}
\usepackage{xcolor}
\usepackage{latexsym}
\usepackage{graphicx} 
\usepackage{amssymb} 
\usepackage[table]{xcolor}  
\usepackage{placeins}

\usepackage{booktabs}   
\usepackage{caption}    
\usepackage{array}      
\usepackage{svg}        
\usepackage[most]{tcolorbox}
\usepackage{subcaption}
\usepackage{multirow}
\usepackage[T1]{fontenc}
\usepackage{placeins}

\usepackage[utf8]{inputenc}
\usepackage{dblfloatfix} 

\usepackage{microtype}

\usepackage{inconsolata}

\usepackage{graphicx}
\usepackage{amsmath}   
\usepackage{caption}    
\usepackage{array}      

\usepackage{tikz}
\usetikzlibrary{positioning, shapes, fit}

\title{\textit{Point of Order}: Action-Aware LLM Persona Modeling for Data-Grounded Civic Deliberation}

\author{
Scott Merrill \quad Shashank Srivastava \\
University of North Carolina at Chapel Hill \\
\texttt{\{smerrill, ssrivastava\}@cs.unc.edu}
}

\begin{document}
\maketitle
\begin{abstract}

LLM-based simulations can enable controlled studies of civic deliberation, but current systems lack speaker-attributed data and methods for evaluating long-form institutional behavior. ASR transcripts typically use anonymous labels such as \texttt{Speaker\_1}, preventing models from learning stable participant behavior across meetings. We present a reproducible pipeline that converts public Zoom recordings into speaker-attributed transcripts enriched with persona profiles, topics, and pragmatic \textit{action tags} such as \texttt{[propose\_motion]}. Using this pipeline, we release three public datasets of government deliberation (Appellate Court hearings, School Board meetings, and Municipal Council sessions) and fine-tune LLM personas on this action-aware data. We evaluate simulations along four dimensions: persona fidelity, persona consistency, institutional fidelity, and behavioral coherence. Action-aware fine-tuning cuts perplexity by 67\%, doubles classifier-based persona fidelity, increases vote attempts by up to $3.6\times$, and improves deliberative responsiveness by up to $70\%$. Human evaluations show that simulated excerpts are often hard to distinguish from real deliberations, indicating a practical foundation for data-grounded civic simulation studies.
\end{abstract}

\section{Introduction}

\begin{figure}[h!]

    \centering
    \makebox[\columnwidth][c]{%
        \includegraphics[width=0.685\columnwidth]{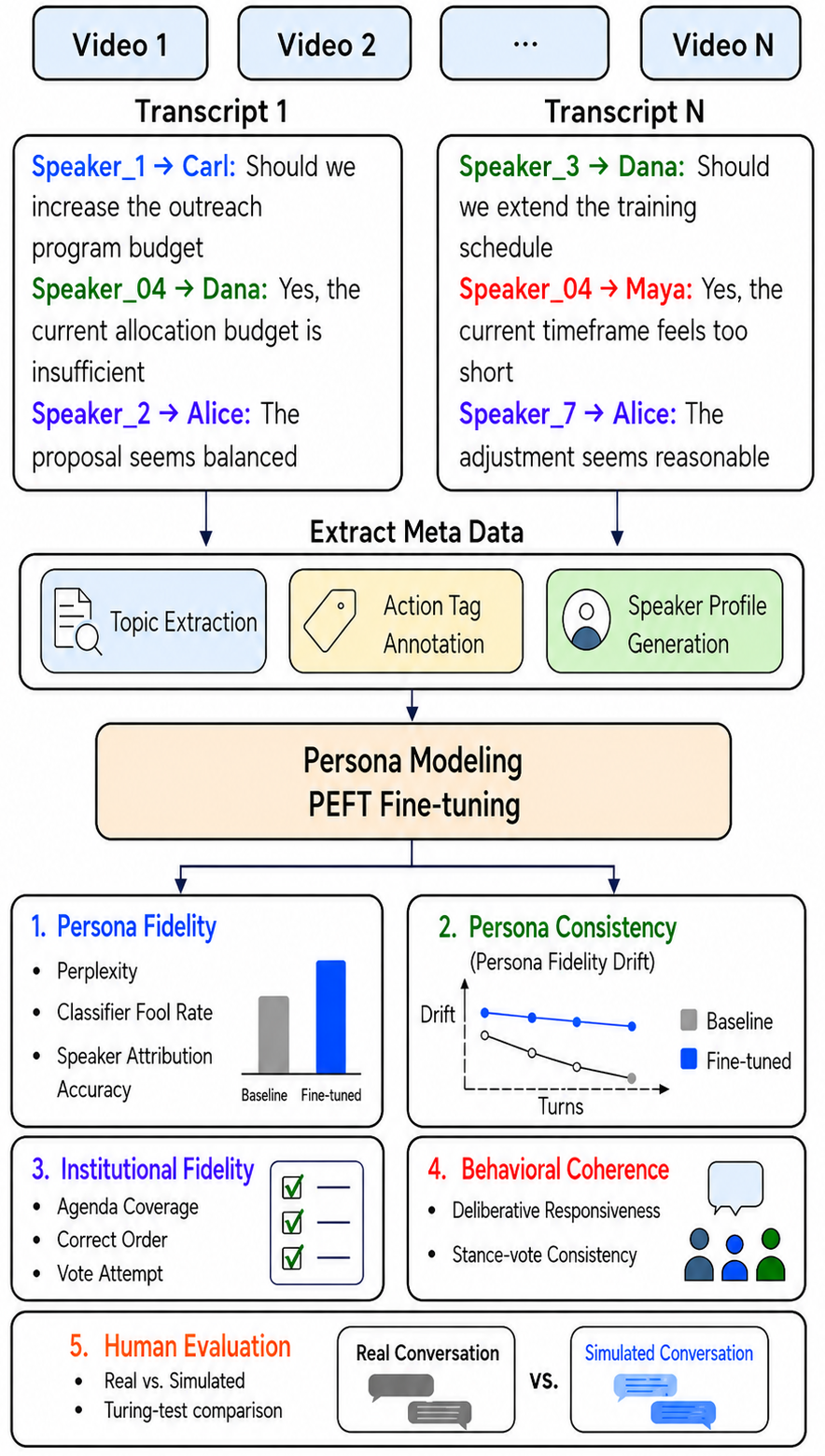}%
    }
    \caption{Public Zoom recordings are converted into speaker-attributed transcripts and linked across videos to build large-scale datasets. Structured topics, action tags, and speaker profiles condition PEFT fine-tuned LLM personas, which are evaluated across four dimensions and validated through human judgments.}
    \label{fig:main_figure}
\end{figure}

Deliberative forums such as courtrooms, city councils, and school boards shape high-stakes decisions in law, infrastructure, and education. These settings are not just generic multi-party conversations: they involve recurring participants, formal roles, procedural constraints, and decision points. LLM-based simulations could support controlled ``what-if'' analyses of such settings. For example, how deliberations change under different agendas, participation rules, or procedural structures. However, building such simulations requires models that can represent stable participants and institution-specific behavior over extended interactions.

A central bottleneck is data. Although millions of public meeting recordings are available online, ASR-generated transcripts typically use anonymous labels such as \texttt{Speaker\_1} and \texttt{Speaker\_2}, preventing models from learning consistent participant behavior across meetings. We address this with a lightweight multimodal speaker-linking pipeline (Figure~\ref{fig:main_figure}, top) that uses visual cues, audio, and textual context to assign stable speaker identities in standard public Zoom recordings, without requiring specialized metadata.

Speaker identity alone, however, is insufficient for realistic institutional simulation. Civic meetings move rapidly between formal procedure, spontaneous discussion, questions, motions, clarifications, and votes. To model this structure, we introduce \textit{action-aware persona modeling} (Figure~\ref{fig:main_figure}, middle): LLM personas are trained on speaker-attributed transcripts enriched with persona profiles, meeting topics, and pragmatic action tags such as \texttt{[propose\_motion]} and \texttt{[ask\_clarification]}. This supervision allows models to learn not only how individual speakers sound, but also what kinds of actions they take in institutional context.

Using this pipeline, we release three speaker-labeled datasets of government deliberation (Appellate Court hearings, School Board meetings, and Municipal Council sessions) covering 1.72M words. We also release the speaker-linking and annotation pipeline as a reusable artifact for constructing similar datasets from public meeting videos.

Finally, we introduce a multidimensional evaluation framework for long-form institutional simulation (Figure~\ref{fig:main_figure}, bottom), measuring \textit{persona fidelity}, \textit{persona consistency}, \textit{institutional fidelity}, and \textit{behavioral coherence}. Across three model families, action-aware fine-tuning reduces perplexity from $20.37$ to $6.64$ ($-67\%$), roughly doubles classifier-based persona fidelity, increases vote attempt rates by up to $3.6\times$, and improves deliberative responsiveness by up to $70\%$. Turing-style human evaluations further show that simulated excerpts are often difficult to distinguish from real transcripts. Together, these results suggest that action-aware supervision provides a practical foundation for data-grounded simulations of civic deliberation. Our contributions are:
\begin{itemize}[noitemsep, topsep=0pt, parsep=0pt, partopsep=0pt,leftmargin=10pt]
    \item A reusable multimodal diarization and speaker-linking pipeline enabling scalable, accurate identity assignment in public meeting videos.
    \item An action-aware persona modeling framework showing that structured metadata substantially enhances simulation fidelity.
    \item Three new datasets of speaker-labeled governmental deliberations spanning courts, school boards, and municipal councils covering $\sim$1.72M words in total.
    \item A multidimensional evaluation framework measuring persona fidelity, persona consistency, institutional fidelity, behavioral coherence, and human-perceived realism.
\end{itemize}

\begin{figure}[t!]
    \centering
    \includegraphics[width=1\columnwidth]{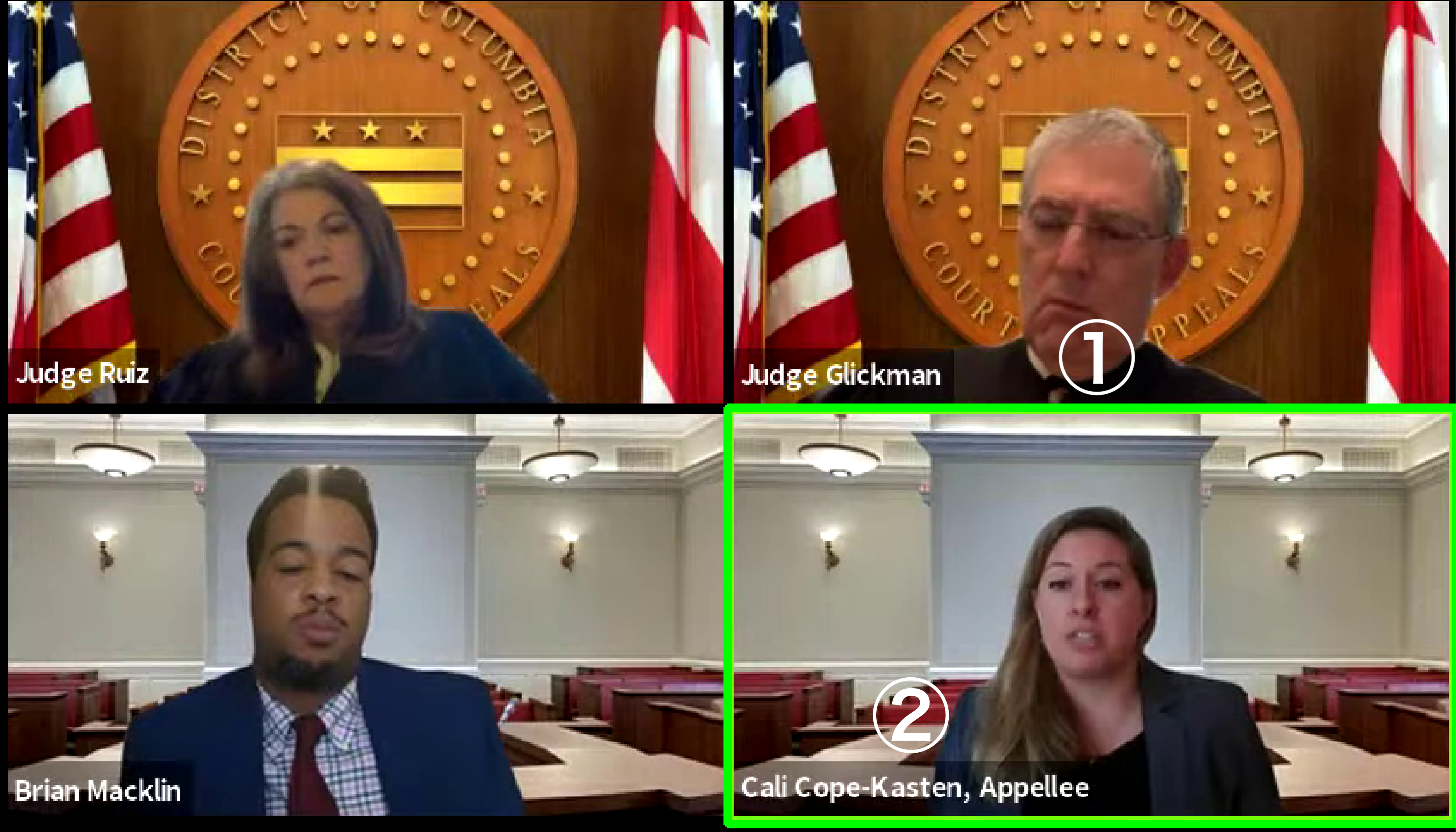}
    \caption{
    (1) Active speakers are identified using the highlighted speaker tile in Zoom. 
    (2) OCR extracts the name-region from the detected speaker tile. 
    }
    \label{fig:zoom_diarization}
\end{figure}

\section{Related Work}
\label{sec:related}
\paragraph{Speaker Diarization and Linking.}
Speaker diarization remains difficult in noisy, multi-party settings~\cite{landini2020vbx, bredin2019pyannote}; end-to-end systems improve robustness but still struggle with acoustic mismatch and varying speaker counts~\cite{fujita2019eend, horiguchi2021eend_eda, ryant2021diharddiarizationchallenge}. The harder task of \textbf{speaker linking} (maintaining identity across recordings) often relies on high-resolution video, lip-sync cues, or metadata~\cite{LeungSpeakerLinking, roth2020ava_activespeaker, xu2021ava_avd, liu22t_interspeech}, which are rarely available in real-world meetings. We instead use visual cues from  Zoom videos to achieve reliable linking from standard public recordings without specialized video or metadata.

\paragraph{Persona Modeling.}
Persona modeling typically uses prompting or fine-tuning: Prompting is sensitive to phrasing and prone to drift~\cite{zheng2024, wang2024}, while PEFT enables more persistent persona control~\cite{hu2021lora, peft, binz2025centaur}. We extend this to real-world deliberation, where profiles alone are insufficient; achieving fidelity requires \textbf{behaviorally grounded supervision}. Our action tags guide models toward realistic behavior rather than stylistic mimicry~\cite{mahajan-shaikh-2024-persona}.

\paragraph{Multi-Agent Deliberation.}
LLMs are increasingly used to simulate multi-agent debate~\cite{Ashkinaze_2025, du2023improvingfactualityreasoninglanguage, wu2023autogenenablingnextgenllm, tessler2026aimediationimprovedemocratic, karanam2024towards, zohny2025adept}, but most systems depend on prompted personas, yielding shallow mimicry and limited modeling of complex human deliberation~\cite{hu2024quantifyingpersonaeffectllm, flamino2024}. Evaluation frameworks emphasize short-term persuasion rather than sustained fidelity~\cite{liu-etal-2024-evaluating-large}. We address these limitations with PEFT-tuned models trained on real-world transcripts and grounded in metadata. 

\section{Methods}

Our pipeline converts public gallery-view Zoom recordings into speaker-attributed, metadata-enriched transcripts for persona modeling. These recordings contain cues that ASR transcripts discard: the active speaker is visually highlighted, participant names appear on video tiles, and the audio can be aligned with this visual timeline. We use these cues to recover stable speaker identities across recordings, then add three metadata layers: speaker-level persona profiles, meeting-level topics, and turn-level action tags.

\subsection{Data Collection}

Videos were downloaded from YouTube using \texttt{yt-dlp} \cite{yt-dlp} and processed at 1 frame per second, which is sufficient to capture Zoom highlight transitions while remaining efficient. Active speakers were identified by applying a color mask and contour extraction \cite{opencv_library} to locate the highlighted tile. The name region of the active tile was cropped and passed to OCR \cite{cui2025paddleocr30technicalreport}; for low-resolution recordings, we applied EDSR super-resolution \cite{lim2017enhanceddeepresidualnetworks} before OCR. This produced a per-second timeline of active speaker identities across video qualities from 240p to 1080p (Figure~\ref{fig:zoom_diarization}). Audio was transcribed with Whisper \cite{radford2022whisper}, and each segment was assigned to the active speaker at the corresponding timestamp. OCR-derived names were normalized and clustered with fuzzy matching to merge spelling variants and produce canonical speaker labels across recordings.

\subsection{Extracting Metadata}

We enrich each speaker-attributed transcript with persona profiles, meeting topics, and action tags.

\paragraph{Persona Profile Extraction}
\label{subsec:persona_extraction}
Persona profiles are derived from each speaker's 25 longest monologues, since sustained turns provide the clearest evidence of goals, policy positions, values, boundaries, and rhetorical style. A first \texttt{GPT-5} pass extracts these dimensions from each monologue; a second pass consolidates them into a compact speaker profile for downstream dialogue conditioning \cite{openai2025gpt5}. Prompts appear in Appendix~\ref{app:extraction}, Fig.~\ref{fig:persona_prompts}.

\paragraph{Topic Extraction}
\label{subsec:topic_extraction}
Meeting topics are extracted by segmenting each transcript into contiguous 1,024-token chunks, summarizing each chunk into 2--3 bullet points with \texttt{GPT-5}, and consolidating these summaries into a small set of meeting-level topic labels. This preserves the main agenda context without requiring the full transcript history. Prompts appear in Appendix~\ref{app:extraction}, Fig.~\ref{fig:topic_extraction_prompts}.

\paragraph{Action Tag Extraction}
Action tags capture the pragmatic function of each utterance, such as proposing a motion, asking for clarification, acknowledging a prior turn, calling a vote, giving logistical information, or requesting an action. Each dataset uses a compact, domain-specific tagset of 15--30 categories. Tags are action-focused rather than sentiment-focused, utterance-level rather than document-level, and allow multiple labels when a turn serves more than one function. \texttt{GPT-5} assigns tags using the dataset-specific taxonomy, providing scalable supervision for modeling what speakers do in institutional context, not only what they say.

\subsection{Dataset Format}

Diarized transcripts are serialized into a compact ChatML-style schema \cite{openai2023cookbook}. Each training example contains preceding speaker-labeled context and a target utterance from the speaker being modeled. Context is truncated at 1,024 tokens; when truncation occurs mid-turn, the speaker label is preserved and the visible text is prefixed with an ellipsis. In the tagged condition, action tags are prepended to each utterance and sorted alphabetically. Examples appear in Appendix~\ref{app:chatml_format}, Fig.~\ref{fig:dataset_format}.

\section{Datasets}
\label{sec:datasets}

Using our gallery-view diarization pipeline, we construct three speaker-labeled deliberation datasets spanning distinct institutional settings: DC Court of Appeals hearings, Albemarle County School Board meetings, and Waipā District Council sessions. These datasets differ in size, formality, and interaction structure, allowing us to test whether the pipeline works across several kinds of public deliberation. \autoref{tab:dataset_comparison} summarizes the corpus statistics; additional dataset details, action tagsets, and procedures for finding gallery-view recordings are provided in Appendices~\ref{app:finding_datasets}, \ref{app:datasets}, and \ref{app:action_tagging}.

\textbf{DC Court of Appeals.}
This dataset contains hearings from the highest court in the District of Columbia, each involving three judges and one appellee \cite{dccourts2025}. It is the smallest collection, but contains dense, structured exchanges: judges question appellees, clarify legal arguments, and move through formal procedures.

\textbf{Albemarle County School Board.}
This dataset contains seven-member school board meetings overseeing Albemarle County Public Schools, along with contributions from the superintendent, staff, and community participants \cite{acps_schoolboard2025}. These meetings mix policy deliberation, administrative updates, staff reports, procedural motions, and public comment, and have the highest average number of participants per transcript.

\textbf{Waipā District Council.}
This dataset comprises transcripts from multiple Waipā District Council committees, including discussions among councillors, the Mayor, committee chairs, staff, and community representatives \cite{waipadc2025}. It is the largest corpus, capturing semi-formal governance interactions across committees. In this paper, we focus on the Audit and Risk Committee, while the broader collection supports future work on other council settings.

\begin{table}[t]
\centering
\resizebox{\linewidth}{!}{%
\begin{tabular}{lccc}
\hline
\textbf{Statistic} & \textbf{DC Appeals} & \textbf{Albemarle} & \textbf{Waipā} \\
\hline
Number of transcripts & 10 & 32 & 80 \\
Average people in conversation & 8.6 & 21.38 & 12.125 \\
Total words across all transcripts & 193,712 & 594,253 & 933,272 \\
Average words per transcript file & 19,371 & 18,570 & 11,666 \\
\hline
\end{tabular}%
}
\caption{Meta-statistics of the three speaker-labeled deliberation datasets.}
\label{tab:dataset_comparison}
\end{table}

\subsection{Dataset Validation}

We validate dataset quality with three MTurk studies, each using three independent workers per item and standard attention checks. In \textbf{Speaker Identification}, workers match an utterance to the correct speaker from a predefined list. In \textbf{Speaker Consistency}, workers judge whether two utterances from different videos come from the same speaker. In \textbf{Transcription Quality}, workers rate transcript accuracy and readability on a 5-point Likert scale. Full task interfaces are shown in Appendix~\ref{app:dataset_eval}.

The resulting labels and transcripts are reliable. Workers correctly identify speakers in 98.5\% of utterances and verify cross-video speaker links with 95.3\% accuracy (Figs.~\ref{fig:diarization_speaker_identification}--\ref{fig:diarization_speaker_consistency}). Transcript quality is also rated positively: 55.9\% of items are marked ``Strongly agree'' and 29.8\% ``Agree'' for accuracy/readability, with few negative ratings (Fig.~\ref{fig:diarization_transcription_quality}).

We also report inter-annotator agreement. Speaker identification achieves substantial agreement (Fleiss' $\kappa=0.765$). Speaker consistency has lower chance-corrected agreement ($\kappa=0.263$), largely because 80.3\% of responses fall into the same ``Yes'' label; direct agreement is much higher, with 83.8\% pairwise exact agreement and 75.0\% unanimous agreement. For transcription quality, annotators agree within one Likert step 85.3\% of the time on the 5-point scale and 92.3\% on a collapsed 3-point scale.

Finally, we validate the GPT-assigned action tags in a separate MTurk study (Appendix~\ref{app:data_val}). Human annotators match the GPT tag in 68.0\% of individual judgments and 81.5\% after majority-vote aggregation. Because pragmatic action labels can overlap and require interpretation, we view this as evidence that the tags are noisy but broadly aligned with human judgments, making them suitable as scalable supervision for action-aware modeling.

\section{Experiments}

\subsection{PEFT Fine-Tuning}
To study the impact of fine-tuning across architectures, three state-of-the-art transformer-based LLMs were selected: \texttt{LLaMA-3.1-70B} \cite{grattafixori2024llama3herdmodels}, \texttt{Qwen-2.5-72B} \cite{yang2024qwen2_5}, and \texttt{GPT-OSS-120B} \cite{openai2025gptoss120bgptoss20bmodel}.  \texttt{LLaMA-3.1-70B} and  \texttt{Qwen-2.5-72B} are dense decoder-only transformers differing in pretraining data and multilingual coverage, while \texttt{GPT-OSS-120B} is a Mixture-of-Experts (MoE) model optimized for reasoning-intensive tasks.

Two fine-tuning scenarios were evaluated: using transcripts with action tags and without tags (see Appendix~\ref{app:chatml_format}, Fig. \ref{fig:dataset_format}). Fine-tuning employed QLoRA \cite{qlora}, which applies quantization-aware low-rank updates while retaining full-precision gradients. Learning rate and LoRA $\alpha$ were optimized via grid search. Each model was trained for five epochs, with intermediate checkpoints saved for evaluation, and the best checkpoint selected based on perplexity on a held-out test set.  Performance was compared to baseline full-precision pretrained models using only structured system prompts (Section~\ref{sec:system_prompt}) for persona modeling (no task-specific fine-tuning).

\subsection{System Prompts}
\label{sec:system_prompt}
In addition to fine-tuning, structured \textit{system prompts} were used to guide model behavior at inference time (see Appendix~\ref{app:system_prompts}, Fig.~\ref{fig:persona-prompt}). These prompts integrate five layers of information to support role-consistent dialogue:

\begin{enumerate}[noitemsep, topsep=0pt, parsep=0pt, partopsep=0pt]
\item \textbf{Persona Description:} A concise summary of the agent’s defining traits, policy positions, and communication style (Section~\ref{subsec:persona_extraction}).
\item \textbf{In-Context Examples:} Representative utterances from training transcripts illustrating the speaker’s lexical choices, rhetorical style, and argumentation patterns.
\item \textbf{Micro-Profile:} Quantitative behavioral summaries derived from training transcripts. e.g., question rate, directive rate, and average sentiment—polarity.
\item \textbf{Conversation Context:} A list of meeting topics (Section~\ref{subsec:topic_extraction}) and participants. 
\item \textbf{Instruction Prompt:} Explicit guidance specifying the agent’s role.
\end{enumerate}


\subsection{Simulations}
To evaluate long-form institutional simulation, we deploy trained models as conversational agents in multi-party meeting simulations. Each dataset begins with agents receiving a seed prompt, agenda topics (see Appendix~\ref{app:sim_topics}), participant lists, and brief instructions (see Appendix~\ref{app:system_prompts}, Fig.~\ref{fig:time-aware-simulation}). An artificial clock advances one minute per utterance plus one minute per 100 words, with simulations running until the meeting duration is exceeded. Responses are generated in round-robin order, and the full conversation history is truncated to 1{,}024 tokens, consistent with the training setup.

We tested two simulation variants, shown in \autoref{fig:time-aware-simulation}. In the \textit{time-unaware} baseline, agents received only the meeting agenda and prior discussion context. In the \textit{time-aware} variant, agenda items were timestamped and agents were instructed to prepend each utterance with the simulated time. This comparison tests whether explicit temporal cues help agents maintain topic coherence over long, multi-party simulations.

\section{Evaluation Metrics}
\label{sec:evaluation_metrics}

We evaluate long-form simulations along four dimensions. \textit{Persona fidelity} asks if generated turns resemble the target speaker. \textit{Persona consistency} asks if this resemblance persists as the simulation unfolds. \textit{Institutional fidelity} asks if the meeting follows the expected agenda structure and procedural milestones. \textit{Behavioral coherence} asks if speakers behave in ways that are internally consistent.

\subsection{Persona Fidelity}

Persona fidelity measures how closely an agent's generated speech matches the target persona. We use three complementary metrics: perplexity, classifier fool rate, and speaker attribution accuracy.

\vspace{0.2cm}
\noindent \textbf{Perplexity (PPL)}
PPL measures how well a model predicts held-out utterances from the target speaker. Lower PPL indicates a better fit to the speaker's linguistic patterns.

\vspace{0.2cm}
\noindent \textbf{Classifier Fool Rate (CFR)}
CFR measures whether generated utterances are plausible for the target speaker. For each speaker $s \in S$, we train a one-vs-all classifier $C_s$ to distinguish utterances by $s$ from utterances by all other speakers. If agent $s$ generates $N_s$ utterances, CFR is the fraction classified by $C_s$ as real utterances from $s$:
\[
CFR(s) = \frac{1}{N_s}\sum_{i=1}^{N_s}\mathbb{I}[C_s(x_i^{(s)})=1].
\]
A high CFR means the generated language is convincing as speech from the target persona.

\paragraph{Speaker Attribution Accuracy (SAA)}
SAA measures whether generated utterances are not only plausible, but distinctive. A multi-class classifier $A$ predicts the speaker identity for each generated utterance $x_i^{(s)}$. SAA is the fraction of generated utterances attributed to the intended speaker:
\[
SAA = \frac{1}{\sum_{s\in S}N_s}\sum_{s\in S}\sum_{i=1}^{N_s}\mathbb{I}[A(x_i^{(s)})=s].
\]
CFR and SAA capture different aspects of persona fidelity. CFR asks whether an utterance could plausibly come from the target speaker; SAA asks whether it is distinguishable from utterances by other speakers in the same setting. Because CFR and SAA depend on reliable speaker classifiers, we exclude speakers whose held-out classifier performance falls below $F_1 < 0.50$ in the main calculation, and report all-speaker results in App \ref{app:all_speaker}.

\subsection{Persona Consistency}

Persona consistency measures whether agents maintain speaker-specific behavior over long simulations rather than drifting toward generic dialogue. We compute CFR and SAA separately on the first and final 10\% of each simulation and report the early-to-late change. Larger declines indicate weaker long-form persona consistency.

\begin{table*}[h!]
\centering
\scriptsize
\setlength{\tabcolsep}{4pt}
\captionsetup{width=0.97\textwidth}

\resizebox{0.97\textwidth}{!}{%
\begin{tabular}{lcccccccccc}
\toprule
\multirow{2}{*}{Model / Setting} & \multicolumn{3}{c}{Persona Fidelity} & \multicolumn{2}{c}{Persona Consistency} & \multicolumn{3}{c}{Institutional Fidelity} & \multicolumn{2}{c}{Behavioral Coherence} \\
\cmidrule(lr){2-4} \cmidrule(lr){5-6} \cmidrule(lr){7-9} \cmidrule(lr){10-11}
 & PPL $\downarrow$ & CFR $\uparrow$ & SAA $\uparrow$ & $\Delta$CFR $(|\cdot|\downarrow)$ & $\Delta$SAA $(|\cdot|\downarrow)$ & Coverage $\uparrow$ & Order $\uparrow$ & Vote Attempt $\uparrow$ & Stance/Vote $\uparrow$ & Responsiveness $\uparrow$ \\
\midrule
\multicolumn{11}{l}{\textbf{GPT}}\\
\hspace{1em}Fine-Tuned w/ tags & \textbf{8.23 ± 0.38} & \textbf{0.53 ± 0.09} & 0.20 ± 0.08 & 0.14 ± 0.00 & \textbf{0.03 ± 0.06} & 0.88 ± 0.05 & 0.83 ± 0.09 & 0.47 ± 0.20 & 0.701 ± 0.196 & 0.632 ± 0.033 \\
\hspace{1em}Fine-Tuned w/o tags & 13.52 ± 0.74 & 0.51 ± 0.06 & \textbf{0.23 ± 0.09} & 0.07 ± 0.04 & -0.05 ± 0.02 & \textbf{0.94 ± 0.03} & \textbf{0.89 ± 0.06} & \textbf{0.55 ± 0.03} & 0.856 ± 0.077 & \textbf{0.679 ± 0.055} \\
\hspace{1em}Baseline w/ tags & 18.34 ± 0.85 & 0.26 ± 0.11 & 0.11 ± 0.05 & \textbf{-0.02 ± 0.10} & -0.14 ± 0.16 & 0.49 ± 0.05 & \textbf{0.89 ± 0.11} & 0.13 ± 0.08 & \textbf{0.889 ± 0.059} & 0.571 ± 0.143 \\
\hspace{1em}Baseline w/o tags & 26.40 ± 1.76 & 0.26 ± 0.09 & 0.16 ± 0.04 & -0.06 ± 0.06 & -0.18 ± 0.10 & 0.63 ± 0.11 & 0.64 ± 0.03 & 0.17 ± 0.06 & 0.812 ± 0.157 & 0.523 ± 0.137 \\
\midrule
\multicolumn{11}{l}{\textbf{LLaMA}}\\
\hspace{1em}Fine-Tuned w/ tags & \textbf{5.93 ± 0.39} & \textbf{0.59 ± 0.05} & \textbf{0.30 ± 0.12} & 0.04 ± 0.08 & -0.03 ± 0.02 & 0.63 ± 0.13 & 0.91 ± 0.06 & \textbf{0.30 ± 0.09} & 0.740 ± 0.164 & 0.664 ± 0.016 \\
\hspace{1em}Fine-Tuned w/o tags & 8.52 ± 0.38 & 0.58 ± 0.06 & 0.28 ± 0.13 & \textbf{0.02 ± 0.08} & -0.09 ± 0.02 & 0.47 ± 0.25 & 0.85 ± 0.06 & 0.30 ± 0.15 & 0.627 ± 0.131 & 0.658 ± 0.050 \\
\hspace{1em}Baseline w/ tags & 10.36 ± 0.13 & 0.38 ± 0.06 & 0.18 ± 0.08 & 0.04 ± 0.05 & \textbf{0.01 ± 0.05} & \textbf{0.69 ± 0.26} & \textbf{0.94 ± 0.06} & 0.10 ± 0.05 & \textbf{0.925 ± 0.043} & \textbf{0.831 ± 0.025} \\
\hspace{1em}Baseline w/o tags & 15.06 ± 0.59 & 0.38 ± 0.08 & 0.20 ± 0.07 & 0.03 ± 0.01 & -0.04 ± 0.05 & 0.47 ± 0.21 & 0.88 ± 0.06 & 0.09 ± 0.09 & 0.653 ± 0.175 & 0.825 ± 0.032 \\
\midrule
\multicolumn{11}{l}{\textbf{Qwen}}\\
\hspace{1em}Fine-Tuned w/ tags & \textbf{5.69 ± 0.31} & \textbf{0.61 ± 0.04} & \textbf{0.33 ± 0.14} & 0.09 ± 0.03 & \textbf{0.08 ± 0.06} & 0.75 ± 0.14 & 0.89 ± 0.07 & 0.30 ± 0.15 & 0.589 ± 0.099 & \textbf{0.804 ± 0.059} \\
\hspace{1em}Fine-Tuned w/o tags & 8.51 ± 0.37 & 0.56 ± 0.06 & 0.31 ± 0.13 & 0.05 ± 0.05 & 0.10 ± 0.07 & 0.73 ± 0.15 & \textbf{0.93 ± 0.05} & \textbf{0.32 ± 0.12} & 0.578 ± 0.011 & 0.786 ± 0.050 \\
\hspace{1em}Baseline w/ tags & 14.54 ± 1.65 & 0.36 ± 0.09 & 0.15 ± 0.06 & 0.07 ± 0.03 & -0.12 ± 0.09 & 0.80 ± 0.11 & 0.88 ± 0.04 & 0.13 ± 0.10 & \textbf{0.667 ± 0.092} & 0.483 ± 0.180 \\
\hspace{1em}Baseline w/o tags & 19.27 ± 2.63 & 0.33 ± 0.10 & 0.15 ± 0.06 & \textbf{0.03 ± 0.01} & -0.10 ± 0.13 & \textbf{0.81 ± 0.16} & 0.92 ± 0.08 & 0.16 ± 0.03 & 0.312 ± 0.084 & 0.465 ± 0.167 \\
\bottomrule
\end{tabular}
}

\caption{Simulation quality across persona fidelity, persona consistency, institutional fidelity, and behavioral coherence. Values are mean $\pm$ SE across datasets. Lower PPL and smaller absolute consistency drift are better; higher is better for all other metrics. Per-model and per-dataset breakdowns for persona fidelity and consistency are in Appendix~\ref{app:model_eval}.}
\label{tab:simulation_realism_main_results}
\end{table*}

\subsection{Institutional Fidelity}

Institutional fidelity measures whether simulations preserve the meeting's organizational structure and procedural flow. We evaluate three properties: \textbf{agenda item coverage}, if target agenda items are meaningfully discussed; \textbf{agenda item ordering}, if they are addressed in the expected sequence; and \textbf{vote attempt rate}, if the simulation attempts a vote or decision when appropriate. These metrics test whether simulations progress through the right topics and procedural milestones, rather than merely producing locally plausible dialogue. We compute them with GPT-5 and validate the judgments against MTurk annotations in Appendix~\ref{app:metric_validation}.

\subsection{Behavioral Coherence}

Behavioral coherence measures whether simulated speakers behave consistently across discussion, response, and decision-making. We evaluate \textbf{stance-vote consistency}, if a speaker's deliberative comments align with the vote they later cast, and \textbf{deliberative responsiveness}, if speakers respond appropriately to questions or requests directed toward them. These are computed using GPT-5 and validated against MTurk annotations in Appendix~\ref{app:metric_validation}.

\subsection{Human Evaluation}

Finally, we measure perceived realism with a Turing-test-style human evaluation. Crowdworkers see two short four-turn excerpts side by side, one from a real transcript and one from a simulation, and are asked to identify the human-generated conversation (Appendix~\ref{app:simulation_result}, Fig.~\ref{fig:mturk_simulation_main}). Lower identification accuracy means the simulated dialogue is harder to distinguish from real meeting dialogue.

\section{Results}
\label{sec:results}
\autoref{tab:simulation_realism_main_results} summarizes performance across our full evaluation suite with more detailed results shown in Appendix~\ref{app:model_eval}.

\paragraph{Persona fidelity.}
Fine-tuning improves persona fidelity across all model families and metrics. For \texttt{GPT}, \texttt{LLaMA}, and \texttt{Qwen}, respectively, PPL decreases from $26.40$, $15.06$, and $19.27$ to $13.52$, $8.52$, and $8.51$; CFR increases from $0.26$, $0.38$, and $0.33$ to $0.51$, $0.58$, and $0.56$; and SAA increases from $0.16$, $0.20$, and $0.15$ to $0.23$, $0.28$, and $0.31$. Action tags further improve both baseline and fine-tuned models, consistently reducing PPL and increasing CFR even when the underlying model was not trained with tags. This suggests that action tags expressed in natural-language help models produce more contextually appropriate speech.

\paragraph{Persona consistency.}
Fine-tuning generally reduces long-horizon persona drift, especially in SAA. For \texttt{GPT}, \texttt{LLaMA}, and \texttt{Qwen}, $\Delta$SAA changes from $-0.18$, $-0.04$, and $-0.10$ to $-0.05$, $-0.09$, and $+0.10$ without action tags, and from $-0.14$, $+0.01$, and $-0.12$ to $+0.03$, $-0.03$, and $+0.08$ with action tags. Thus, fine-tuning substantially improves SAA stability for \texttt{GPT} and \texttt{Qwen}, while the \texttt{LLaMA} baseline already has stable speaker attribution scores. CFR drift is smaller and more mixed. Surprisingly, fine-tuned models often have higher CFR and SAA as simulations progress. This may reflect the training distribution: fine-tuned models are trained to generate target turns conditioned on preceding dialogue (\autoref{fig:dataset_format}), so later simulation turns more closely resemble the fine-tuning setting.

\paragraph{Institutional fidelity.}
Baseline and fine-tuned models both perform reasonably well on several structural aspects of institutional fidelity. Coverage is strong for some baselines, especially \texttt{Qwen} ($0.80$--$0.81$), though fine-tuning substantially improves \texttt{GPT} coverage from $0.49$--$0.63$ to $0.88$--$0.94$. Ordering is high in most settings, with baseline scores from $0.64$ to $0.94$ and fine-tuned scores from $0.83$ to $0.93$, suggesting that prompt-only models can often preserve the broad meeting sequence. The clearest benefit of fine-tuning is vote attempts: \texttt{GPT} increases from $0.13$--$0.17$ to $0.47$--$0.55$, \texttt{LLaMA} from $0.09$--$0.10$ to $0.30$, and \texttt{Qwen} from $0.13$--$0.16$ to $0.30$--$0.32$. This likely reflects repeated exposure to motions, calls for votes, and decision points in the training transcripts.

\paragraph{Behavioral coherence.}
Both baseline and fine-tuned models often achieve high behavioral coherence. Action tags seem to help align expressed stances with later votes; in the baseline setting, stance-vote consistency rises from $0.812$ to $0.889$ for \texttt{GPT}, $0.653$ to $0.925$ for \texttt{LLaMA}, and $0.312$ to $0.667$ for \texttt{Qwen}; in the fine-tuned setting, tags improve \texttt{LLaMA} from $0.627$ to $0.740$ and \texttt{Qwen} from $0.578$ to $0.589$, though \texttt{GPT} decreases from $0.856$ to $0.701$. Fine-tuning itself does not consistently improve stance-vote consistency, but it does improve deliberative responsiveness for weaker baselines: \texttt{GPT} rises from $0.523$--$0.571$ to $0.632$--$0.679$, and \texttt{Qwen} rises from $0.465$--$0.483$ to $0.786$--$0.804$.

\paragraph{Human evaluation.}
In Turing tests, MTurk workers correctly distinguished human from simulated transcripts only 45.5\% of the time (below the 50\% expected by chance; see Appendix~\ref{app:simulation_result}, Fig.~\ref{fig:mturk_simulation_main}), indicating that simulations were largely indistinguishable from real transcripts.  
Detection varied by model: \texttt{GPT} simulations were identified 60\% of the time, versus 36\% for \texttt{LLaMA} and 42\% for \texttt{Qwen}. This likely reflects stylistic differences: \texttt{GPT}’s more structured, chain-of-thought reasoning yields coherent but unusually polished discourse, whereas \texttt{LLaMA} and \texttt{Qwen} produce less structured exchanges that appear more natural to evaluators.

\begin{figure}[h!]
\centering
\footnotesize 
\captionsetup{belowskip=1pt,aboveskip=1pt}

\begin{tcolorbox}[width=\columnwidth,left=0.5mm,right=0.5mm,bottom=0.5mm, colback=gray!10!white,colframe=black!50!white,title=(a) Albemarle County School Board Simulation]
\setlength{\parskip}{0pt} 
\setlength{\baselineskip}{0.95em}
\textbf{Graham Paige:} Good morning, everybody. I am calling this special board meeting for the Albemarle County School Board to order.\\
\textbf{David Oberg:} We're now going to have our moment of silence.\\
\textbf{Ellen Osborne:} All right, thank you. Shall we get started?\\
\textbf{Graham Paige:} Due to the pandemic, we must read this statement at least once a month into our meetings. The minutes of this meeting should reflect that this meeting is being held electronically, pursuant to Virginia state law and Albemarle County ordinance, because the COVID-19 pandemic makes it impracticable and unsafe to assemble a quorum physically in a single location. The meeting is being communicated to the public electronically through Zoom and streamed live on the Albemarle County Public Schools website. And public comment may be offered through the Zoom waiting room feature in accordance with the school board's protocol or by written submission.\\
\textbf{Jonno Alcaro:} Um, yeah, could we also do roll call?\\
\textbf{Judy Le:} Judy Le, White Hall District.\\
\textbf{Kate Acuff:} Kate Acuff, Jack Jouett District.\\
\textbf{Katrina Callsen:} Katrina Callsen, Scottsville District.\\
\textit{(Introductions continue …)}
\end{tcolorbox}

\begin{tcolorbox}[width=\columnwidth,left=0.5mm,right=0.5mm, bottom=1mm, colback=yellow!10!white,colframe=black!50!white,title=(b) DC Court of Appeals Simulation]
\setlength{\parskip}{0pt}
\setlength{\baselineskip}{0.95em}

\textbf{Courtroom Clerk:} All persons having business for the Honorable Chief Judge and Associate Judges are residing with the District of Columbia Court of Appeals. Draw near and give your attention. God save the United States and this Honorable Court. This Honorable Court is now in session. Please come forward.\\
\textbf{Judge Mcleese:} So good morning everyone. The court will take up the matter of John F. General versus USA. And uh, you're representing yourself today, right?\\
\textbf{John F. General:} Yes, Your Honor.\\
\textbf{Judge Deahl:} Can you tell us briefly what happened here? What did the police do wrong?\\
\textbf{John F. General:} Yes, Your Honor. I was brought to a station-house and questioned for over an hour and a half. I was not free to leave. The door was locked, and I was sharply questioned about multiple allegations, including prior uncharged conduct. During that time, they also took a saliva DNA sample from me without my formal consent.\\
\textbf{Judge Glickman:} Um, can you help me understand why you think you weren't free to leave? Wasn't it true that you could've gotten up and walked out anytime you wanted?
\end{tcolorbox}

\caption{(a) School Board simulation following typical meeting procedures (roll calls, introductions). (b) Court simulation following standard courtroom procedures (openings, questioning, representation verification).}
\label{fig:dialogs}
\end{figure}

\paragraph{Emergent Procedural Behaviors}
\autoref{fig:dialogs} shows representative simulations from fine-tuned models for an Albemarle County School Board meeting and a DC Court of Appeals hearing. In both cases, generated dialogue is not only coherent, but also follows recognizable institutional structure. The school board simulation moves through a call to order, a moment of silence, a mandated public-meeting statement, roll call, and district-based introductions. The court simulation similarly follows an expected sequence: clerk opening, case introduction, representation check, and questioning.

Across the broader set of simulations, we observe similar recurring procedural moves, including pledges of allegiance, agenda amendments, formal acknowledgments, and transitions into votes or questioning (Appendix~\ref{app:simulation_result}). These examples complement the institutional-fidelity metrics above, suggesting that action-aware training helps models recover not only speaker style but also some of the procedural scaffolding of civic deliberation.

\begin{figure*}[h!]
    \centering
    \includegraphics[width=1.0\textwidth]{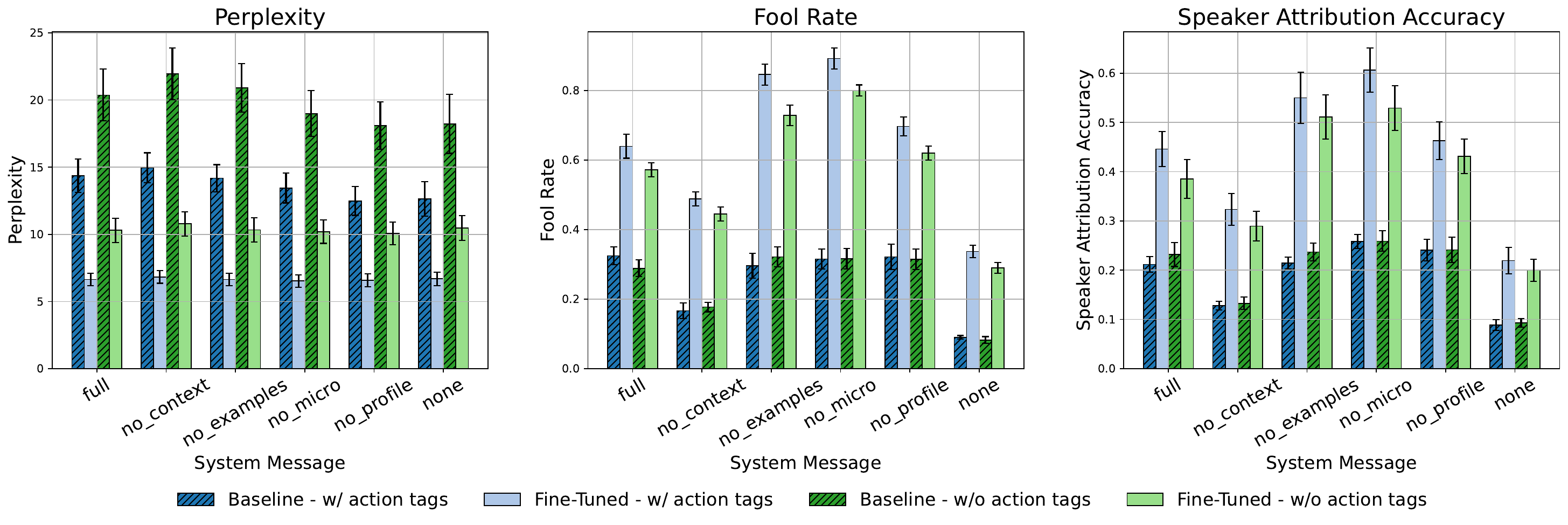}
    \caption{
Summary of persona fidelity metrics across system message configurations. Perplexity ($\downarrow$ better), fool rate ($\uparrow$ better), and speaker attribution accuracy ($\uparrow$ better), are averaged over all datasets and model families.
    }
    \label{fig:system_message}

\end{figure*}

\subsection{Ablations}

\noindent \textbf{Prompt Structure}
We ablate six system-prompt variants and report persona-fidelity results in \autoref{fig:system_message}; prompt details are in Appendix~\ref{app:system_prompts}. The main pattern is that prompting matters most before fine-tuning. Non-fine-tuned models are brittle: removing persona cues (\texttt{no\_profile}) or the full system prompt (\texttt{none}) sharply reduces CFR and SAA, often causing models to fall back to generic summarization. Fine-tuned models are less prompt-dependent. Since they already encode speaker-specific behavior, concise prompts often work better, while extra details such as \texttt{no\_micro} or \texttt{no\_examples} can introduce noise and slightly degrade consistency.

\vspace{0.1cm}
\noindent \textbf{Temporal Grounding}
To isolate the effect of temporal cues, we compare time-aware and time-unaware variants of the prompt-only baseline without action tags (Appendix~\ref{app:simulation_result}, Fig.~\ref{fig:simulation_comparison}). Temporal grounding improves agenda-item coverage for all models: \texttt{GPT} ($0.626 \rightarrow 0.830$), \texttt{LLaMA} ($0.467 \rightarrow 0.949$), and \texttt{Qwen} ($0.815 \rightarrow 0.907$). It also increases vote attempts, especially for \texttt{LLaMA} ($0.093 \rightarrow 0.591$) and \texttt{Qwen} ($0.156 \rightarrow 0.500$). Agenda-order effects are smaller and model-dependent. Overall, lightweight temporal grounding helps simulations stay aligned with the agenda and reach later procedural milestones.

\begin{figure}[h]
    \centering
    \includegraphics[width=0.65\columnwidth]{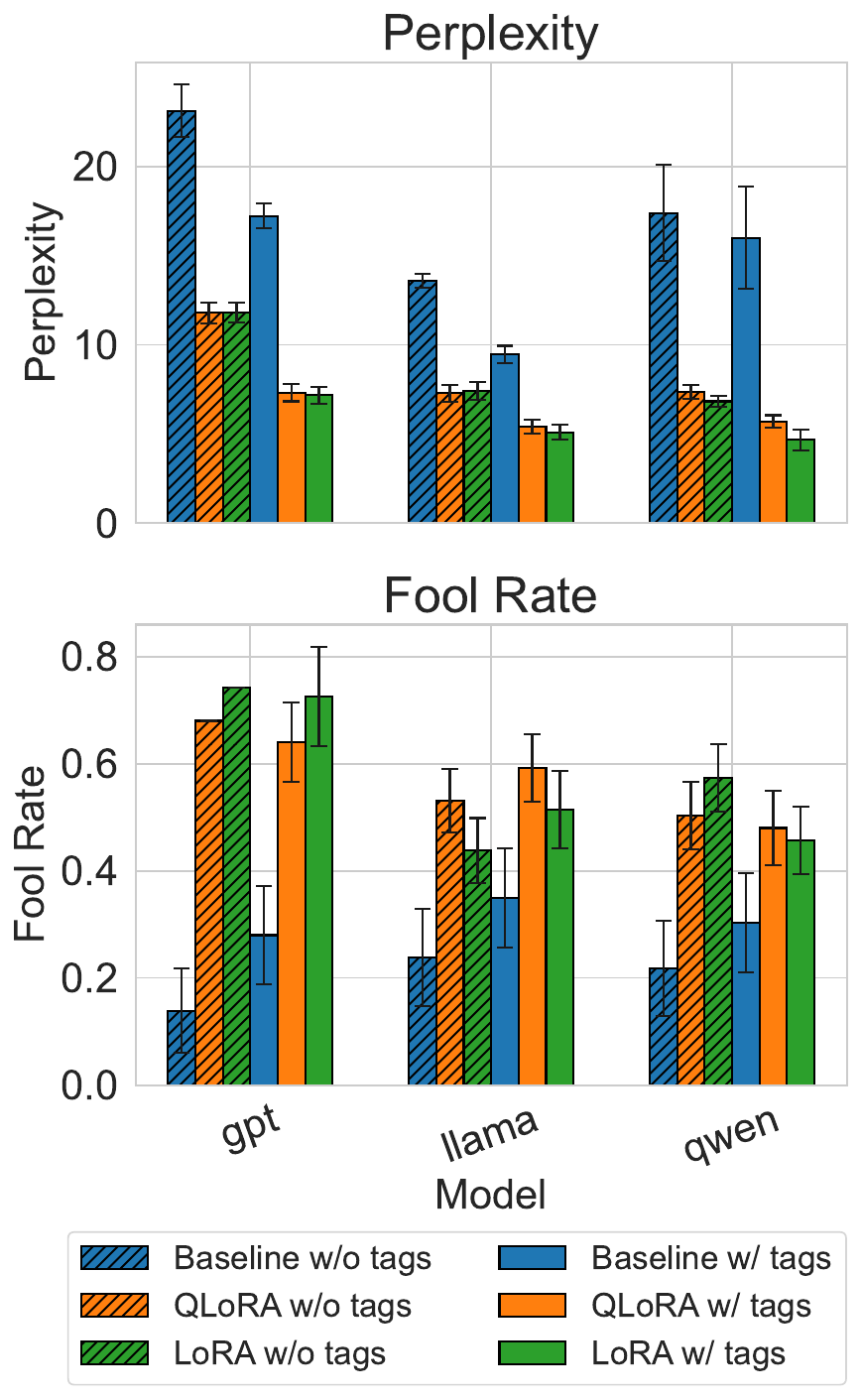}
    \caption{LoRA achieves marginally lower perplexity than QLoRA but differences in fool rate are negligible.}
    \label{fig:LORA_perplexity}
\end{figure}

\vspace{0.1cm}
\noindent \textbf{Fine-Tuning Method}
We compare QLoRA and LoRA by fine-tuning the speaker with the most training data in each dataset. LoRA achieves slightly lower PPL, but CFR and SAA are nearly unchanged (Appendix~\ref{app:fine-tune-ab}; \autoref{fig:LORA_perplexity}), suggesting comparable generation quality. We use QLoRA as the default for its memory efficiency.




\section{Conclusion}
\label{sec:discussion}

We presented a reproducible pipeline for constructing speaker-attributed, action-annotated transcripts from public meeting videos, and used it to release three datasets of government deliberation. Together with our evaluation framework, these resources support data-grounded study of civic simulations across persona fidelity, persona consistency, institutional fidelity, and behavioral coherence. Our results show that structured metadata improves generation, while fine-tuning improves CFR/SAA, reduces persona drift, and increases procedural outcomes such as vote attempts.

Data-grounded simulations can make institutional behavior more measurable and experimentally controllable. This enables controlled studies of procedural counterfactuals, while future work should move toward richer interaction dynamics and stronger decision-level validation.

\section*{Limitations}
The method requires access to Zoom meetings recorded in gallery view, which limits its applicability. Many real-world deliberations, collaborative environments, or public events are not recorded this way, restricting the universality of the approach.  That said, gallery view recordings exist across a wide range of domains including academic seminars and public-interest groups; providing substantial breadth for empirical study. This data can meaningfully inform practical questions, such as how workplace teams might be structured to improve coordination or how lectures and discussions could be organized to strengthen student engagement and learning outcomes. 

The approach also depends on Whisper ASR transcriptions. Real-world meetings frequently include overlapping speech, variable audio quality, and background noise, all of which can reduce transcription accuracy. This issue is amplified in less structured environments, such as debates or public hearings, where interruptions and crosstalk are common.  In such settings, additional preprocessing steps may be required. Thus, while Whisper performs well in many contexts, the reliance on ASR-generated text imposes practical constraints, particularly for noisy or high-overlap audio.

Our work enables realistic simulation of deliberative processes, with both promising and concerning implications. Simulations could support civic participation, governance training, and policy exploration by democratizing access to deliberation dynamics. But realistic persona models also pose risks: generated transcripts could spread misinformation, manipulate public opinion, or displace genuine participation in democratic processes. We release datasets and code for research use, expecting users to consider these risks. Future work should establish community guidelines for responsible deployment of generative civic models, especially in high-stakes domains like policy-making and public engagement.

\bibliography{custom}
\FloatBarrier
\clearpage
\appendix

\section{Persona and Topic Extraction Prompts}
\label{app:extraction}

\begin{figure}[h!]
\footnotesize
\centering

\begin{tcolorbox}[colback=gray!3,colframe=black!15,
title=\textbf{Persona Extraction — Stage 1: Structured Extraction Prompt},
width=\linewidth]

\textbf{System:} You are a precise extractor. Only extract information explicitly stated or strongly implied. Return only JSON, nothing else.

\medskip
\textbf{Task:} For each numbered monologue, extract the following fields:
\texttt{identity}, \texttt{goals}, \texttt{tone}, \texttt{firm\_boundaries},
\texttt{beliefs\_policy}, \texttt{evidence\_spans}, and \texttt{confidence}.

\medskip
\textbf{Output:}

\begin{verbatim}
{
  "monologue_extractions": [
    {
      "index": 0,
      "identity": [...],
      "goals": [...],
      "tone": [...],
      "firm_boundaries": [...],
      "beliefs_policy": [...],
      "evidence_spans": [...],
      "confidence": "low/medium/high"
    }
  ]
}
\end{verbatim}

\end{tcolorbox}

\vspace{6pt}

\begin{tcolorbox}[colback=gray!3,colframe=black!15,
title=\textbf{Persona Extraction — Stage 2: Consolidation Prompt},
width=\linewidth]

\textbf{System:} You are an expert at synthesizing persona traits into coherent, compact descriptions.

\medskip
\textbf{User:} Given a collection of per-monologue extractions from one speaker,
consolidate them into 1–3 short paragraphs (30–200 tokens).

Capture the speaker’s \textit{identity, goals, tone, boundaries, and policy beliefs}
using natural descriptive prose.

\medskip
Avoid redundancy and do not invent any information beyond what is supported by the text.
Return a concise persona summary suitable for downstream dialogue conditioning.

\end{tcolorbox}

\caption{
Two-stage prompting strategy for persona extraction.
\textbf{Stage 1} uses a schema-based \texttt{GPT-5} prompt to extract structured communicative traits from each speaker’s longest monologues.
\textbf{Stage 2} merges these traits into a coherent natural-language persona summary enabling persona-aware modeling in deliberative settings.
}
\label{fig:persona_prompts}

\end{figure}


\begin{figure}[t]
\footnotesize
\centering

\begin{tcolorbox}[colback=gray!3,colframe=black!15,
title=\textbf{Topic Extraction — Stage 1: Chunk Summarization Prompt},
width=\linewidth]

\textbf{System:} You are an assistant that extracts the main discussion topics
from a council meeting transcript. Keep summaries concise and clear.

\medskip
\textbf{User:} Summarize the main topics discussed in the following meeting
transcript chunk in 2--3 bullet points:

\medskip
\texttt{[Chunk text of up to 1,024 tokens here...]}

\end{tcolorbox}

\vspace{6pt}

\begin{tcolorbox}[colback=gray!3,colframe=black!15,
title=\textbf{Topic Extraction — Stage 2: Topic Consolidation Prompt},
width=\linewidth]

\textbf{System:} You are an assistant that consolidates multiple bullet-point
summaries into a concise final list of meeting topics.

\medskip
\textbf{Rules:}

\begin{itemize}
\item Merge duplicate or overlapping items
\item Organize notes by topic
\item Limit to $\leq 5$ main topics, each with $\leq 2$ bullet points
\item Include only information clearly supported by the summaries
\end{itemize}

\textbf{User:} Here are summaries of meeting chunks.
Consolidate and organize notes by topic.

\medskip
\texttt{[Summarized transcript chunks]}

\end{tcolorbox}

\caption{
Two-stage prompting strategy for topic extraction.
\textbf{Stage 1} summarizes transcript chunks (up to 1,024 tokens) into concise notes using \texttt{GPT-5}.
\textbf{Stage 2} consolidates these summaries into canonical meeting topics,
merging overlapping content and producing a compact representation of the discussion flow.
}
\label{fig:topic_extraction_prompts}

\end{figure}

\FloatBarrier
\clearpage

\section{Dataset Format}
\label{app:chatml_format}
{\setlength{\stripsep}{4pt}
\begin{strip}
\begin{nolinenumbers} 
\centering
\small
\begin{tcolorbox}[colback=gray!5,colframe=gray!40,arc=2mm,width=1.0\textwidth]
\textbf{Raw Transcript} \\[-2pt]
\textbf{grahampaige:} Tonight is Elizabeth's last official night, so we have a special resolution and letter that will be mailed to you, Elizabeth. Dear Elizabeth, all of us on the school board want to offer our congratulations to you this evening, not just because you are completing three months of service to our entire school community, but because of the purpose you had in mind when you applied to join us. You wrote on your application that this position would allow you to bring a more positive outcome on your community by representing student voice. As a member of our Hispanic community, you wanted to ensure that minority voices would be heard. You belong to Monticello High School's Spanish, Math, and Art Honor Societies, and your transcript reveals no fewer than nine AP classes. Let us add one more: the School Board Student Representative Honor Society. Your service has brought our strategic plan’s four values—equity, excellence, family and community, and wellness—to life. Please accept our appreciation for your leadership. When we created this position, our hope was that it would enhance the desire of our young leaders to make the world a better place. On this test, let me add one more grade to your transcript: A+. So thank you so much for your contributions, Elizabeth. Do you have any words to tell us? \\[4pt]
\textbf{elizabethalbarran:} Yeah, I just wanted to say thank you for the opportunity and for allowing me to speak on behalf of the student body, which I really appreciate. It was a great learning experience, and I thank you for that. \\[4pt]
\textbf{grahampaige:} Thank you so much, Elizabeth. And your class load is just really, really spectacular—to know that you have nine AP classes and all the other things you’re doing. So thanks so much for your contribution to the board. We’ve really appreciated having you on. \\[4pt]
\textbf{davidoberg:} Mr.\ Chair, may I ask a point of personal privilege? Since we’ve not gotten to meet in person, could we please invite all past student representatives to the next meeting to be recognized? \\[4pt]
\textbf{grahampaige:} Very good idea, Mr.\ O’Berg. Ms.\ Johnson, could you begin working on that? Thank you. So Elizabeth, hopefully we’ll see you in April at the county office building—or whenever we end up doing this. \\[6pt]

\textbf{Non-Tagged Dataset Format} \\
\texttt{[ \{ 'role': 'assistant', 'content': "grahampaige: ...Please accept our appreciation for your leadership. When we created this position, our hope was that it would enhance the desire of our young leaders to make the world a better place. On this test, let me add one more grade to your transcript: A+. So thank you so much for your contributions, Elizabeth. Do you have any words to tell us?" \},} \\
\texttt{\{ 'role': 'user', 'content': "elizabethalbarran: Yeah, I just wanted to say thank you for the opportunity..." \},} \\
\texttt{\{ 'role': 'assistant', 'content': "grahampaige: Thank you so much, Elizabeth. And your class load is really spectacular..." \},} \\
\texttt{\{ 'role': 'user', 'content': "davidoberg: Mr. Chair, may I ask a point of personal privilege..." \},} \\
\texttt{\{ 'role': 'assistant', 'content': "grahampaige: Very good idea, Mr. O’Berg. Ms. Johnson, could you begin working on that..." \} ]} \\[6pt]

\textbf{Tagged Dataset Format} \\
\texttt{[ \{ 'role': 'assistant', 'content': "grahampaige: .... Please accept our appreciation for your leadership. When we created this position, our hope was that it would enhance the desire of our young leaders to make the world a better place. On this test, let me add one more grade to your transcript: A+. So thank you so much for your contributions, Elizabeth. Do you have any words to tell us?" \},} \\
\texttt{\{ 'role': 'user', 'content': "elizabethalbarran: [ACKNOWLEDGE] [PUBLIC\_ADDRESSING] Yeah, I just wanted to say thank you for the opportunity..." \},} \\
\texttt{\{ 'role': 'assistant', 'content': "grahampaige: [ACKNOWLEDGE] [OPINION] [PUBLIC\_ADDRESSING] Thank you so much, Elizabeth. And your class load is really spectacular..." \},} \\
\texttt{\{ 'role': 'user', 'content': "davidoberg: [PROCEDURAL\_MOVE] [PUBLIC\_ADDRESSING] [REQUEST\_ACTION] Mr. Chair, may I ask a point of personal privilege..." \},} \\
\texttt{\{ 'role': 'assistant', 'content': "grahampaige: [ACKNOWLEDGE] [LOGISTICS\_INFO] [REQUEST\_ACTION] Very good idea, Mr. O’Berg. Ms. Johnson, could you begin working on that..." \} ]} \\[6pt]

\end{tcolorbox}

\captionof{figure}{
Illustration of dataset encoding and truncation logic. Speaker turns are segmented, serialized into a compact ChatML-style 
and optionally augmented with communicative function tags (e.g., [ACKNOWLEDGE]). 
Each training instance consists of a context window of preceding utterances and a target turn from the speaker of interest. 
Context is truncated at 1{,}024 tokens to maintain coherence within computational limits; when truncation occurs mid-sentence, 
the visible text is prefixed with an ellipsis (\texttt{...}) to indicate omitted material. 
Action tags are prepended and alphabetically sorted before tokenization.
}

\label{fig:dataset_format}
\end{nolinenumbers}
\end{strip}
\FloatBarrier

\section{Finding Datasets}
\label{app:finding_datasets}
The proposed method focuses on “Zoom gallery-style” recordings. While such videos may seem rare, a substantial number are publicly accessible. To identify them systematically, the YouTube Data API \cite{youtube_data_api_docs_2025} was queried using three terms: \textit{``zoom meeting,'' ``virtual meeting,''} and \textit{``zoom webinar,''} targeting content uploaded in 2020—a period marked by a surge in remote meetings due to the COVID-19 pandemic. Only videos longer than 20 minutes were considered, as longer recordings are more likely to capture full multi-participant meetings rather than brief tutorials or unrelated content. Searches were conducted in one-month increments to ensure comprehensive coverage, yielding roughly 10,000 candidate videos.

Not all candidates depicted gallery-style meetings. To refine the selection, a ResNet50 model \cite{he2016deep} was fine-tuned to classify whether a video thumbnail represented a Zoom gallery-style meeting. Positive examples were sampled from frames of 100 curated gallery-style meetings, while negatives were drawn from thumbnails of diverse YouTube content, including news, comedy, and entertainment. Applying this model reduced the candidate set by roughly 40\%, resulting in approximately 6,000 gallery-style videos.

The 6,000 videos identified represent only a small fraction of the full corpus of gallery-style recordings. The search was limited to three queries, 2020 uploads, videos longer than 20 minutes, and thumbnail-based classification. The goal of this exercise, however, is not to quantify the total population, but rather to demonstrate the diversity of meetings that can be diarized and simulated using our proposed pipeline. Even within this subset, the videos encompass a wide range of meeting types, participant configurations, and interaction patterns. Channel IDs and descriptions were extracted and categorized using \texttt{GPT-5}, producing the distribution shown in Figure~\ref{fig:zoom1}, which showcases the diverse and interesting types of meetings that our pipeline can simulate.

\begin{figure}[t]
    \centering
    \includegraphics[width=0.45\textwidth]{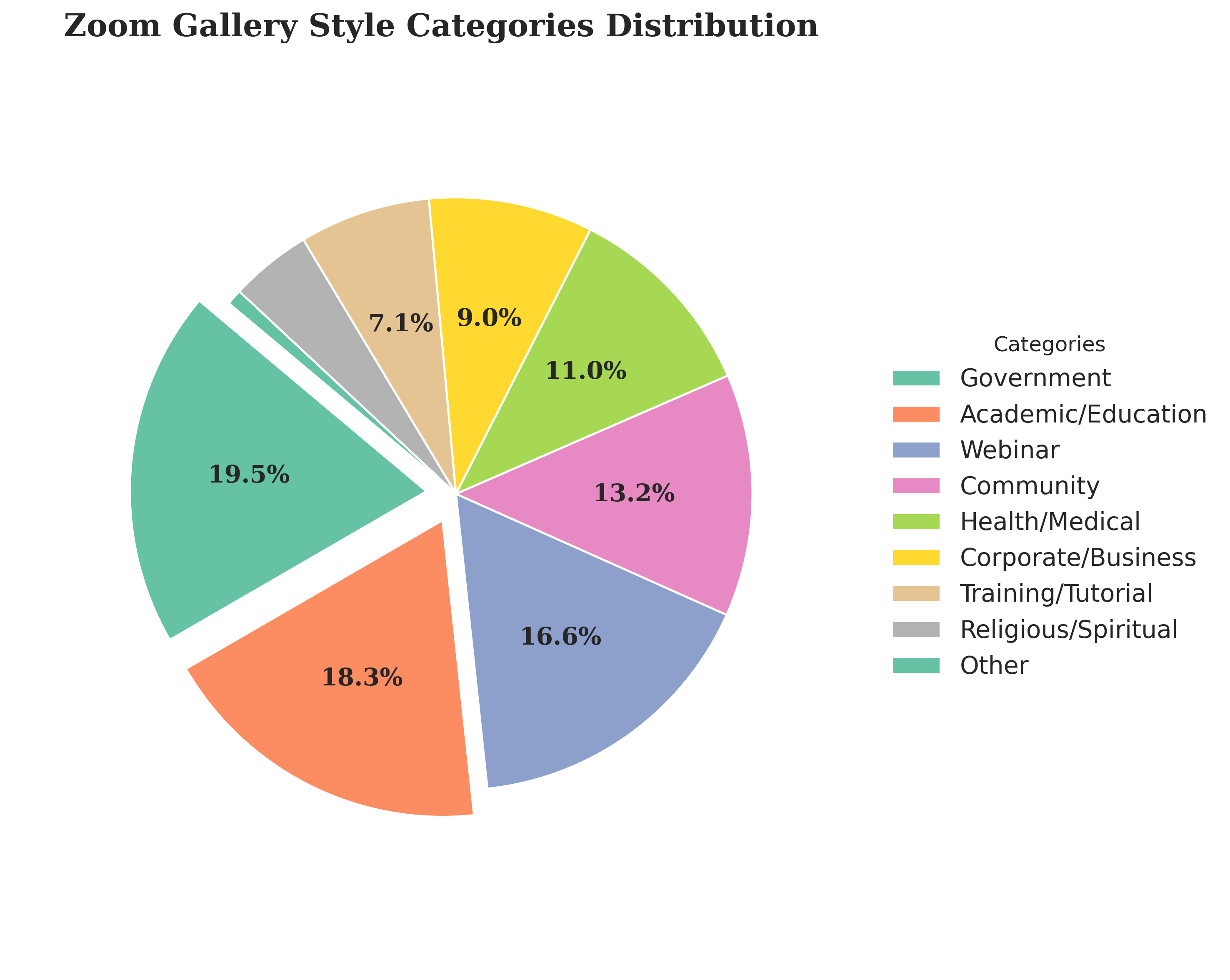}
    \caption{Categorization of publicly available gallery-style Zoom videos.  Candidate videos were retrieved using three search queries (\textit{``zoom meeting,'' ``virtual meeting,'' ``zoom webinar''}) and filtered by length (>20 minutes). Thumbnails were classified using a fine-tuned ResNet50 model to identify gallery-style recordings. Channel IDs and descriptions of the resulting 6,000 videos were then analyzed using \texttt{GPT-5} to categorize channels into broad content types.
    This figure illustrates a wide range of meeting types captured that are publically available and can be studied using our pipeline.}
    \label{fig:zoom1}

\end{figure}
\FloatBarrier
\clearpage
\section{Dataset Descriptions}
\label{app:datasets}
\paragraph{DC Court of Appeals Dataset}
The DC Court of Appeals dataset consists of recordings of hearings from the District of Columbia’s highest court, each involving three judges and one appellee. Judges are appointed by the President and confirmed by the Senate for 15-year terms, and the court also oversees attorney conduct through the D.C. Bar..  These sessions capture formal proceedings—judges’ questions, deliberations, and interactions with the appellant—resulting in highly structured, high-stakes multi-party exchanges. The dataset is well-suited for studying judicial reasoning, formal deliberation, and role-specific communication patterns. Although it is the smallest of the three datasets, each transcript is dense, and a custom action-tag taxonomy captures typical communicative patterns (\autoref{tab:dc_taxonomy}).

The dataset preserves speaker-specific lexical and rhetorical patterns, enabling detailed modeling of persona and conversational style. Table~\ref{tab:all_speaker_summaries} summarizes per-judge statistics, including total tokens, average utterance length, and utterance counts, illustrating differences in participation and speech style. Additional participants, such as the appellee, are included in the transcripts and can be analyzed, though their word counts are not included in these statistics.

Overall, this dataset offers a unique opportunity to study multi-party deliberation in a legally constrained environment, providing a foundation for evaluating fine-tuned models’ ability to replicate persona-specific communication and structured interactions.

\paragraph{Albermale County School Board Dataset}
The Albemarle County School Board dataset captures recordings of meetings held by the seven-member board that governs Albemarle County Public Schools. Meetings typically involve board members, the superintendent, administrative staff, and members of the public, resulting in a mixture of semi-formal deliberation, policy discussion, and public commentary. While the dataset is moderate in the number of meetings, sessions often include a large number of participants, giving it the highest average attendance across the three datasets.

The School Board and Superintendent work in partnership to guide the school system’s priorities and ensure that the district meets its educational goals. This collaborative governance framework creates a rich context for studying multi-party deliberation, decision-making dynamics, and the interplay between formal roles and public input.

Each transcript preserves speaker-specific behavior, including speech patterns, rhetorical strategies, and agenda-driven contributions. Community members’ participation is included in the transcripts, providing additional depth for analyzing interaction patterns and multi-party dialogue. A custom action-tag taxonomy captures common communicative acts (\autoref{tab:albemarle_taxonomy}), supporting detailed modeling of procedural behaviors. Table~\ref{tab:all_speaker_summaries} summarizes board members’ participation, including average utterance length, total tokens, and utterance counts, illustrating variation in engagement and style.

\paragraph{Waipā District Council Dataset}
The Waipā District Council dataset contains recordings from multiple council committees, capturing discussions among councillors, the Mayor, committee chairs, and community representatives. As the largest corpus in this collection, it spans a wide range of semi-formal governance interactions, with rich multi-party exchanges across several committees. While this study primarily focuses on the Audit and Risk Committee, the broader dataset includes meetings from other committees—such as Strategic Planning \& Policy, Finance \& Corporate, and Sustainability \& Climate Change—enabling additional analyses in future work.

The Waipā District Council governs a municipality in New Zealand’s Waikato region, with Cambridge and Te Awamutu as its main towns. The council operates across five wards, overseeing local governance, resource allocation, and community services for a population of approximately 62,200. Council meetings reflect formal decision-making processes, procedural deliberation, and community engagement, providing a rich context for studying multi-party interaction, role-based behavior, and deliberative dynamics in a local government setting.

Each transcript preserves detailed speaker-specific behavior, including speech patterns, rhetorical style, and agenda-driven contributions. Action tags have been defined to capture recurring communicative acts and procedural markers (\autoref{tab:waipa_taxonomy}). Table~\ref{tab:all_speaker_summaries} reports per-member statistics for the Audit and Risk Committee, including total tokens, average utterance length, and utterance counts. While the table focuses on committee members, contributions from community representatives are included in the transcripts, providing additional richness for analyzing multi-party dialogue and emergent procedural behavior.

\begin{table}[t]
\centering
\begin{subtable}[t]{\columnwidth}
\centering
\resizebox{\linewidth}{!}{%
\begin{tabular}{lrrr}
\hline
\textbf{Judge Name} & \textbf{Avg. Utterance} & \textbf{Total Tokens} & \textbf{Utterance Count} \\
\hline
Judge McLeese    & 106.08 & 76,801 & 724 \\
Judge Glickman   & 89.05 & 28,406 & 319  \\
Judge Deahl      & 78.85 & 22,314 & 283  \\
Judge Easterly   & 109.63 & 20,721 & 189  \\
Judge Blackburne & 70.73 & 19,380 & 274  \\
Judge Thompson   & 56.30 & 18,073 & 321  \\
\hline
\end{tabular}%
}
\caption{Per-judge statistics for DC Court of Appeals transcripts.}
\end{subtable}

\begin{subtable}[t]{0.9\linewidth}
\centering
\resizebox{\linewidth}{!}{%
\begin{tabular}{lrrr}
\hline
\textbf{School Board Member} & \textbf{Avg. Utterance} & \textbf{Total Tokens} & \textbf{Utterance Count} \\
\hline
Ellen Osborne    & 79.97  & 4,798    & 60   \\
David Oberg      & 77.40  & 32,044   & 414  \\
Graham Paige     & 48.73  & 162,920  & 3,343 \\
Jonno Alcaro     & 67.71 & 63,110   & 932  \\
Katrina Callsen  & 90.52 & 91,965  & 1,016 \\
Kate Acuff       & 74.17  & 41,017   & 553  \\
Judy Le          & 68.44  & 23,269   & 340  \\
\hline
\end{tabular}%
}
\caption{Per-member statistics for Albemarle County School Board.}
\end{subtable}

\begin{subtable}[t]{0.9\linewidth}
\centering
\resizebox{\linewidth}{!}{%
\begin{tabular}{lrrr}
\hline
\textbf{Council Member} & \textbf{Avg. Utterance} & \textbf{Total Tokens} & \textbf{Utterance Count} \\
\hline
Susan O'Regan    & 93.76  & 163,076  & 1,739 \\
Jim Mylchreest   & 80.97  & 197,770  & 2,442 \\
Clare St Pier    & 100.28  & 139,495   & 1,391 \\
Andrew Brown     & 109.46  & 16,090   & 147   \\
Roger Gordon     & 103.63  & 23,317   & 225   \\
Lou Brown        & 82.31  & 35,970   & 437   \\
Ange Holt        & 74.82  & 175,090  & 2,340 \\
Sue Milner       & 59.16  & 123,947   & 2,095 \\
\hline
\end{tabular}%
}
\caption{Per-member statistics for Waipā District Council participants.}
\end{subtable}

\caption{Per-speaker statistics for all transcripts.}
\label{tab:all_speaker_summaries}
\end{table}

\section{Action Tagging Taxonomies}
\label{app:action_tagging}

This appendix presents the action-tag taxonomies used for our three meeting corpora: DC Court of Appeals (Table~\ref{tab:dc_taxonomy}), Albemarle County School Board (Table~\ref{tab:albemarle_taxonomy}), and Waipā District Council (Table~\ref{tab:waipa_taxonomy}). Each table lists the dataset-specific tags, their corresponding unified category, a short definition, and a canonical example.

The taxonomy was designed to capture communicative acts central to each dataset’s institutional context while remaining compact and practical for both human annotation and downstream computational modeling. When constructing the taxonomy, we followed several guiding principles to ensure clarity, consistency, and analytical utility:

\begin{itemize}
    \item \textbf{Action-focused} --- Tags capture the speaker’s communicative intent (speech act or function) rather than emotional stance or opinion outcomes. Focusing on actions rather than sentiment allows for consistent comparisons across speakers and institutions, and ensures that models trained on these tags learn to predict functional behavior rather than subjective interpretation.
    
    \item \textbf{Compact and informative} --- Each dataset uses a concise tagset (typically 15--30 tags) that maximizes analytical signal while minimizing cognitive load for annotators. A smaller, focused set of tags reduces ambiguity, ensures faster and more reliable annotation, and allows downstream models to more effectively learn patterns of interaction without being overwhelmed by an excessive number of classes.
    
    \item \textbf{Utterance-based} --- The primary unit of annotation is the utterance, reflecting the natural granularity of spoken or transcribed meetings. Multiple tags can be applied when an utterance serves multiple functions, while span-level annotations are reserved for specialized cases such as legal citations. This design ensures that annotations align closely with how participants actually communicate, making both human and automated analysis more interpretable.
    
    \item \textbf{Cross-context applicability} --- While each corpus comes from a different institutional setting, the taxonomy is designed to be flexible enough to capture similar communicative acts across contexts. This supports comparative studies of deliberation, interaction dynamics, and procedural management across institutions, while preserving dataset-specific richness through the mapping to unified categories.
    
    \item \textbf{Practical for computational modeling} --- The taxonomy was constructed with downstream modeling in mind. Categories are mutually intelligible, semantically coherent, and of sufficient frequency to allow robust learning. Prepending tags to the text of each utterance (as shown in Figure~\ref{fig:dataset_format}) facilitates easy integration with models for classification, prediction, and generation tasks.
\end{itemize}

To facilitate cross-dataset analysis, the dataset-specific tagsets were mapped to a set of unified categories:

\begin{itemize}
    \item \textbf{INFO} --- providing or clarifying factual or procedural information.
    \item \textbf{ASK} --- requesting information, clarification, or explanation.
    \item \textbf{ARG} --- making arguments, expressing opinions, or persuading others.
    \item \textbf{CITE} --- explicit reference to authoritative sources such as cases, statutes, or reports.
    \item \textbf{PROC} --- procedural moves that manage session flow, agenda, or turn-taking.
    \item \textbf{INTERACT} --- short, interactive acts such as acknowledgments, agreement/disagreement, interruptions, or backchannels.
    \item \textbf{ACTION} --- calls for action or assignments of concrete tasks.
    \item \textbf{LEGIT} --- statements oriented toward legitimacy, reassurance, or public accountability.
\end{itemize}

For efficient and consistent annotation, utterances were passed to \texttt{GPT-5} one at a time along with the corresponding tag set, and the model was prompted to assign the appropriate tags to each utterance.

\begin{table*}[htbp]
\centering
\small
\caption{DC Court of Appeals — action tags}
\label{tab:dc_taxonomy}
\begin{tabularx}{\textwidth}{@{} l l X X @{}}
\toprule
\textbf{Tag} & \textbf{Category} & \textbf{Definition} & \textbf{Example} \\
\midrule
\texttt{court\_opening} & PROC & Ritual invocation bringing the court into session. & ``The Court is now in session.'' \\
\texttt{case\_call} & PROC & Clerk/judge announcing the docketed case by name/number. & ``Our first case is No.\ 19-CV-0725.'' \\
\texttt{court\_closing} & PROC & Formal adjournment wording. & ``Court is adjourned.'' \\
\texttt{court\_procedural\_remark} & PROC & Administrative/session logistics (recesses, technical notes). & ``We'll take a ten-minute recess.'' \\
\texttt{opening\_argument} & ARG & Counsel's allocated opening statement at the start of their time. & ``May it please the Court, counsel for the appellant...'' \\
\texttt{substantive\_argument} & ARG & Core legal reasoning applying law to facts. & ``Under Smith, the statute requires X of Y.'' \\
\texttt{legal\_citation} (span) & CITE & Span-level explicit reference to case law, statutes, or rules. & ``See Roe v.\ Wade, 410 U.S. 113 (1973).'' \\
\texttt{fact\_statement} & INFO & Recitation or summary of facts from the record. & ``The trial transcript shows on p.\ 12 that...'' \\
\texttt{policy\_argument} & ARG & Argument that appeals to broader policy consequences or systemic effects. & ``Allowing this rule would create perverse incentives...'' \\
\texttt{concession} & ARG / INTERACT & Counsel explicitly acknowledges a weakness or adverse point. & ``We concede that the court found...'' \\
\texttt{rebuttal\_argument} & ARG & Statements made during rebuttal addressing opposing points. & ``In rebuttal, the appellant notes...'' \\
\texttt{reserve\_time} & PROC & Explicit reservation of rebuttal time (procedural). & ``I reserve five minutes for rebuttal.'' \\
\texttt{judge\_question} & ASK & Judge poses a clarifying, hypothetical, or probing question. & ``How does that affect standing?'' \\
\texttt{judge\_comment} & INTERACT & Non-question remarks from a judge (observations or tentative views). & ``I'm troubled by the text of the statute.'' \\
\texttt{judge\_interruption} & INTERACT & Judge interrupts or cuts off counsel (often to redirect). & [interruption during counsel's turn] \\
\texttt{judge\_procedural\_directive} & PROC & Procedural instruction from the bench (e.g., time/order). & ``Please answer the jurisdictional issue first.'' \\
\texttt{acknowledge} & INTERACT & Brief interactive receipts (``Yes, Your Honor''). & ``Thank you, Judge.'' \\
\texttt{hedge} & INTERACT & Cautious or tentative language (``I think'', ``perhaps''). & ``I think the better reading is...'' \\
\texttt{agreement} & INTERACT & Explicit acceptance of a prior proposition. & ``I agree.'' \\
\texttt{disagreement} & INTERACT & Explicit rejection or objection. & ``I disagree.'' \\
\texttt{time\_management} & PROC & Remarks about elapsed or remaining time. & ``You have two minutes left.'' \\
\texttt{case\_transition} & PROC & Movement to the next case or docketed matter. & ``Next case, No.\ 20-CV-...'' \\
\texttt{record\_reference} & INFO / CITE & Explicit pointer to the record (page/line/exhibit). & ``The record at R.12--13 shows...'' \\
\bottomrule
\end{tabularx}
\end{table*}

\begin{table*}[htbp]
\centering
\small
\caption{Albemarle County School Board — action tags}
\label{tab:albemarle_taxonomy}
\begin{tabularx}{\textwidth}{@{} l l X X @{}}
\toprule
\textbf{Tag} & \textbf{Category} & \textbf{Definition} & \textbf{Example} \\
\midrule
\texttt{ask\_question} & ASK & Request for information or clarification from other participants. & ``Can you explain the budget increase?'' \\
\texttt{give\_update} & INFO & Status reports or operational updates. & ``Enrollment increased by 3\% this year.'' \\
\texttt{request\_action} & ACTION & Calls for a specific task or action by staff or members. & ``Please circulate the revised policy.'' \\
\texttt{defer\_to\_expert} & ACTION / PROC & Pass authority or technical issues to an expert/official (delegation). & ``I'll defer to the Superintendent on that.'' \\
\texttt{express\_concern} & ARG / INTERACT & Raise a risk, worry, or problem. & ``I'm concerned about safety on campus.'' \\
\texttt{reassure} & LEGIT & Attempts to calm or assure stakeholders. & ``We will monitor closely and report back.'' \\
\texttt{clarify} & INFO & Correction or clarifying restatement. & ``To be clear, the date is June 1.'' \\
\texttt{policy\_proposal} & ARG & Proposal of a policy, rule, or initiative. & ``I move we adopt policy X.'' \\
\texttt{prioritization} & ARG & Statements identifying or ranking priorities. & ``We should prioritize safety over aesthetics.'' \\
\texttt{logistics\_info} & PROC & Practical scheduling or operational details. & ``The next meeting is Tuesday.'' \\
\texttt{hedge} & INTERACT & Tentative language and uncertainty markers. & ``It may be the case that...'' \\
\texttt{cite\_external\_source} & CITE & References to external reports, studies, or authorities. & ``According to CDC guidance...'' \\
\texttt{opinion} & ARG / INTERACT & Subjective statement of stance. & ``I believe this is the right course.'' \\
\texttt{acknowledge} & INTERACT & Minimal receipt tokens (``okay'', ``thanks''). & ``Thank you.'' \\
\texttt{agreeing} & INTERACT & Explicit agreement with another speaker. & ``I agree.'' \\
\texttt{disagreeing} & INTERACT & Explicit disagreement or objection. & ``I disagree.'' \\
\texttt{conceding} & INTERACT & Accepting another's point after discussion. & ``Fair---I accept that.'' \\
\texttt{persuaded} & INTERACT & Marked change of stance attributed to discussion. & ``I'm now convinced.'' \\
\texttt{follow\_up\_request} & ACTION & Request building on a prior turn (tasking). & ``Can staff prepare a memo?'' \\
\texttt{interrupt\_overlap} & INTERACT & Overlap or interruption in turn-taking. & [overlapping speech] \\
\texttt{public\_addressing} & LEGIT & Remarks explicitly directed to the public/community. & ``To the parents listening at home...'' \\
\texttt{procedural\_move} & PROC & Motions, points of order, and rule-of-order moves. & ``I move to table this item.'' \\
\texttt{vote\_call} & PROC & Initiation or conduct of a formal vote. & ``All those in favor, say `aye'.'' \\
\bottomrule
\end{tabularx}
\end{table*}

\begin{table*}[htbp]
\centering
\small
\caption{Waipā District Council — action tags}
\label{tab:waipa_taxonomy}
\begin{tabularx}{\textwidth}{@{} l l X X @{}}
\toprule
\textbf{Tag} & \textbf{Category} & \textbf{Definition} & \textbf{Example} \\
\midrule
\texttt{ask\_question} & ASK & Request for clarification or detail (probing). & ``Can you expand on the risk assessment?'' \\
\texttt{provide\_information} & INFO & Sharing of data, audit findings, or factual reports. & ``The audit found a 2\% variance.'' \\
\texttt{cite\_reference} & CITE & Reference to statutes, policy documents, or expert sources. & ``Under s.\ 12 of the Act...'' \\
\texttt{clarify\_point} & INFO & Rephrase or simplify for shared understanding. & ``To clarify, the intent is...'' \\
\texttt{express\_concern} & ARG / INTERACT & Highlight risks, trade-offs, or gaps. & ``My concern is the budget impact.'' \\
\texttt{state\_opinion} & ARG & Subjective judgement or stance. & ``I think we should delay.'' \\
\texttt{policy\_proposal} & ARG & Presentation of a concrete policy or option. & ``I propose we adopt option B.'' \\
\texttt{evaluate\_option} & ARG & Weigh pros and cons of alternatives. & ``This will save money but reduce capacity.'' \\
\texttt{prioritize\_issue} & ARG & Statements ranking issues by importance. & ``We must deal with water first.'' \\
\texttt{agree} & INTERACT & Explicit alignment with another speaker. & ``I agree with Councillor X.'' \\
\texttt{disagree} & INTERACT & Explicit opposition. & ``I disagree.'' \\
\texttt{build\_on\_point} & INTERACT & Constructive extension of another's idea. & ``To build on that, we could...'' \\
\texttt{persuasion\_attempt} & ARG / INTERACT & Rhetorical effort to change others' views. & ``Given the evidence, you should support this.'' \\
\texttt{defer\_to\_authority} & LEGIT / ACTION & Reliance on chair, staff, or expert for legitimacy or delegation. & ``I defer to the planning officer.'' \\
\texttt{procedural\_move} & PROC & Agenda control, motions, or governance mechanics. & ``I move we accept the minutes.'' \\
\texttt{call\_vote} & PROC & Initiate a formal vote. & ``All in favor?'' \\
\texttt{summarize\_outcome} & PROC & Chair/member crystallizes the decision or result. & ``The motion passes, 5--3.'' \\
\texttt{defer\_decision} & PROC & Postpone resolution to a later time. & ``Let's take this to the next meeting.'' \\
\texttt{community\_reference} & LEGIT & Invoke residents, iwi, stakeholders, or ratepayers. & ``Local ratepayers are concerned...'' \\
\texttt{equity\_reference} & LEGIT & Explicit mention of fairness or representation. & ``We need to consider equity impacts.'' \\
\texttt{reassure\_stakeholders} & LEGIT & Statements aimed at public confidence or accountability. & ``We will publish a full report.'' \\
\texttt{public\_address} & LEGIT & Statements intended for the public record. & ``For the benefit of residents...'' \\
\texttt{assign\_action} & ACTION & Delegate concrete tasks to staff or members. & ``Ask staff to prepare the draft.'' \\
\texttt{confirm\_next\_step} & ACTION & Agree on follow-up actions and timelines. & ``We'll meet again in two weeks.'' \\
\texttt{report\_back} & ACTION & Update on previously assigned actions. & ``Staff report: action complete.'' \\
\texttt{close\_discussion} & PROC / ACTION & Chair moves to closure or transitions to next agenda item. & ``Moving on to item 5.'' \\
\bottomrule
\end{tabularx}
\end{table*}

\FloatBarrier

\section{Dataset Validation}
\label{app:data_val}
We validate the dataset through a set of crowdsourced human evaluation tasks designed to assess the reliability of the main annotation components: speaker attribution, cross-video speaker consistency, transcription quality, and action-tag labels. For the diarization and transcription components, workers verify whether utterances are assigned to the correct speakers, whether the same speaker can be consistently identified across videos, and whether the transcript is readable and faithful to the underlying utterance. For action tags, workers label the communicative action expressed by a target utterance given two preceding turns of context, allowing us to compare human judgments against GPT-assigned labels. Across tasks, we collect multiple independent annotations per item and include attention-check questions to improve annotation reliability. Together, these evaluations test whether the dataset’s speaker, transcript, and action-level annotations align with human judgments.

\label{app:dataset_eval}
\begin{figure}[h!]
    \centering
    \begin{subfigure}[b]{0.32\textwidth}
        \centering
        \includegraphics[width=\textwidth]{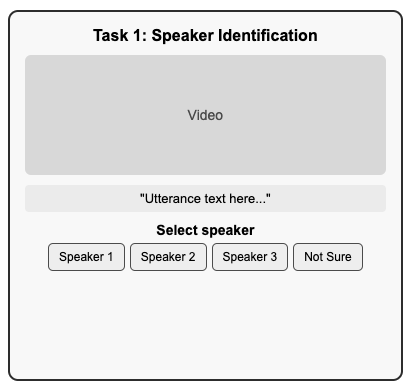}
        \caption{Speaker Identification}
        \label{fig:mturk_1}
    \end{subfigure}
    \hfill
    \begin{subfigure}[b]{0.32\textwidth}
        \centering
        \includegraphics[width=\textwidth]{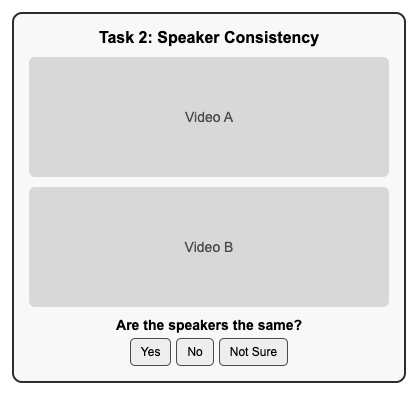}
        \caption{Speaker Consistency}
        \label{fig:mturk_2}
    \end{subfigure}
    \hfill
    \begin{subfigure}[b]{0.32\textwidth}
        \centering
        \includegraphics[width=\textwidth]{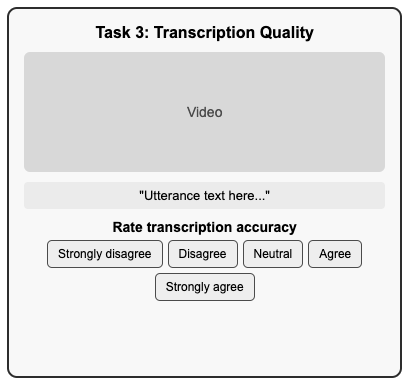}
        \caption{Transcription Quality}
        \label{fig:mturk_3}
    \end{subfigure}
    \caption{
    Overview of the human evaluation tasks conducted on Amazon Mechanical Turk (MTurk). 
    (\subref{fig:mturk_1}): Workers select the correct speaker for a single utterance, producing the Speaker Identification Accuracy. 
    (\subref{fig:mturk_2}): Workers determine whether two utterances from different videos are from the same speaker, yielding Cross-Video Speaker Consistency. 
    (\subref{fig:mturk_3}): Workers rate the transcription of an utterance on a 5-point Likert scale, producing the Transcription Quality Score. 
    Each task included three independent workers per item and attention-check questions to ensure reliability.
    }
    \label{fig:mturk_tasks_all}
    \end{figure}
    \begin{figure}
    \centering
    \begin{subfigure}[b]{0.32\textwidth}
        \centering
        \includegraphics[width=\textwidth]{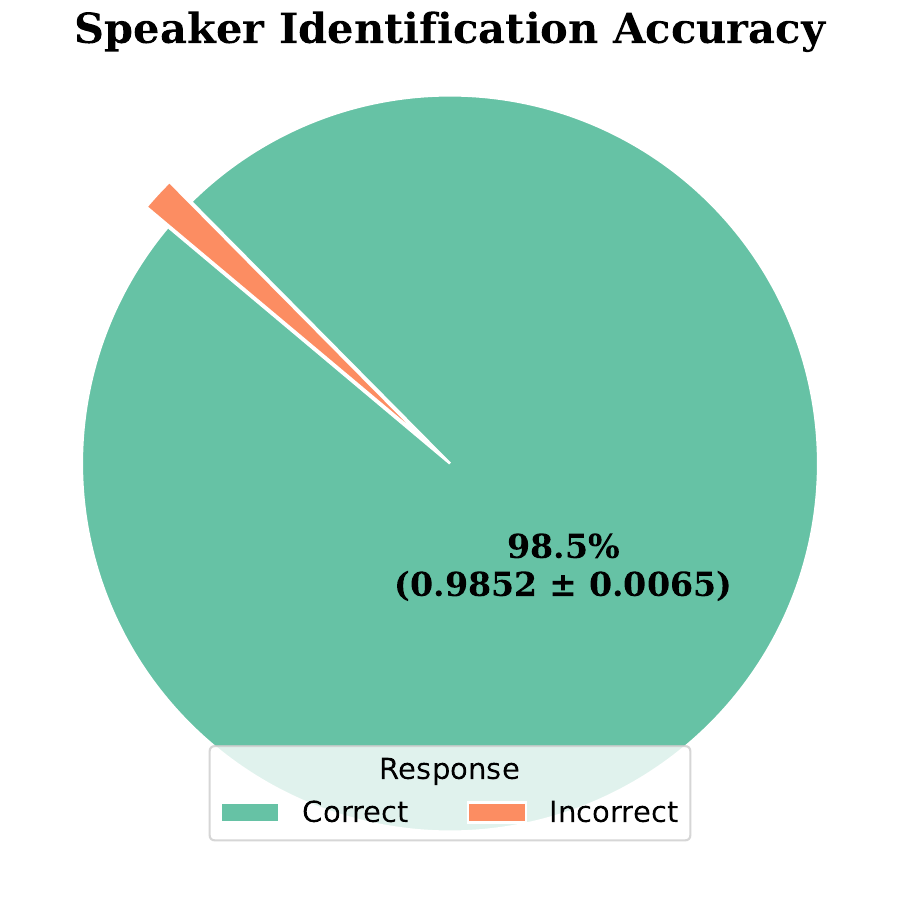}
        \caption{Speaker Identification Accuracy}
        \label{fig:diarization_speaker_identification}
    \end{subfigure}
    \hfill
    \begin{subfigure}[b]{0.32\textwidth}
        \centering
        \includegraphics[width=\textwidth]{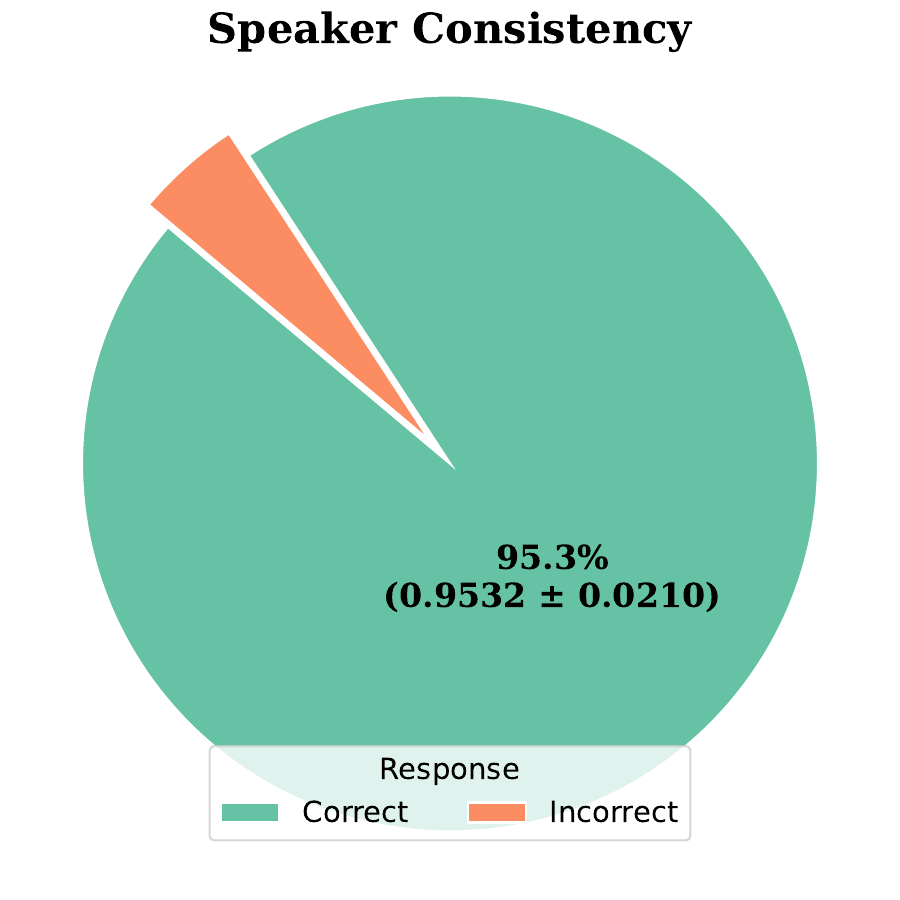}
        \caption{Cross-Video Speaker Consistency
        }
        \label{fig:diarization_speaker_consistency}
    \end{subfigure}
    \hfill
    \begin{subfigure}[b]{0.32\textwidth}
        \centering
        \includegraphics[width=\textwidth]{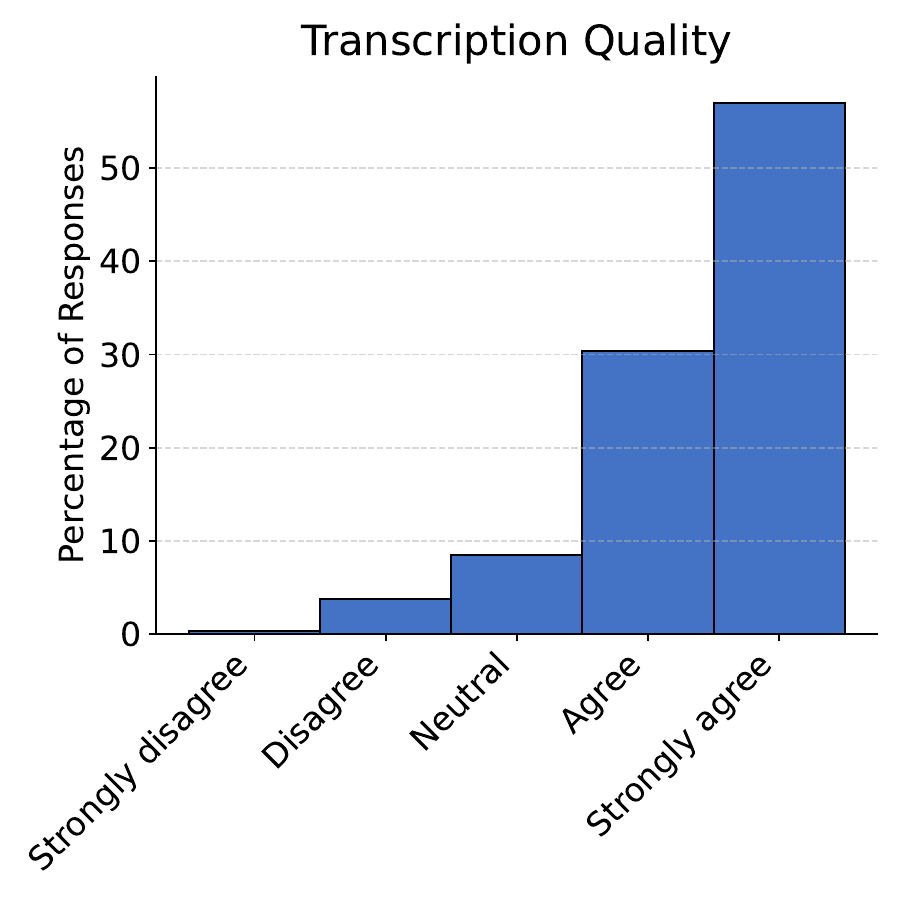}
        \caption{Transcription Quality Score}
        \label{fig:diarization_transcription_quality}
    \end{subfigure}
    \caption{
    Results of the three human evaluation tasks assessing dataset quality. 
    (\subref{fig:diarization_speaker_identification}) Speaker identification accuracy confirms strong alignment between ground-truth and human-labeled speakers.
    (\subref{fig:diarization_speaker_consistency}) Cross-video speaker consistency indicates that speakers can be correctly mapped across different videos.
    (\subref{fig:diarization_transcription_quality}) Transcription quality ratings confirm high perceived fidelity and readability.
    Together, these metrics demonstrate the reliability and human-validated integrity of the dataset.
    }
    \label{fig:diarization_result}
\end{figure}

\paragraph{Transcript Validation.}
Human judgments indicate that the core transcript and speaker annotations are highly reliable. Workers correctly identified the speaker for 98.5\% of utterances and matched speakers across videos with 95.3\% accuracy (Figs.~\ref{fig:diarization_speaker_identification}--\ref{fig:diarization_speaker_consistency}). Transcript ratings were also strongly positive: 55.9\% of ratings were ``Strongly agree'' and 29.8\% were ``Agree'' that the transcript was accurate and readable, with only 4.0\% negative ratings (Fig.~\ref{fig:diarization_transcription_quality}). These results suggest that the dataset provides a reliable basis for downstream speaker- and utterance-level modeling.

\paragraph{Action Tag Validation.}
We also evaluate whether GPT-assigned action tags agree with human judgments using the annotation task shown in Fig.~\ref{fig:action_tag_task}. At the individual-worker level, workers agreed with the GPT tag in 68.0\% of annotations (Fig.~\ref{fig:action_tag_worker_level}). After aggregating worker responses by majority vote, item-level agreement rose to 81.5\% (Fig.~\ref{fig:action_tag_item_level}). Because action tagging requires interpreting the communicative function of an utterance in context, the task involves some inherent judgment and ambiguity. Nonetheless, the results show that human annotators generally agree with the GPT-assigned labels, indicating that our action tags are reliable.

\begin{figure}[h!]
    \centering
    \begin{subfigure}[b]{0.32\textwidth}
        \centering
        \includegraphics[width=\textwidth]{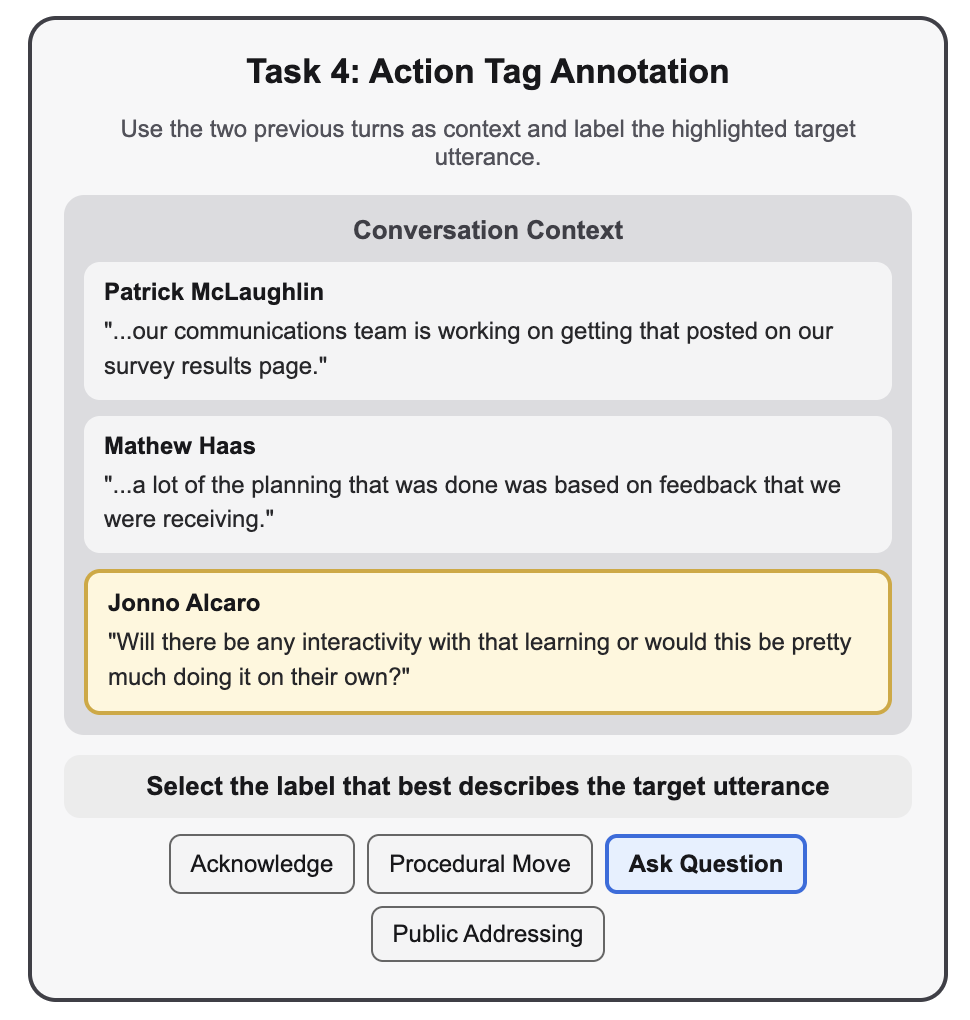}
        \caption{Annotation Task}
        \label{fig:action_tag_task}
    \end{subfigure}
    \hfill
    \begin{subfigure}[b]{0.32\textwidth}
        \centering
        \includegraphics[width=\textwidth]{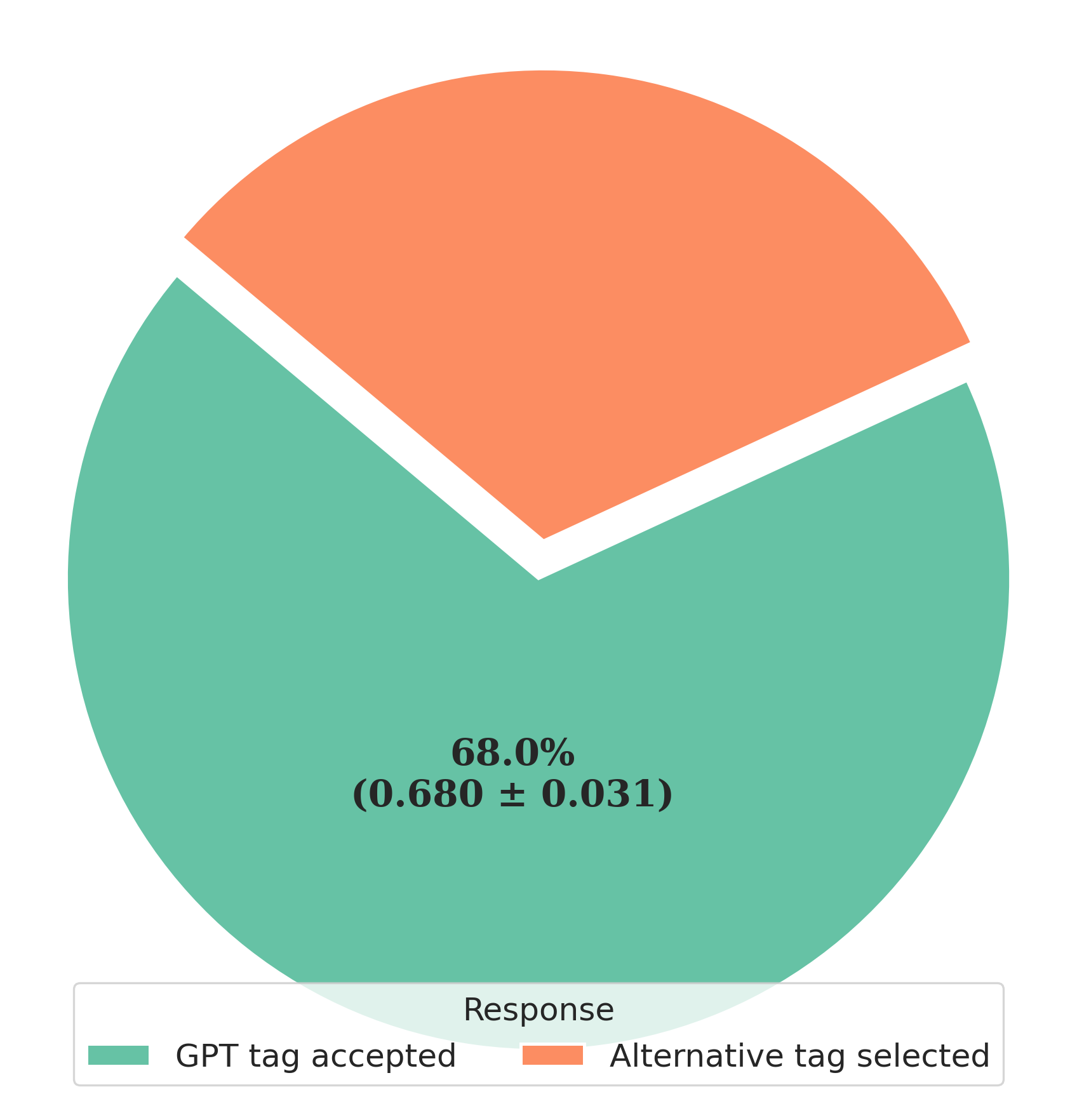}
        \caption{Worker-Level Agreement}
        \label{fig:action_tag_worker_level}
    \end{subfigure}
    \hfill
    \begin{subfigure}[b]{0.32\textwidth}
        \centering
        \includegraphics[width=\textwidth]{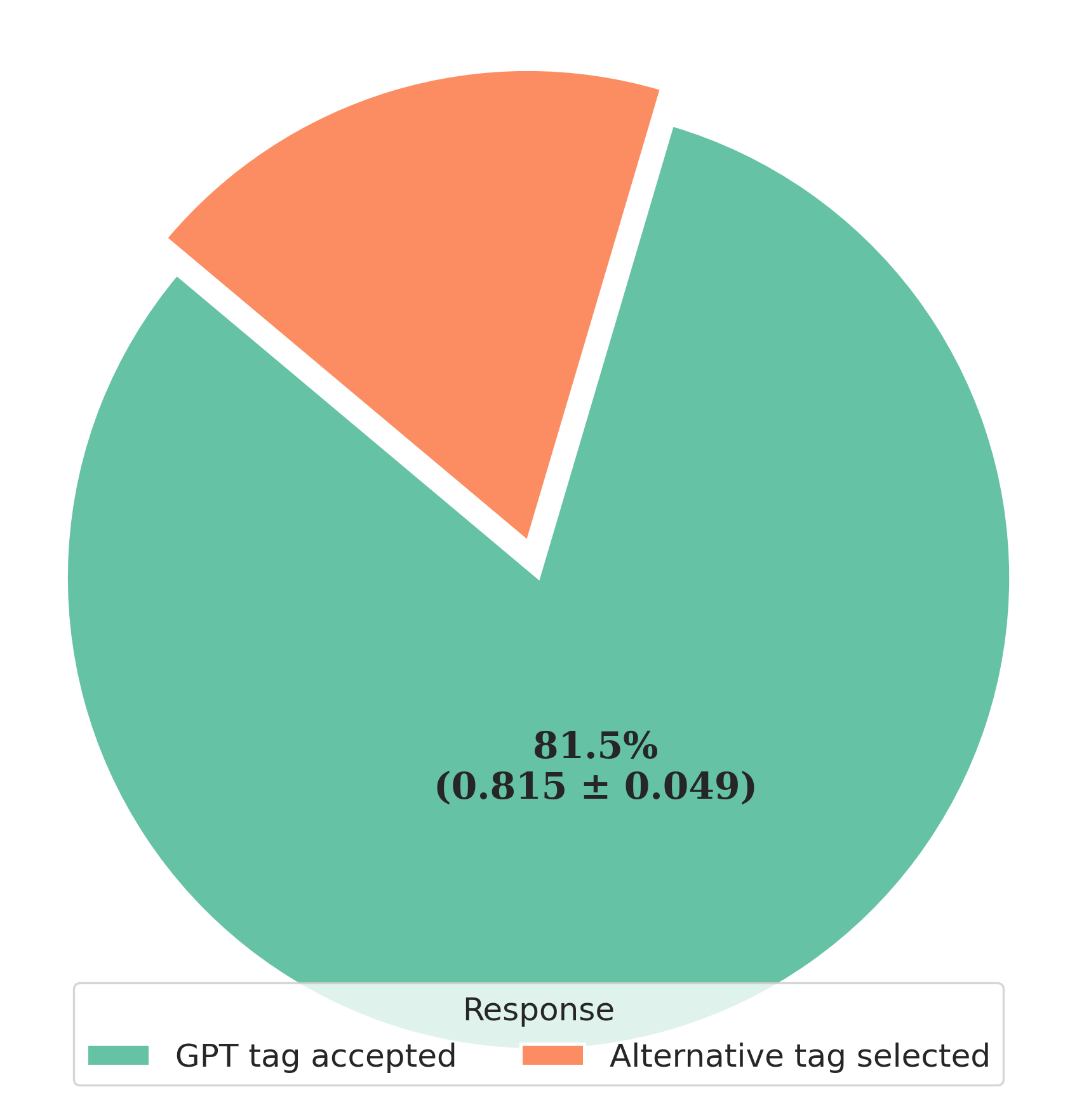}
        \caption{Item-Level Majority Vote}
        \label{fig:action_tag_item_level}
    \end{subfigure}
    \caption{
    Human validation of action-tag annotations.
    (\subref{fig:action_tag_task}) MTurk workers are shown two preceding turns for context and a highlighted target utterance, then select the action label that best describes the target utterance.
    (\subref{fig:action_tag_worker_level}) Worker-level agreement measures the proportion of individual annotations that match the GPT-assigned action tag.
    (\subref{fig:action_tag_item_level}) Item-level agreement measures whether the GPT-assigned action tag matches the majority label selected by workers for each item.
    These results assess whether the automatically assigned action tags align with human judgments.
    }
    \label{fig:action_tag_results}
\end{figure}

\FloatBarrier

\section{Classifier Training for Style and Identity Evaluation}
\label{app:clf_results}

This appendix provides full implementation and evaluation details for the classifiers used to compute the \textbf{Fool Rate} and \textbf{Speaker Attribution Accuracy} metrics described in Section~\ref{sec:evaluation_metrics}. Both classifiers are based on fine-tuned \texttt{DeBERTa-v3-large} models and are designed to assess whether generated utterances exhibit realistic, persona-consistent linguistic style.

\subsection{Fool Rate Classifier (One-vs-All)}
\label{app:foolrate}

The \textit{Fool Rate} classifier evaluates whether a fine-tuned agent can reproduce a target speaker’s stylistic patterns closely enough to “fool” a discriminative model. For each speaker $s$, a separate binary classifier $C_s$ is trained to predict whether an utterance $x$ originates from that speaker (\textit{positive}) or from any other (\textit{negative}). High fool rates indicate that an agent’s generated outputs align strongly with the target persona’s stylistic and lexical characteristics.

Each utterance in the dataset is treated as an independent text sample, with only content-rich utterances (more than ten words) included to ensure sufficient stylistic signal. Data are split 70/30 into training and test sets. All available training utterances from the target speaker serve as positive examples, while an equal number of randomly sampled utterances from other speakers form the negative class.  Each one-vs-all classifier is initialized from \texttt{microsoft/deberta-v3-large}. Training employs the AdamW optimizer with a learning rate of $1\times10^{-6}$, and early stopping is applied based on validation F1 score with a patience of 5.

All classifiers achieve strong discriminative performance, confirming their sensitivity to stylistic features. Detailed per-speaker precision, recall, accuracy, and F1 scores for the Albemarle County School Board, DC Court of Appeals, and Waipā District Council datasets are reported in \autoref{tab:CFR_SAA_METRIC}.

\subsection{Speaker Attribution Accuracy (Multi-Class)}
\label{app:attribution}

The \textit{Speaker Attribution} model assesses whether generated utterances remain uniquely identifiable to their respective personas. In contrast to the \emph{Fool Rate} classifier, which measures how convincingly an agent imitates a single target speaker, this model evaluates the distinctiveness of each persona’s linguistic and stylistic signature across the entire speaker set. High attribution accuracy indicates that persona-specific characteristics are preserved and consistently detectable in a multi-speaker context.

Only utterances longer than twenty words are included to ensure sufficient linguistic context for classification. To address class imbalance, the dataset is upsampled so that each speaker contributes an equal number of examples, followed by a 70/30 train–test split. We fine-tune \texttt{microsoft/deberta-v3-large} using AdamW with a learning rate of $1\times10^{-6}$ and apply early stopping with a patience of 5, monitored on macro-averaged precision.

The multi-class classifier achieves strong discriminative performance, confirming that persona-specific linguistic features remain recoverable across speakers. Detailed per-speaker precision, recall, accuracy, and F1 scores for the Albemarle County School Board, DC Court of Appeals, and Waipā District Council datasets datasets are presented in \autoref{tab:CFR_SAA_METRIC}.

\begin{table*}[t]
\centering
\begin{tabular}{c}

\begin{subtable}[t]{\textwidth}
\centering
\begin{tabular}{lcccc}
\toprule
\textbf{Speaker} & \textbf{Precision} & \textbf{Recall} & \textbf{Accuracy} & \textbf{F1 Score} \\
\midrule
Ellen Osborne & $0.7778$ ($0.3333$) & $0.8235$ ($0.2500$) & $0.8000$ ($0.2500$) & $0.8000$ ($0.2857$) \\
David Oberg & $0.7791$ ($0.5833$) & $0.9306$ ($0.6034$) & $0.8333$ ($0.6034$) & $0.8481$ ($0.5932$) \\
Graham Paige & $0.9174$ ($0.9000$) & $0.8706$ ($0.7105$) & $0.8963$ ($0.7105$) & $0.8934$ ($0.7941$) \\
Jonno Alcaro & $0.8182$ ($0.5210$) & $0.7920$ ($0.6139$) & $0.8088$ ($0.6139$) & $0.8049$ ($0.5636$) \\
Katrina Callsen & $0.8732$ ($0.8000$) & $0.8105$ ($0.6621$) & $0.8464$ ($0.6621$) & $0.8407$ ($0.7157$) \\
Kate Acuff & $0.7395$ ($0.4250$) & $0.8980$ ($0.8095$) & $0.7919$ ($0.8095$) & $0.8111$ ($0.5574$) \\
Judy Le & $0.7011$ ($0.6000$) & $0.9531$ ($0.4821$) & $0.7734$ ($0.4821$) & $0.8079$ ($0.5347$) \\
\bottomrule
\end{tabular}
\caption{Albemarle dataset: One-vs-all classifiers (multi-class scores in parentheses).}
\end{subtable} \\[8em]

\begin{subtable}[t]{\textwidth}
\centering
\begin{tabular}{lcccc}
\toprule
\textbf{Speaker} & \textbf{Precision} & \textbf{Recall} & \textbf{Accuracy} & \textbf{F1 Score} \\
\midrule
Judge McLeese & $0.8421$ ($0.6512$) & $0.7059$ ($0.4828$) & $0.7868$ ($0.4828$) & $0.7680$ ($0.5545$) \\
Judge Glickman & $0.7297$ ($0.1818$) & $0.9000$ ($0.1250$) & $0.7869$ ($0.1250$) & $0.8060$ ($0.1481$) \\
Judge Deahl & $0.6275$ ($0.5385$) & $0.9412$ ($0.4828$) & $0.6957$ ($0.4828$) & $0.7529$ ($0.5091$) \\
Judge Easterly & $0.7727$ ($0.7524$) & $0.9444$ ($0.8158$) & $0.8378$ ($0.8158$) & $0.8500$ ($0.7828$) \\
Judge Blackburn & $0.6857$ ($0.8333$) & $0.9231$ ($0.6207$) & $0.7547$ ($0.6207$) & $0.7869$ ($0.7115$) \\
Judge Thompson & $0.6829$ ($0.8000$) & $0.9032$ ($0.6621$) & $0.7419$ ($0.6621$) & $0.7778$ ($0.7245$) \\
\bottomrule
\end{tabular}
\caption{DCAppeals dataset: One-vs-all classifiers (multi-class scores in parentheses).}
\end{subtable} \\[8em]

\begin{subtable}[t]{\textwidth}
\centering
\begin{tabular}{lcccc}
\toprule
\textbf{Speaker} & \textbf{Precision} & \textbf{Recall} & \textbf{Accuracy} & \textbf{F1 Score} \\
\midrule
Susan O'regan & $0.6623$ ($0.5024$) & $0.9107$ ($0.7222$) & $0.7240$ ($0.7222$) & $0.7669$ ($0.5926$) \\
Jim Mylchreest & $0.9058$ ($0.6788$) & $0.8737$ ($0.7529$) & $0.8914$ ($0.7529$) & $0.8895$ ($0.7139$) \\
Claire St Pierre & $0.7938$ ($0.5882$) & $0.9549$ ($0.7031$) & $0.8534$ ($0.7031$) & $0.8669$ ($0.6406$) \\
Andrew Brown & $0.7083$ ($0.3478$) & $0.6800$ ($0.2963$) & $0.7059$ ($0.2963$) & $0.6939$ ($0.3200$) \\
Roger Gordon & $0.7826$ ($0.6667$) & $0.8780$ ($0.3902$) & $0.8193$ ($0.3902$) & $0.8276$ ($0.4923$) \\
Lou Brown & $0.7627$ ($0.5507$) & $0.7258$ ($0.7170$) & $0.7520$ ($0.7170$) & $0.7438$ ($0.6230$) \\
Ange Holt & $0.8525$ ($0.7667$) & $0.8595$ ($0.6216$) & $0.8560$ ($0.6216$) & $0.8560$ ($0.6866$) \\
Sue Milner & $0.7500$ ($0.7091$) & $0.9065$ ($0.3750$) & $0.8022$ ($0.3750$) & $0.8208$ ($0.4906$) \\
\bottomrule
\end{tabular}
\caption{Waipā dataset: One-vs-all classifiers (multi-class scores in parentheses).}
\end{subtable}

\end{tabular}
\caption{Accuracy Metrics for Classifiers used for Fool Rates and Speaker Attribution Accuracy.}
\label{tab:CFR_SAA_METRIC}
\end{table*}

\FloatBarrier  
\clearpage

\section{Simulation Agendas}
\label{app:sim_topics}

\begin{figure}[h!]
\footnotesize
\centering

\begin{tcolorbox}[colback=gray!3,colframe=black!15,
title=\textbf{DCAppeals Dataset},
width=\linewidth]

\begin{itemize}\setlength\itemsep{2pt}

\item \textbf{Opening Statement by Appellant}
\begin{itemize}\setlength\itemsep{1pt}
\item Appellant introduces the case background and identifies the key legal question on appeal.
\item Explain that the central issue concerns whether Mr. General was ``in custody'' for Miranda purposes during a 1 hour 27 minute station-house interview.
\end{itemize}

\item \textbf{Custody Determination and Miranda Framework}
\begin{itemize}\setlength\itemsep{1pt}
\item Explore the factual circumstances of the interrogation, including duration, setting, tone, and restrictions on movement.
\item Analyze whether a reasonable person in Mr. General’s position would have felt free to leave.
\end{itemize}

\item \textbf{Harmless Error and Impact on the Verdict}
\begin{itemize}\setlength\itemsep{1pt}
\item Discuss whether any statements obtained during the interview were introduced at trial and how they influenced the jury’s perception of guilt or intent.
\item Evaluate whether the government has met its burden to prove the error was harmless beyond a reasonable doubt.
\end{itemize}

\item \textbf{Preservation, Waiver, and Standard of Review}
\begin{itemize}\setlength\itemsep{1pt}
\item Clarify whether the Miranda custody issue was raised at the trial level or preserved for appeal.
\item Discuss the implications of waiver, forfeiture, and plain-error review.
\end{itemize}

\item \textbf{Closing Argument and Relief Requested}
\begin{itemize}\setlength\itemsep{1pt}
\item Summarize the appellant’s key legal points regarding custody, Miranda applicability, and prejudicial error.
\item Reiterate that Mr. General’s statements should have been suppressed and the conviction reversed.
\item \textbf{Decision:} Judges vote on granting relief.
\end{itemize}

\end{itemize}
\end{tcolorbox}

\caption{Simulation agenda for the DCAppeals dataset.}
\label{fig:sim_agenda_dcappeals}
\end{figure}

\begin{figure}[p]
\footnotesize
\centering

\begin{tcolorbox}[colback=gray!3,colframe=black!15,
title=\textbf{Albemarle Dataset},
width=\linewidth]

\begin{itemize}\setlength\itemsep{2pt}

\item \textbf{Health \& Safety: COVID-19 Response and Elementary Return Plan}
\begin{itemize}\setlength\itemsep{1pt}
\item Assess the adequacy of current safety measures (masks, distancing, breaks) and readiness of facilities and staff.
\item Balance family safety concerns with the goal of returning PreK--3 students to in-person or hybrid learning.
\item \textbf{Decision:} Board Vote on mask/safety guidelines and PreK--3 return.
\end{itemize}

\item \textbf{Technology and AI Use in Classrooms}
\begin{itemize}\setlength\itemsep{1pt}
\item Explore ways technology and AI can enhance learning while mitigating risks to critical thinking and engagement.
\item Decide on district guidelines, limits, and staff training for responsible AI and technology use.
\item \textbf{Decision:} Board Vote on AI/technology policy.
\end{itemize}

\item \textbf{High School Center 2: Design, Budget, and Sustainability Goals}
\begin{itemize}\setlength\itemsep{1pt}
\item Review the design’s alignment with collaborative learning and community values while considering sustainability.
\item Confirm budget, timeline, and long-term vision alignment before advancing the project.
\item \textbf{Decision:} Board Vote on schematic advancement.
\end{itemize}

\item \textbf{Meeting Wrap-Up \& Dismissal}
\begin{itemize}\setlength\itemsep{1pt}
\item Summarize key decisions and clarify any remaining questions.
\item Confirm next meeting date and officially adjourn.
\item \textbf{Decision:} Board Vote to adjourn.
\end{itemize}

\end{itemize}
\end{tcolorbox}

\caption{Simulation agenda for the Albemarle dataset.}
\label{fig:sim_agenda_albemarle}
\end{figure}

\begin{figure}[p]
\footnotesize
\centering

\begin{tcolorbox}[colback=gray!3,colframe=black!15,
title=\textbf{Waipā Dataset},
width=\linewidth]

\begin{itemize}\setlength\itemsep{2pt}

\item \textbf{Insurance Market, Placement and Strategy}
\begin{itemize}\setlength\itemsep{1pt}
\item Transition from Marsh to Aon and renewal of liability/indemnity policies, with market update on global insurance conditions.
\item Shift to a risk-based insurance strategy including asset valuations, loss modelling, and emerging cyber coverage concerns.
\item \textbf{Decision:} Committee Vote on risk-based strategy and Aon placement.
\end{itemize}

\item \textbf{Top Risks, Risk Appetite and Oversight}
\begin{itemize}\setlength\itemsep{1pt}
\item Endorse updated top risks and risk appetite statement, including new fraud/bribery/corruption portfolio, with review of emerging risk themes.
\item Discuss heat map, mitigation measures, and proposal for climate-change workshop with ongoing monitoring.
\item \textbf{Decision:} Committee Vote on risk register and workshop.
\end{itemize}

\item \textbf{Funding, Capital Programme and Asset Management}
\begin{itemize}\setlength\itemsep{1pt}
\item Update on Waka Kotahi funding decisions, council discretion on reallocation, and review of deferred projects due to Three Waters reform.
\item Endorse activity management framework and align asset valuations/probable-maximum-loss modelling with insurance limits.
\item \textbf{Decision:} Committee Vote on asset framework.
\end{itemize}

\item \textbf{Meeting Wrap-Up \& Dismissal}
\begin{itemize}\setlength\itemsep{1pt}
\item Summarize key decisions and clarify any remaining questions.
\item Confirm next meeting date and officially adjourn.
\item \textbf{Decision:} Committee Vote to adjourn.
\end{itemize}

\end{itemize}
\end{tcolorbox}

\caption{Simulation agenda for the Waipā dataset.}
\label{fig:sim_agenda_waipa}
\end{figure}
\FloatBarrier

\section{System Prompts}
\label{app:system_prompts}

\begin{figure}[h!]
\footnotesize
\centering

\begin{tcolorbox}[colback=gray!3,colframe=black!15,
title=\textbf{Persona Prompt},
width=\linewidth]

\begin{itemize}\setlength\itemsep{2pt}

\item \textbf{Persona Description}
\begin{itemize}\setlength\itemsep{1pt}
\item Graham Paige, chair of the Albemarle County School Board, is a thoughtful and respectful advocate for equity in education.
\item He values inclusive decision-making, student representation, and appreciation for teachers and community members.
\item His tone is formal yet warm, emphasizing gratitude and collaboration.
\end{itemize}

\item \textbf{In-Context Examples}
\begin{itemize}\setlength\itemsep{1pt}
\item \textit{jonnoalcaro:} Graham, could you share your thoughts on teacher morale?
\item \textit{grahampaige:} Certainly. Our teachers have shown remarkable resilience, and I deeply appreciate their dedication to students' success during challenging times.
\end{itemize}

\item \textbf{Micro-Profile}
\begin{itemize}\setlength\itemsep{1pt}
\item Avg.\ response length $\approx$ 45 words
\item Question frequency: $\sim$60\%
\item Politeness rate: $\sim$45\%
\item Sentiment: mildly positive ($\approx$0.5)
\item \textbf{Behavioral guidance:} Respond politely, often ask questions, and maintain a warm, professional tone.
\end{itemize}

\item \textbf{Conversation Context}
\begin{itemize}\setlength\itemsep{1pt}
\item Topics: teacher morale, collective bargaining, meeting procedures.
\item Participants: jonnoalcaro, grahampaige, ellenosborne, katrinacallsen.
\end{itemize}

\item \textbf{Instruction}
\begin{itemize}\setlength\itemsep{1pt}
\item You are \textbf{grahampaige} in a school board meeting.
\item Speak naturally and concisely, using everyday language and light hesitations.
\item Ask questions only to listed participants and follow the behavioral frequencies above.
\end{itemize}

\end{itemize}
\end{tcolorbox}

\caption{Example of the structured persona prompt used to condition speaker-specific conversational agents.}
\label{fig:persona-prompt}
\end{figure}

\begin{figure}[p]
\footnotesize
\centering

\begin{tcolorbox}[colback=gray!3,colframe=black!15,
title=\textbf{Time-Aware vs.\ Time-Unaware Simulation Prompt},
width=\linewidth]

\begin{itemize}\setlength\itemsep{2pt}

\item \textbf{Agenda}
\begin{itemize}\setlength\itemsep{1pt}
\item \textcolor{red}{09:00--09:30:} Budget Review
\item \textcolor{red}{09:30--09:45:} Vote on Budget Proposal
\end{itemize}

\item \textbf{Rules}
\begin{itemize}\setlength\itemsep{1pt}
\item Speak only as your assigned persona, addressing listed participants.
\item Follow the agenda in order and stay on topic.
\item Keep utterances short (1--2 sentences) and natural.
\item \textcolor{red}{Before speaking, always state the current time and agenda item: [current\_time=HH:MM, agenda\_item=...]}
\end{itemize}

\item \textbf{Instruction}
\begin{itemize}\setlength\itemsep{1pt}
\item You are the assigned meeting participant.
\item Speak naturally and briefly, referencing the current agenda item.
\end{itemize}

\item \textbf{Example Output}
\begin{itemize}\setlength\itemsep{1pt}
\item \textit{\underline{Time-Aware:}} \textcolor{red}{[current\_time=09:05, agenda\_item=Budget Review]} \textit{alice:} Uh, so, let's start with the budget numbers.
\item \textit{\underline{Time-Unaware:}} \textit{alice:} Let's start with the budget numbers.
\end{itemize}

\item \textbf{Key Difference}
\begin{itemize}\setlength\itemsep{1pt}
\item The time-aware setup adds explicit temporal structure through agenda timestamps and per-turn time/agenda annotations.
\item These cues help produce more chronologically grounded and agenda-consistent dialogue.
\end{itemize}

\end{itemize}
\end{tcolorbox}

\caption{Time-aware and time-unaware simulation prompt variants. In the time-aware setting, agenda items include explicit timestamps (shown in red), and agents are instructed to state the current time and agenda item before each utterance. The time-unaware setting omits these temporal cues, leaving agents to follow the agenda without explicit per-turn time grounding.}
\label{fig:time-aware-simulation}
\end{figure}

\FloatBarrier

\section{Validation of GPT-Based Metrics}
\label{app:metric_validation}
To evaluate institutional fidelity and behavioral coherence at scale, we use GPT-based metrics for agenda item coverage, vote attempt detection, stance-vote consistency, and deliberative responsiveness. Because these metrics require interpreting meeting context rather than matching simple surface patterns, we validate the GPT judgments against human annotations collected through MTurk. Each validation task presents crowdworkers with a short excerpt from either a real or simulated meeting and asks them to make the same type of judgment used by the automated evaluator.

For each task, we annotate 50 examples with three independent MTurk workers per example. Each task includes an ``Unclear / not sure'' option, allowing workers to mark genuinely ambiguous examples rather than forcing a binary decision. We report agreement between the GPT-based metric and human annotations at two levels. First, we compute a worker-level match rate, treating each individual worker judgment as a separate comparison. Second, we compute a majority-vote match rate, aggregating the three worker labels for each example and comparing the GPT judgment against the resulting human consensus label.

Before computing agreement, we drop worker annotations labeled \textit{Unclear / not sure}. For the majority-vote analysis, we also drop examples for which the retained worker labels do not produce a unique majority. This filtering reduces label noise by excluding cases where the human judgment is ambiguous or where workers do not converge on a single interpretation.  Throughout this section, match rate refers to the fraction of human judgments for which the GPT-based metric agrees with the corresponding worker or majority-vote label.
\newpage
\subsection{Deliberative Responsiveness}
\label{app:dr}

Deliberative responsiveness measures whether simulated speakers respond appropriately to questions or requests made in the conversation. This metric tests a local but important form of behavioral coherence: agents should not merely produce plausible standalone utterances, but should react coherently to the immediately preceding conversational context.

For the human validation task, workers are shown two turns: a question turn and the next response turn. They are asked whether the response addresses the question. The answer choices are \textit{Yes}, \textit{No}, and \textit{Unclear / not sure}. A ``Yes'' label indicates that the response directly answers, acknowledges, or otherwise substantively addresses the question. A ``No'' label indicates that the response ignores the question, changes topic, or is otherwise disconnected. The ``Unclear / not sure'' option is used when the exchange is too ambiguous to classify confidently.

As shown in \autoref{fig:turk_deliberative_responsiveness}, GPT-5-based labels for deliberative responsiveness judgments align well with human annotations. At the worker level, the GPT metric matches individual MTurk labels with a match rate of $0.773 \pm 0.033$. After aggregating workers by majority vote, the match rate increases to $0.800 \pm 0.059$. This suggests that the GPT-based responsiveness metric generally captures the same local question-answer relationship identified by human annotators, while majority voting modestly reduces annotation noise.

\begin{figure}[h!]
    \centering
    \begin{subfigure}[t]{0.95\linewidth}
        \centering
        \includegraphics[width=\linewidth]{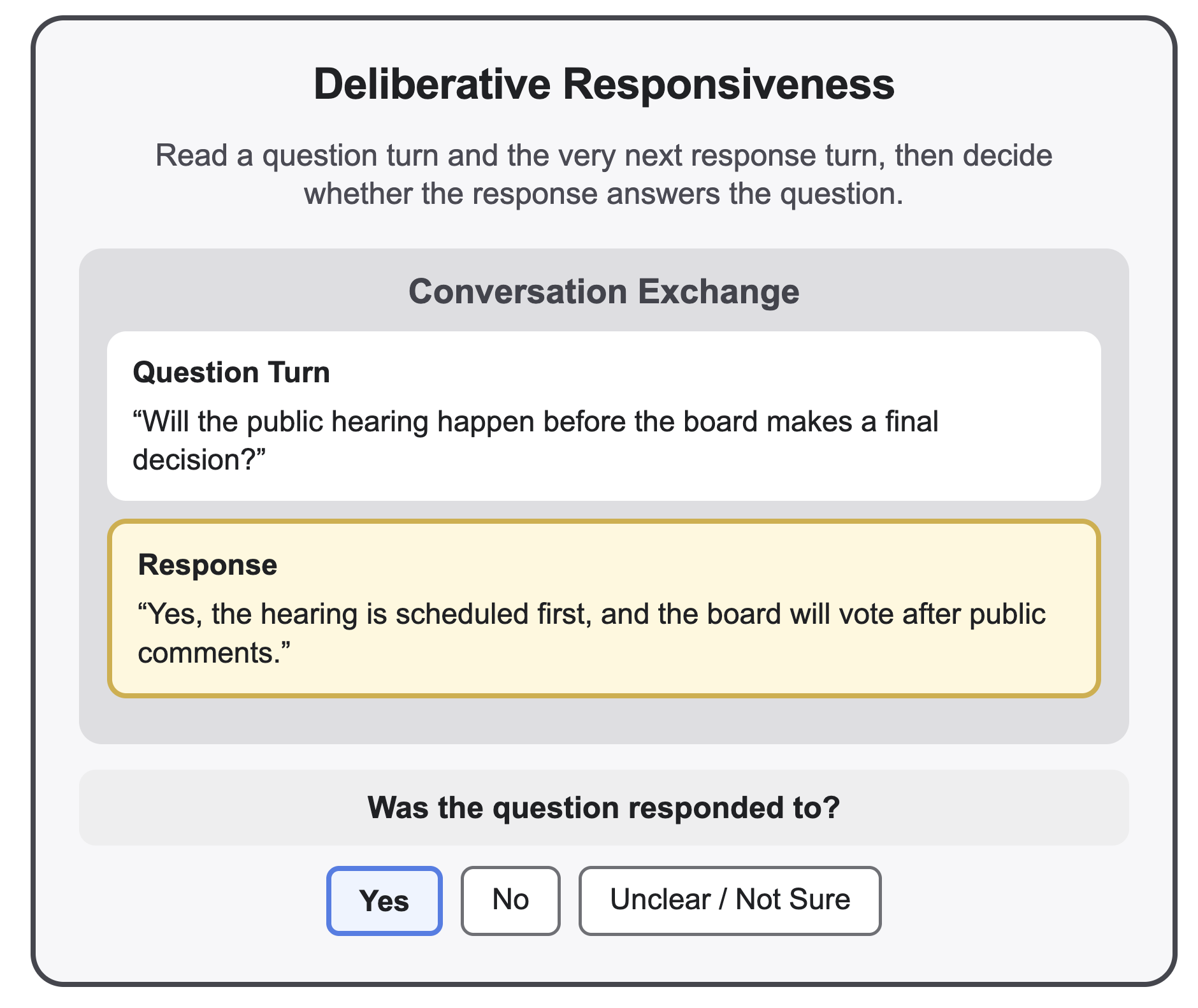}
        \caption{Task interface.}
        \label{fig:turk_deliberative_responsiveness_ui}
    \end{subfigure}

    \vspace{0.75em}

    \begin{subfigure}[t]{0.95\linewidth}
        \centering
        \includegraphics[width=\linewidth]{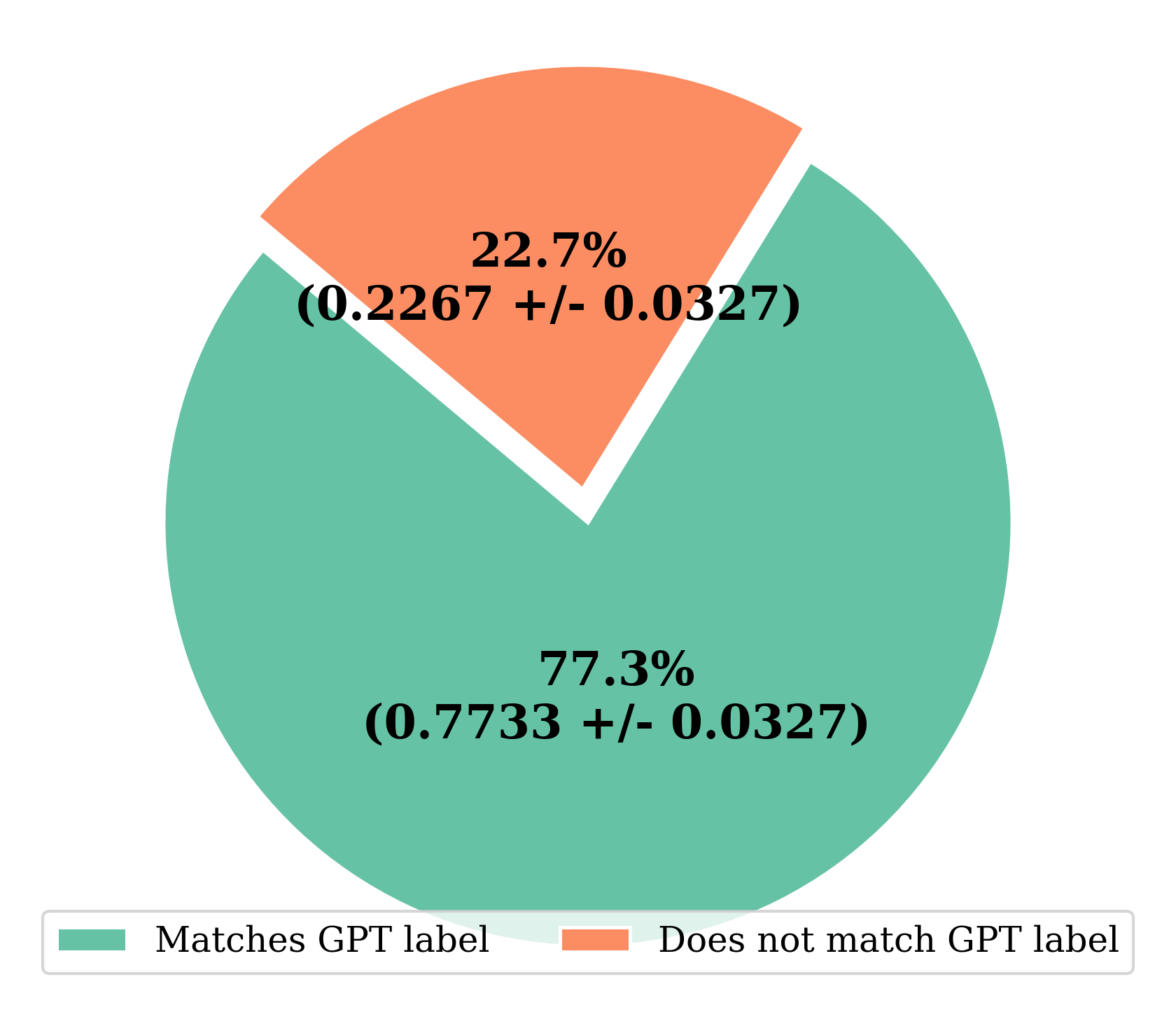}
        \caption{Worker-level match rate.}
        \label{fig:turk_deliberative_responsiveness_wl}
    \end{subfigure}

    \vspace{0.75em}

    \begin{subfigure}[t]{0.95\linewidth}
        \centering
        \includegraphics[width=\linewidth]{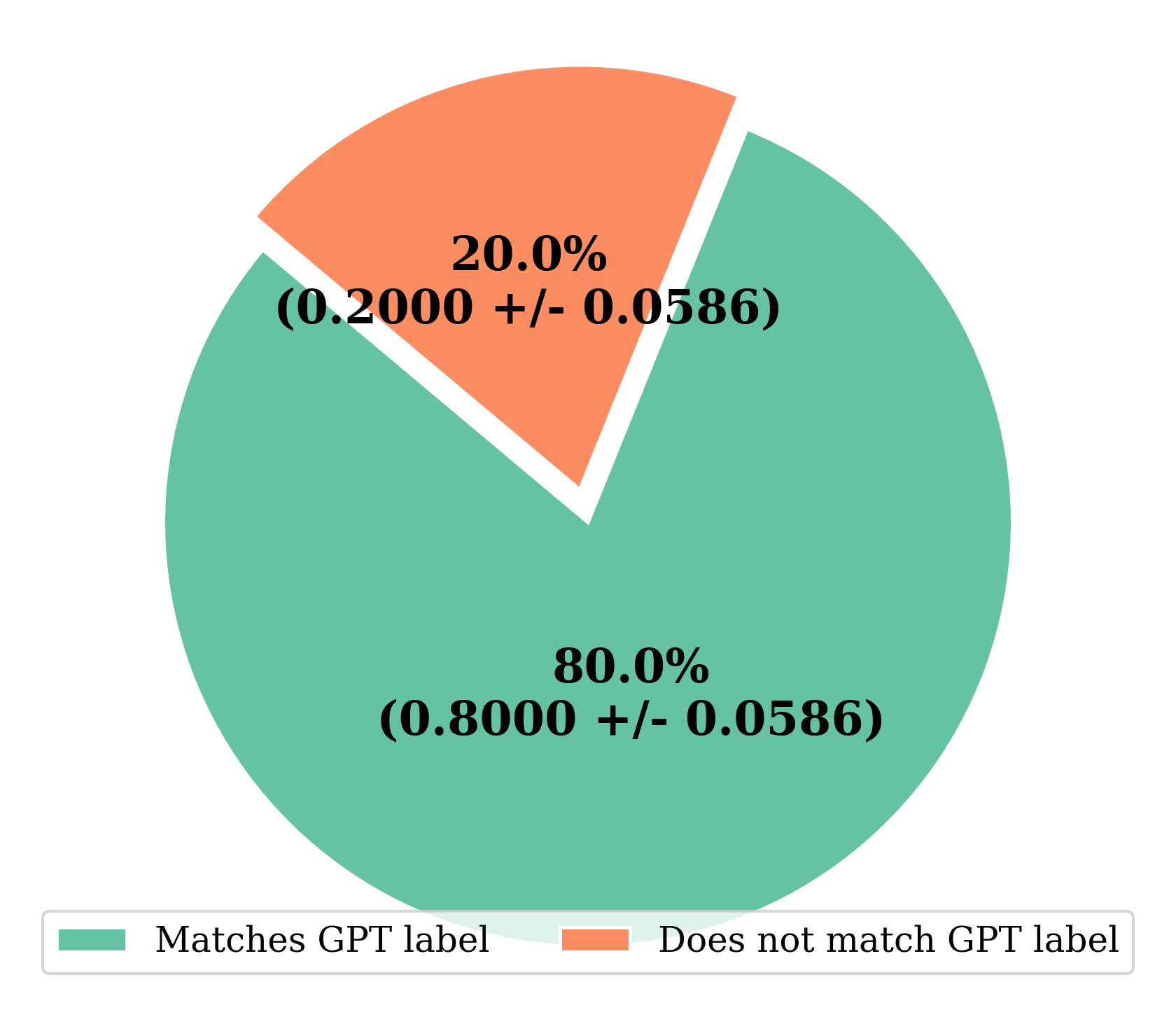}
        \caption{Majority-vote match rate.}
        \label{fig:turk_deliberative_responsiveness_mv}
    \end{subfigure}

    \caption{
    MTurk validation for deliberative responsiveness. Workers are shown a question turn and the following response turn, then judge whether the response addresses the question.
    }
    \label{fig:turk_deliberative_responsiveness}
\end{figure}

\FloatBarrier
\subsection{Stance-Vote Consistency}
\label{app:svc}

Stance-vote consistency evaluates whether a speaker's deliberative statements are consistent with the vote they later cast. This metric captures a central aspect of behavioral coherence: simulated agents should behave as coherent decision-makers whose expressed positions during discussion align with their subsequent procedural actions.

For the human validation task, workers are shown the vote under review, the recorded vote, and up to three earlier turns from the same speaker. They are then asked whether the speaker's previous statements are consistent with the vote. The answer choices are \textit{Yes}, \textit{No}, and \textit{Unclear / not sure}. A ``Yes'' label indicates that the earlier statements support or are compatible with the recorded vote. A ``No'' label indicates that the earlier statements conflict with the recorded vote. The ``Unclear / not sure'' option is used when the prior turns do not provide enough evidence to determine the speaker's stance.

As shown in \autoref{fig:turk_stance_vote} the GPT metric matches individual worker labels with a match rate of $0.728 \pm 0.041$. Under majority-vote aggregation, the match rate is $0.714 \pm 0.062$. The lower agreement relative to the other tasks is expected: determining whether a prior statement supports or conflicts with a later vote often requires interpreting implicit stance, partial evidence, or procedural context. Nevertheless, the GPT-based metric shows substantial alignment with human judgments on this more ambiguous behavioral property.

\begin{figure}[h!]
    \centering
    \begin{subfigure}[h!]{0.9\linewidth}
        \centering
        \includegraphics[width=\linewidth]{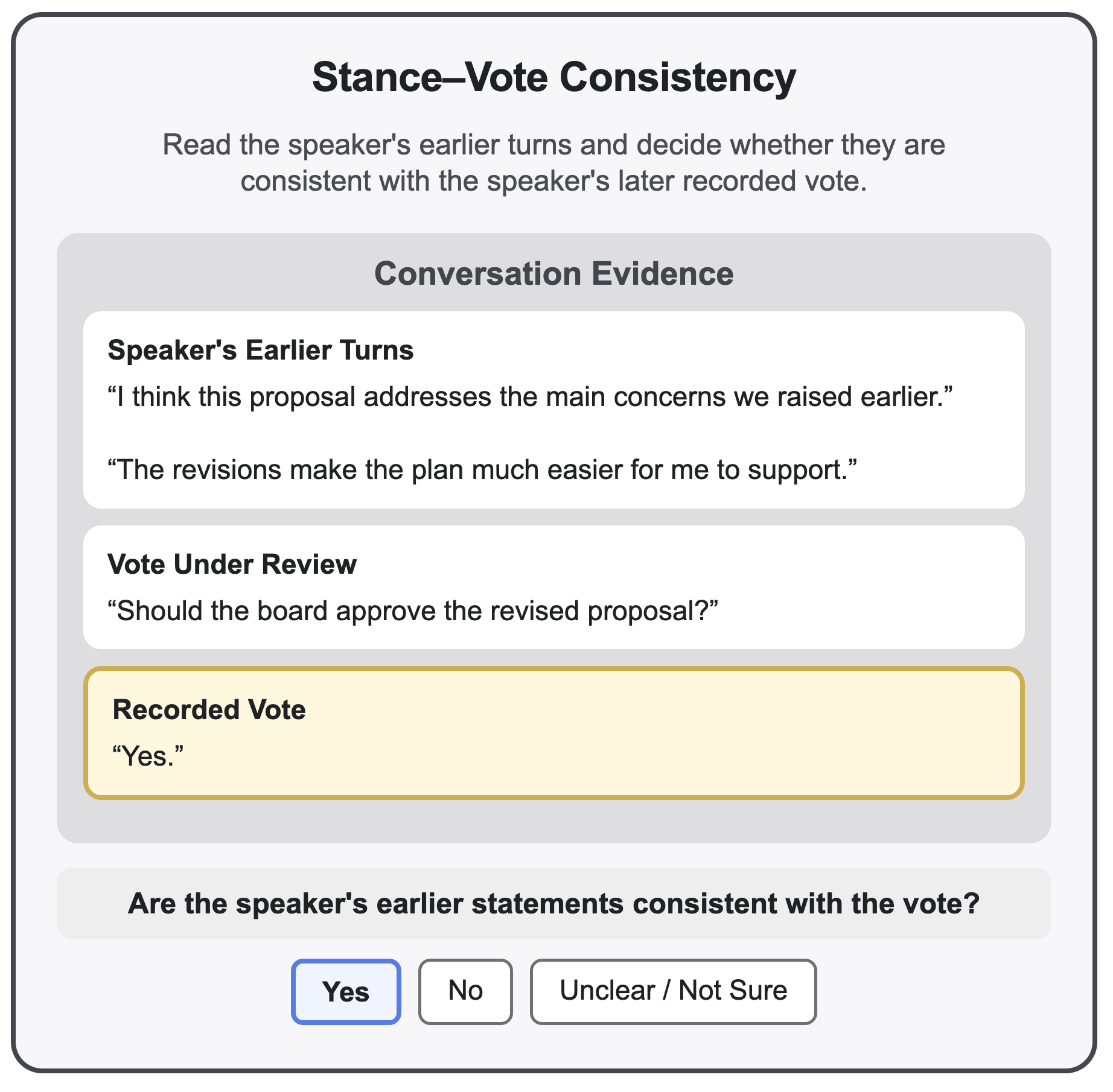}
        \caption{Task interface.}
        \label{fig:turk_stance_vote_ui}
    \end{subfigure}

    \vspace{0.25em}

    \begin{subfigure}[h!]{0.9\linewidth}
        \centering
        \includegraphics[width=\linewidth]{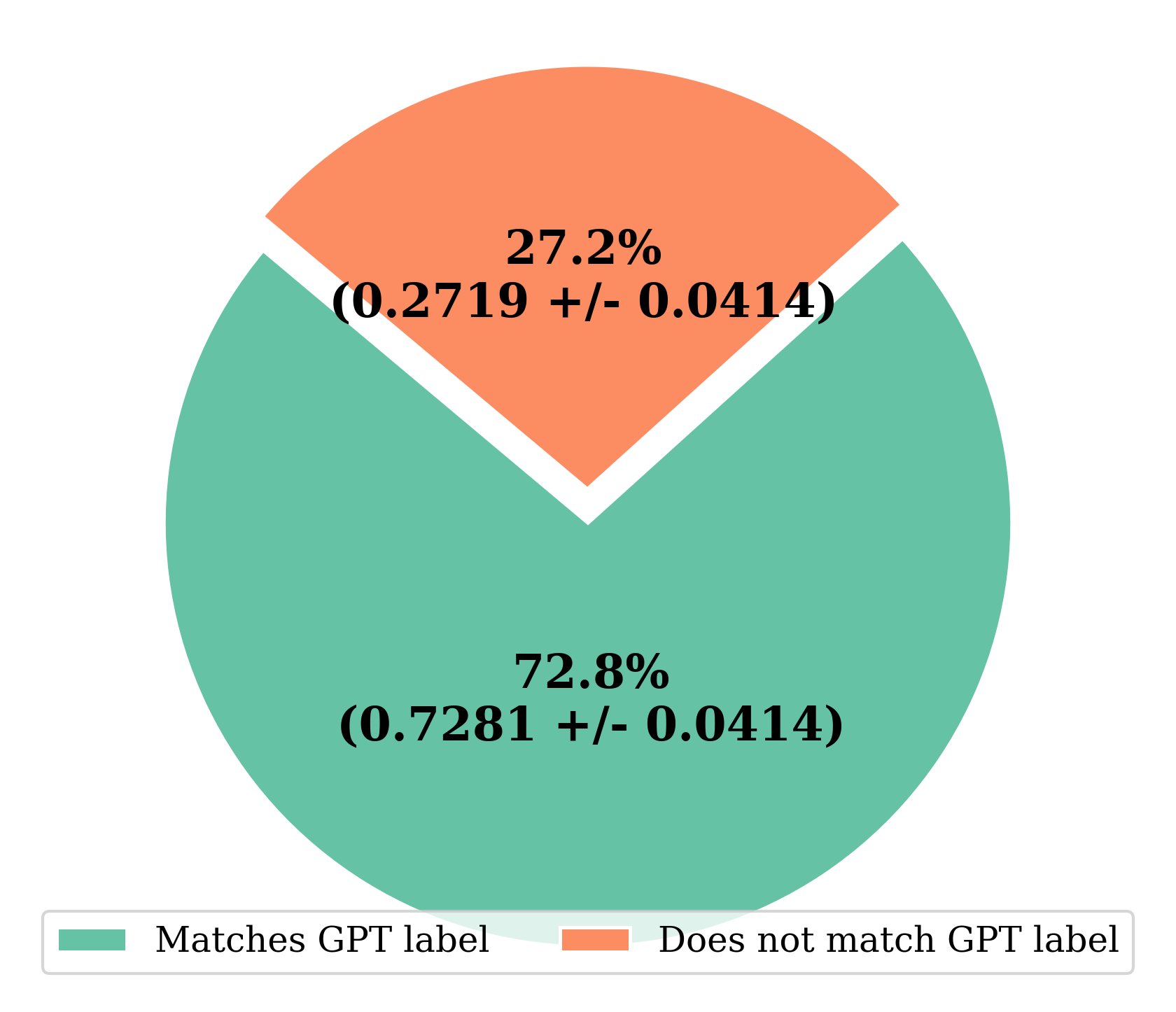}
        \caption{Worker-level match rate.}
        \label{fig:turk_stance_vote_wl}
    \end{subfigure}

    \vspace{0.25em}

    \begin{subfigure}[h!]{0.9\linewidth}
        \centering
        \includegraphics[width=\linewidth]{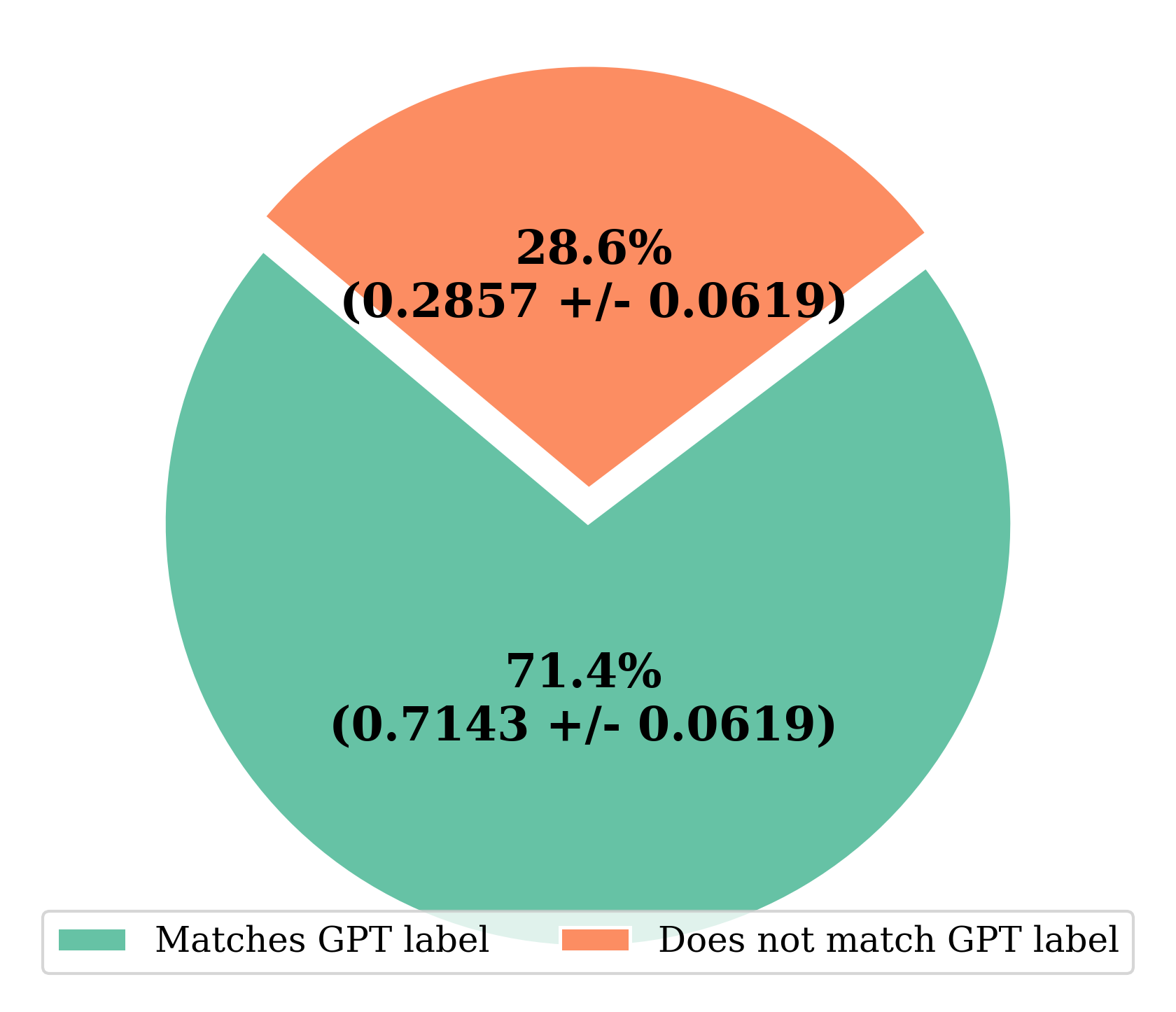}
        \caption{Majority-vote match rate.}
        \label{fig:turk_stance_vote_mv}
    \end{subfigure}

    \caption{
    MTurk validation for stance-vote consistency. Workers are shown a speaker's earlier turns, the later decision point, and the recorded vote, then judge whether the earlier stance is consistent with the vote. 
    }
    \label{fig:turk_stance_vote}
\end{figure}

\FloatBarrier
\subsection{Topic Coverage}
\label{app:tc}

Topic coverage validates the agenda item coverage metric used to assess institutional fidelity. Agenda item coverage asks whether the simulation meaningfully discusses each target agenda item from the original meeting. This tests whether a simulation follows the substantive structure of the meeting, rather than producing generic civic dialogue that appears plausible but omits required agenda content.

For the human validation task, workers are shown a short conversation excerpt and asked to choose which predefined agenda item is being discussed. The choices consist of candidate agenda items, along with an \textit{Unclear / not sure} option. A correct topic match indicates that the excerpt contains enough substantive content for a human reader to identify the agenda item being addressed. This provides a validation signal for whether GPT-based agenda coverage judgments correspond to human-recognizable topic discussion.

As shown in \autoref{fig:turk_coverage}, GPT-based topic coverage judgments align strongly with human topic annotations. At the worker level, the GPT metric matches individual MTurk labels with a match rate of $0.853 \pm 0.027$. After majority-vote aggregation, the match rate increases to $0.920 \pm 0.038$. This increase suggests that many apparent disagreements at the individual-worker level reflect annotation noise or borderline examples rather than systematic disagreement between the GPT evaluator and human judgments.

\begin{figure}[h!]
    \centering
    \begin{subfigure}[h!]{0.9\linewidth}
        \centering
        \includegraphics[width=\linewidth]{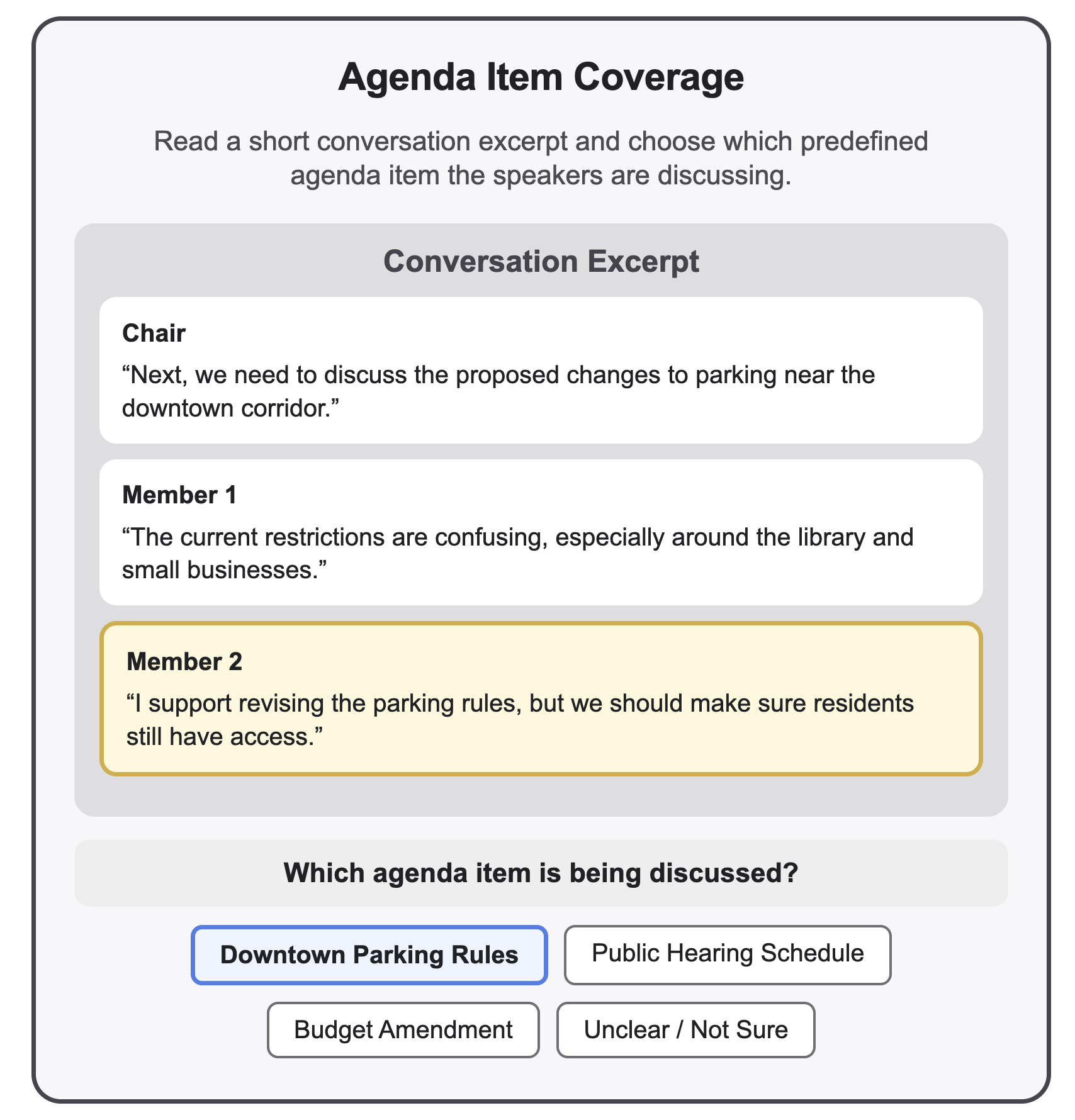}
        \caption{Task interface.}
        \label{fig:turk_coverage_ui}
    \end{subfigure}

    \vspace{0.25em}

    \begin{subfigure}[h!]{0.9\linewidth}
        \centering
        \includegraphics[width=\linewidth]{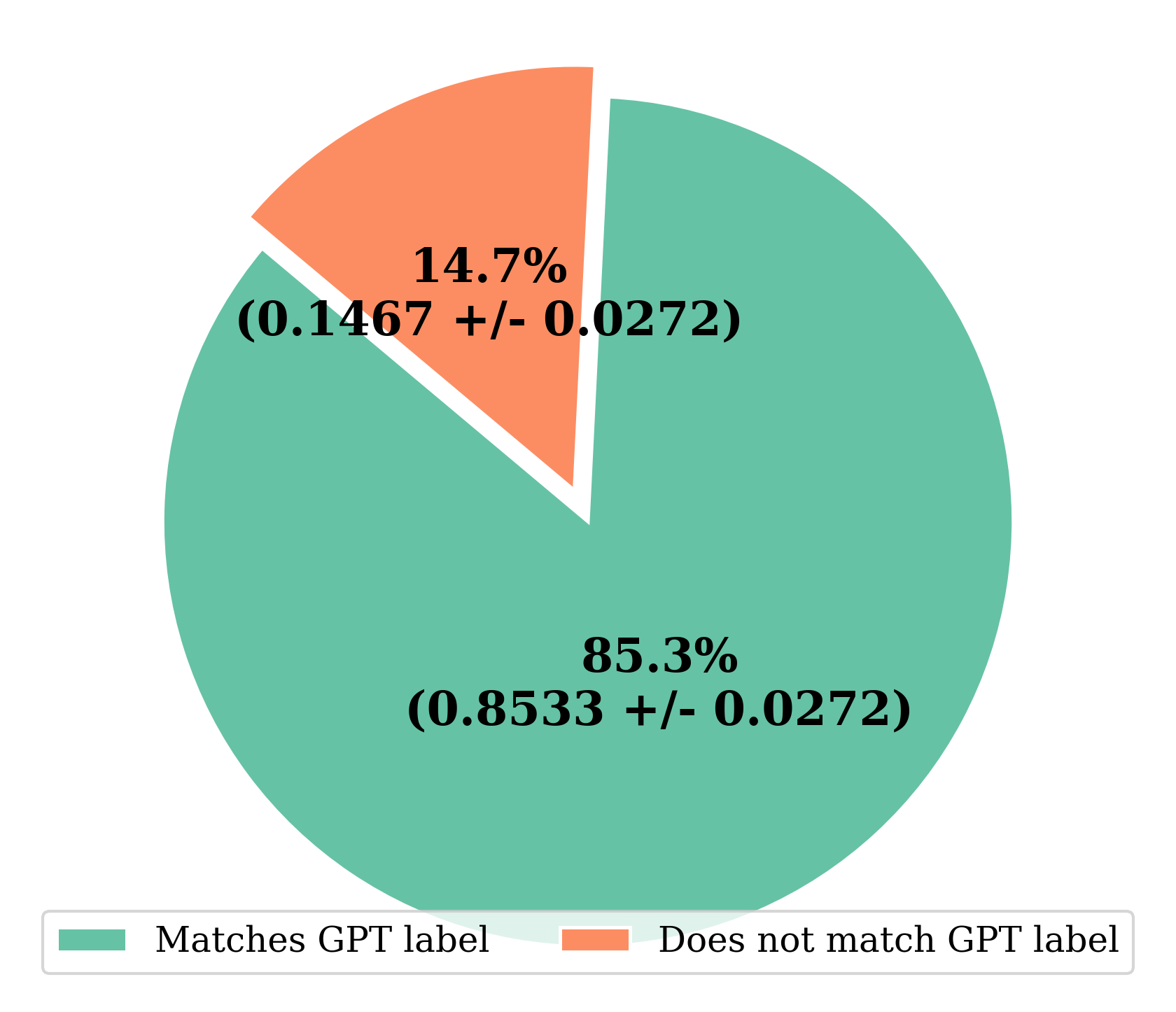}
        \caption{Worker-level match rate.}
        \label{fig:turk_coverage_wl}
    \end{subfigure}

    \vspace{0.25em}

    \begin{subfigure}[h!]{0.9\linewidth}
        \centering
        \includegraphics[width=\linewidth]{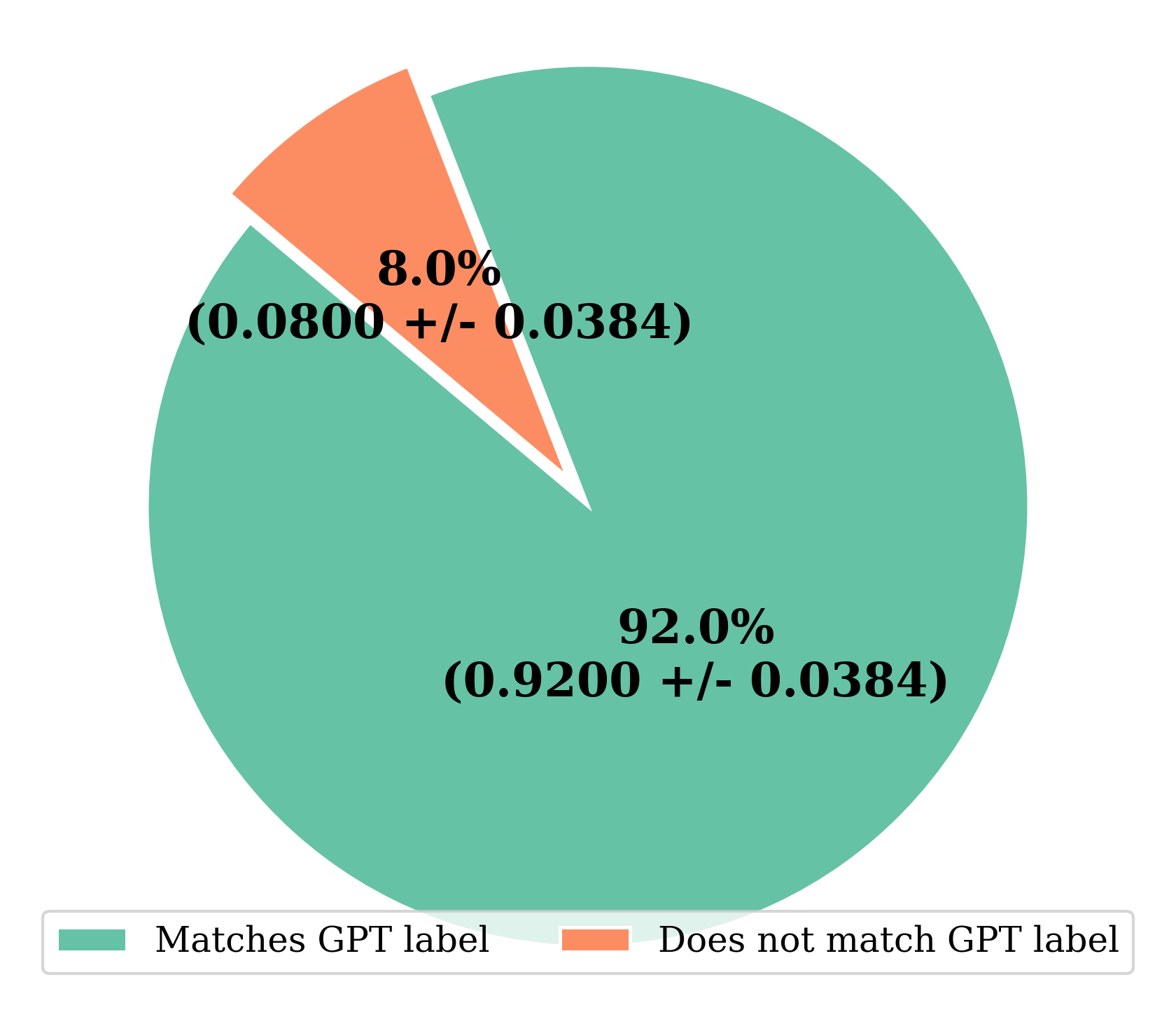}
        \caption{Majority-vote match rate.}
        \label{fig:turk_coverage_mv}
    \end{subfigure}

    \caption{
    MTurk validation for agenda item coverage. Workers are shown a short conversation excerpt and asked to select the predefined agenda item being discussed.
    }
    \label{fig:turk_coverage}
\end{figure}

\FloatBarrier
\subsection{Vote Attempt}
\label{app:vote_attempt}

Vote attempt rate measures whether a simulation attempts a vote or decision when the agenda item calls for one. This is a institutional fidelity metric: a meeting simulation should not only discuss the correct topics, but should also reproduce the decision-making steps that occur in the original institutional process.

For the human validation task, workers are shown a short conversation block and asked whether a vote occurs anywhere in the block. The answer choices are \textit{Yes}, \textit{No}, and \textit{Unclear / not sure}. A ``Yes'' label indicates that the excerpt contains a clear voting action, such as calling for a vote, recording votes, or asking members to indicate approval. A ``No'' label indicates that the block contains discussion but no vote attempt. The ``Unclear / not sure'' option is used when the language is too ambiguous to determine whether a vote occurred.

As shown in \autoref{fig:turk_vote_attempt}, vote attempt detection shows the strongest alignment between GPT-based and human judgments. The GPT metric matches individual worker labels with a match rate of $0.893 \pm 0.023$. Under majority-vote aggregation, the match rate is $0.900 \pm 0.042$. This high agreement indicates that the GPT-based vote attempt metric reliably captures the procedural signal that human annotators identify when determining whether a meeting excerpt contains a voting action.

\begin{figure}[h!]
    \centering
    \begin{subfigure}[h!]{0.85\linewidth}
        \centering
        \includegraphics[width=\linewidth]{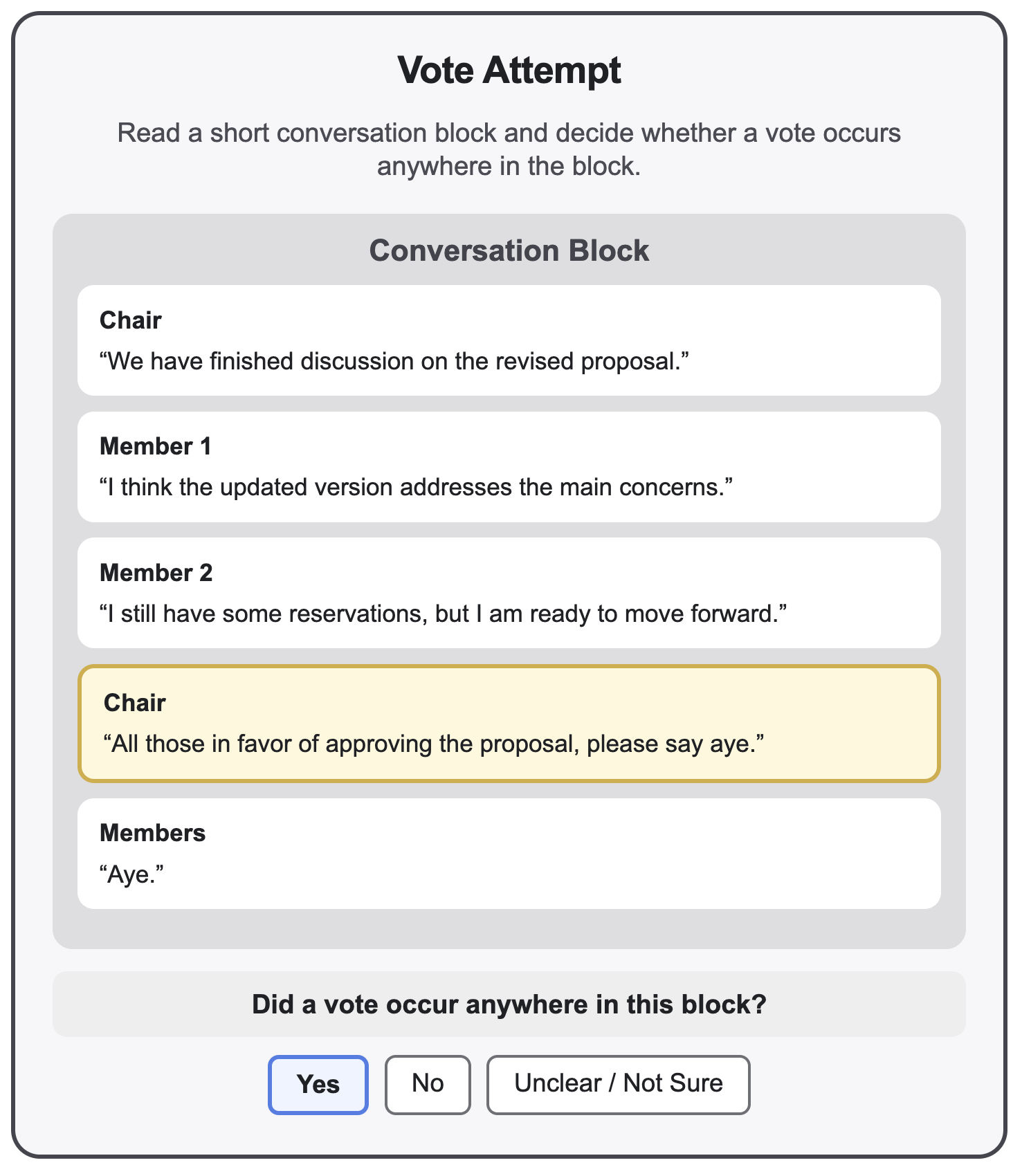}
        \caption{Task interface.}
        \label{fig:turk_vote_attempt_ui}
    \end{subfigure}

    \vspace{0.25em}

    \begin{subfigure}[h!]{0.85\linewidth}
        \centering
        \includegraphics[width=\linewidth]{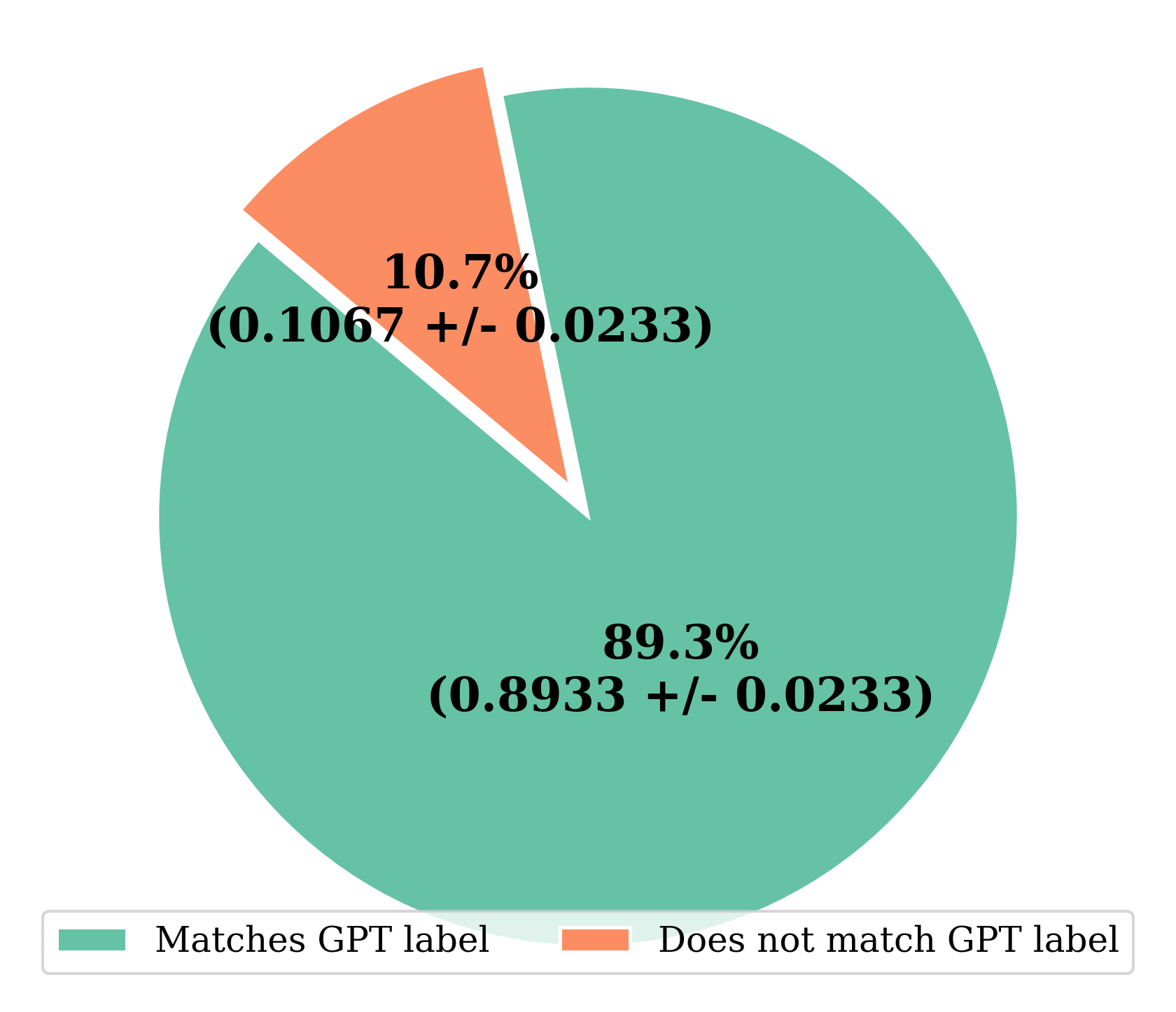}
        \caption{Worker-level match rate.}
        \label{fig:turk_vote_attempt_wl}
    \end{subfigure}

    \vspace{0.25em}

    \begin{subfigure}[h!]{0.85\linewidth}
        \centering
        \includegraphics[width=\linewidth]{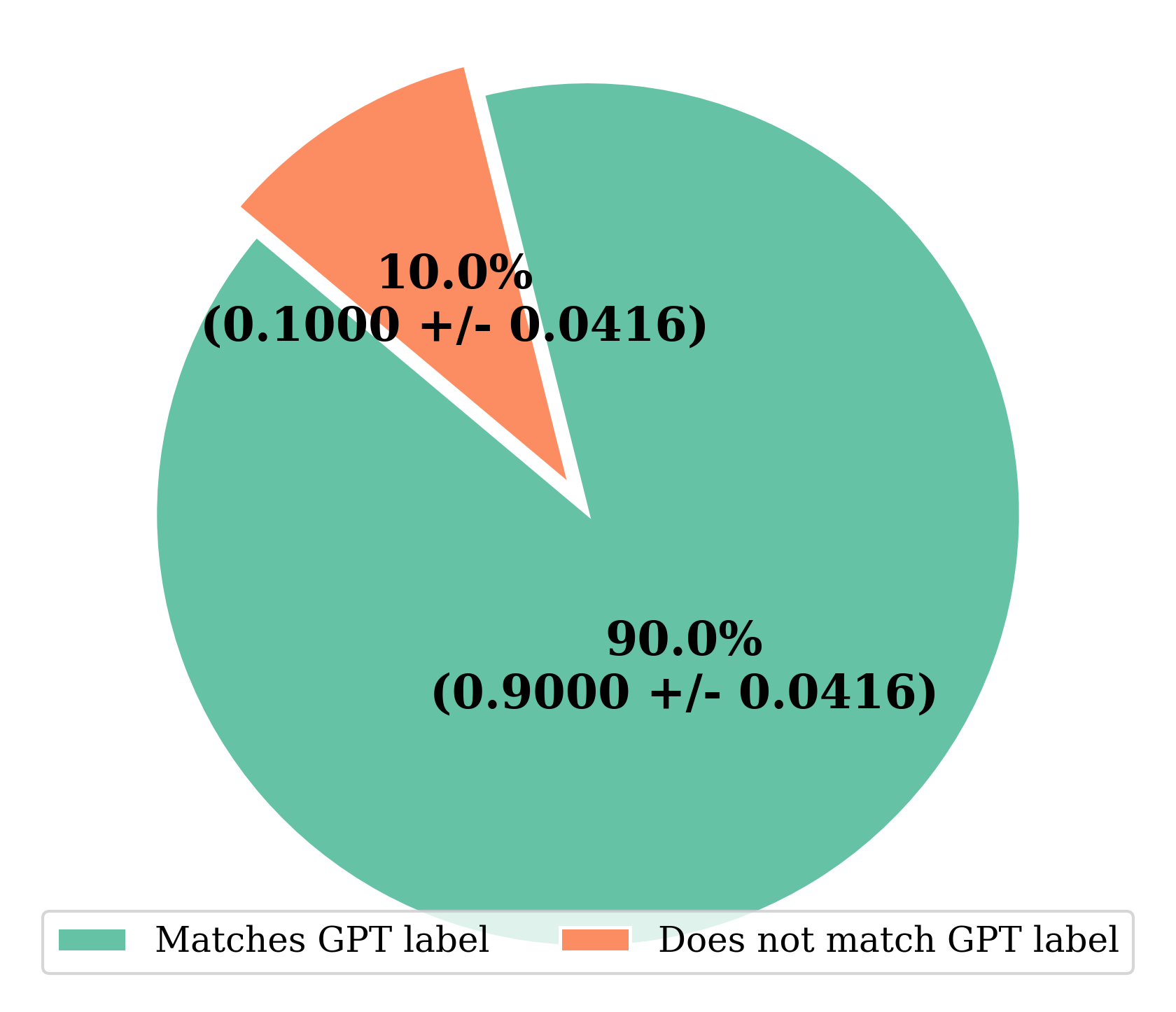}
        \caption{Majority-vote match rate.}
        \label{fig:turk_vote_attempt_mv}
    \end{subfigure}

    \caption{
    MTurk validation for vote attempt detection. Workers are shown a short conversation block and asked whether a vote occurs anywhere in the excerpt.
    }
    \label{fig:turk_vote_attempt}
\end{figure}

\FloatBarrier

\section{Additional Evaluation Results}

\subsection{All-Speaker Persona Fidelity}
\label{app:all_speaker}
In the main text, we report SAA on the subset of speakers whose speaker-attribution classifiers achieve held-out F1 above $50\%$. This filtering is necessary because SAA is only interpretable when the underlying classifier can reliably distinguish the target speaker on real held-out utterances. Speakers with low classifier F1 generally have lower support in the training set, making it difficult for the classifier to learn a stable attribution signal. Including these speakers would therefore conflate failures of the generative model with failures of the evaluator.

For completeness, \autoref{app:simulation_saa_all_speakers} reports the same SAA metric with all speakers retained. The results are materially similar to the main-text results: fine-tuned models continue to outperform prompt-only baselines, and fine-tuning with action tags achieves the highest SAA for each model family. This robustness is expected because the reliability filter excludes only four speakers---Ellen Osborne, Judge Glickman, Andrew Brown, and Roger Gordon.

\begin{table}[ht]
\centering
\scriptsize
\setlength{\tabcolsep}{4pt}
\begin{tabular}{lc}
\toprule
Model / Setting & SAA  $\uparrow$ \\
\midrule
\multicolumn{2}{l}{\textbf{GPT}}\\
\hspace{1em}Fine-Tuned w/ tags & \textbf{0.26 ± 0.04} \\
\hspace{1em}Fine-Tuned w/o tags & 0.26 ± 0.06 \\
\hspace{1em}Baseline w/ tags & 0.16 ± 0.04 \\
\hspace{1em}Baseline w/o tags & 0.19 ± 0.03 \\
\midrule
\multicolumn{2}{l}{\textbf{LLaMA}}\\
\hspace{1em}Fine-Tuned w/ tags & \textbf{0.30 ± 0.09} \\
\hspace{1em}Fine-Tuned w/o tags & 0.27 ± 0.11 \\
\hspace{1em}Baseline w/ tags & 0.18 ± 0.06 \\
\hspace{1em}Baseline w/o tags & 0.19 ± 0.06 \\
\midrule
\multicolumn{2}{l}{\textbf{Qwen}}\\
\hspace{1em}Fine-Tuned w/ tags & \textbf{0.33 ± 0.12} \\
\hspace{1em}Fine-Tuned w/o tags & 0.30 ± 0.11 \\
\hspace{1em}Baseline w/ tags & 0.18 ± 0.05 \\
\hspace{1em}Baseline w/o tags & 0.17 ± 0.04 \\
\bottomrule
\end{tabular}
\caption{
Dataset-averaged SAA with all speakers retained, reported as mean $\pm$ SE across datasets.
}
\label{app:simulation_saa_all_speakers}
\end{table}

\subsection{Per-Dataset Persona Fidelity}
\label{app:model_eval}

\autoref{fig:main_comparison} provides a visual comparison of persona fidelity across the three evaluation metrics discussed in \autoref{sec:results}. Each row corresponds to one metric---perplexity (PPL), classifier fool rate (CFR), and speaker attribution accuracy (SAA)---while columns represent the three datasets used in our study. For each model, we report results under four configurations: baseline prompting without action tags, baseline prompting with action tags, fine-tuning without action tags, and fine-tuning with action tags. Error bars denote standard errors computed across speakers within each dataset.

Several consistent patterns emerge. First, fine-tuning substantially improves persona fidelity across models and datasets. Averaged over all model--dataset pairs, fine-tuning with action tags reduces PPL from $14.37$ to $6.64$ relative to the tagged baseline, while fine-tuning without action tags reduces PPL from $20.37$ to $10.31$ relative to the untagged baseline. This reduction appears for every model and dataset, indicating that fine-tuning helps models better match the local lexical and conversational structure of the target meetings.

Second, action tags provide an additional and highly consistent improvement in PPL. In every model--dataset pair, adding action tags lowers PPL for both baseline and fine-tuned models. For example, in Waipa, \texttt{GPT} PPL decreases from $30.445 \pm 0.751$ to $19.641 \pm 0.709$ under baseline prompting, and from $15.419 \pm 0.435$ to $9.109 \pm 0.270$ after fine-tuning. Similar reductions appear for \texttt{LLaMA} and \texttt{Qwen}, suggesting that action tags help constrain generation toward the intended conversational act and reduce uncertainty over the next utterance.

Third, fine-tuning sharply improves classifier fool rate. Averaged across model--dataset pairs, CFR increases from $0.325$ to $0.639$ when comparing tagged baseline prompting to tagged fine-tuning, and from $0.289$ to $0.572$ when comparing untagged baseline prompting to untagged fine-tuning. These gains indicate that fine-tuned models produce utterances that better match the stylistic distribution of the target speakers. Action tags also tend to improve CFR, particularly in the fine-tuned setting, though the effect is not uniform across all datasets and models.

Finally, SAA also improves with fine-tuning, but remains more difficult than CFR. With action tags, average SAA increases from $0.212$ under baseline prompting to $0.446$ after fine-tuning. Without action tags, average SAA increases from $0.232$ to $0.385$. However, SAA gains are less uniform than PPL or CFR: in a few untagged settings, fine-tuning does not improve speaker attribution, and action tags do not always increase SAA for baseline models. This gap reflects the distinction between plausibility and distinctiveness. Fine-tuned models often generate utterances that appear broadly realistic for the meeting context, but reproducing the subtle speaker-specific cues needed for unique attribution remains a harder form of persona fidelity.

\begin{figure*}[h!]
\centering
\includegraphics[width=\textwidth]{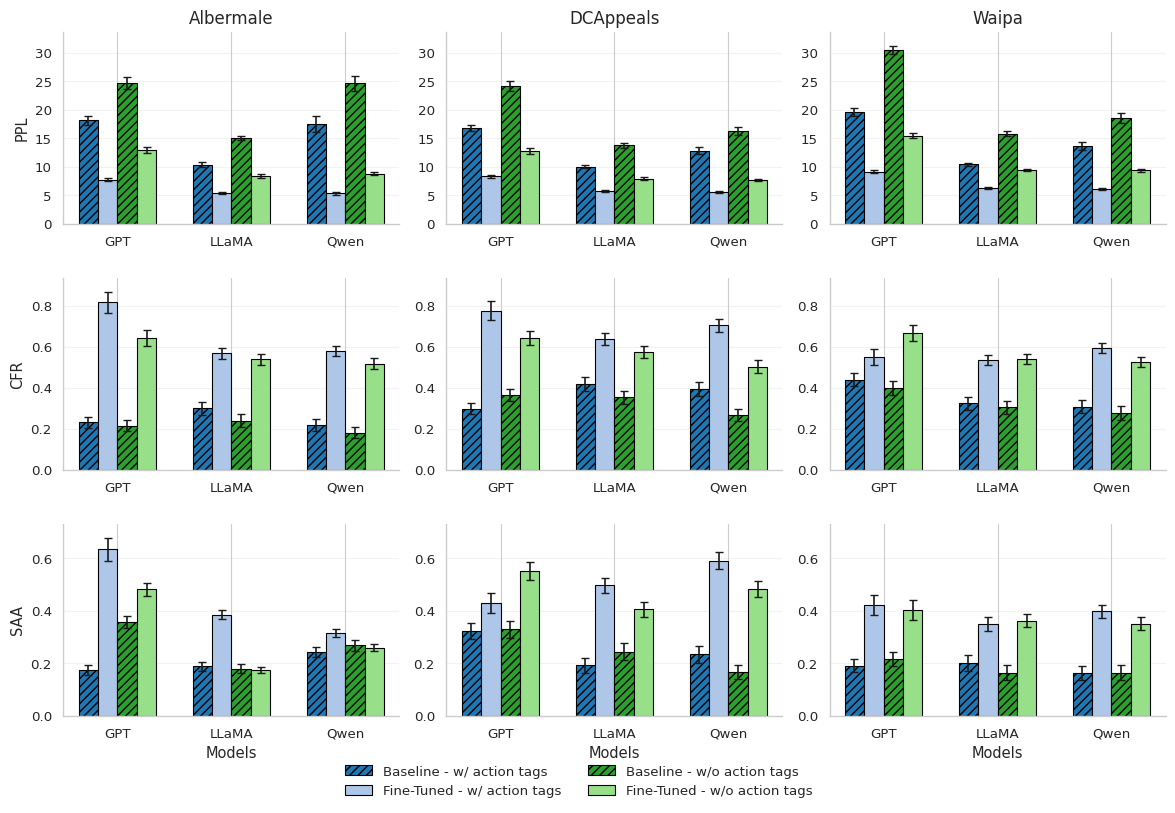}\par\vspace{6pt}
\caption{Per-model comparison across datasets.
\textbf{Perplexity} (top row): Fine-tuning substantially lowers PPL, and adding action tags further reduces surprisal for fine-tuned models.
\textbf{Fool rate} (middle row): classifier-based fooling rises sharply after fine-tuning, with the fine-tuned + tags condition generally highest.
\textbf{Speaker-attribution accuracy} (bottom row): improves with fine-tuning but often remains lower than fool rate, indicating that outputs can appear plausible for a target speaker without replicating the subtle stylistic fingerprints required for unique attribution.}
\label{fig:main_comparison}
\end{figure*}
\FloatBarrier
\subsection{Per-Dataset Persona Consistency}

\begin{figure*}[h!]
\centering
\includegraphics[width=\textwidth]{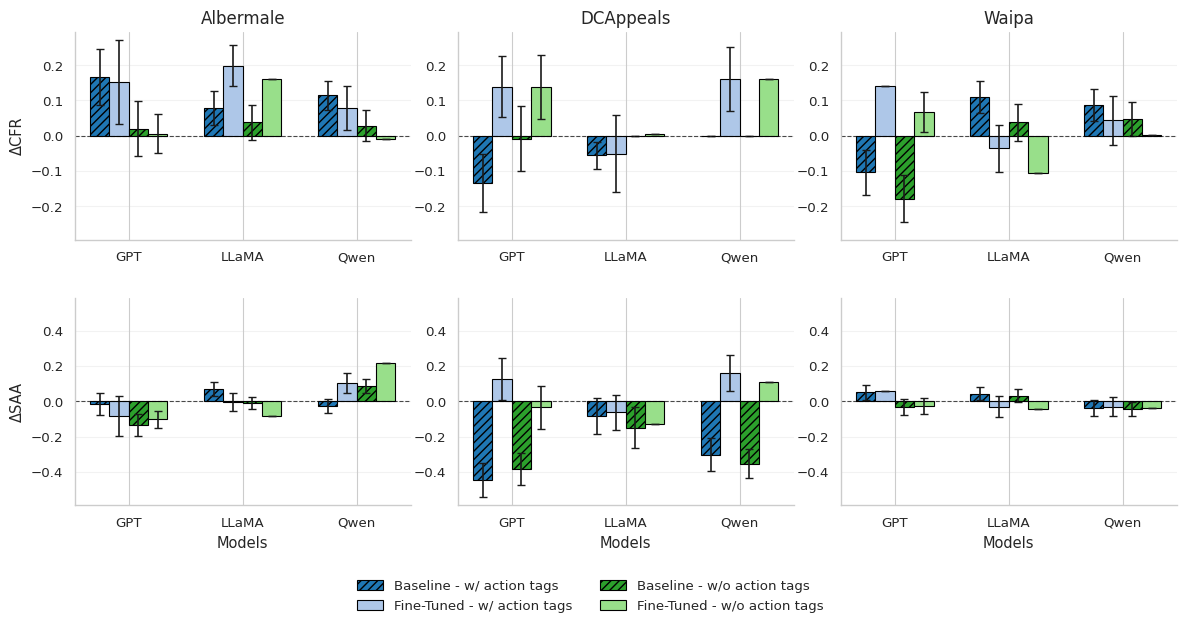}\par\vspace{6pt}
\caption{Persona consistency across long simulations. We report early-minus-late drift in CFR (top row) and SAA (bottom row); values closer to zero indicate less change over the course of a simulation.}
\label{fig:persona_consistency}
\end{figure*}

We next evaluate whether personas remain stable over the course of long simulations. For each run, we compare model behavior in early versus late turns and report signed drift in CFR and SAA, where values closer to zero indicate greater consistency. Overall, fine-tuning produces personas that are less likely to deteriorate over time, while prompt-only baselines show larger and less predictable drift.

This pattern is clearest for speaker attribution accuracy. Across model--dataset-condition comparisons, fine-tuned models reduce the average absolute SAA drift from $0.128$ to $0.080$, and improve the absolute SAA drift in 12 of 18 matched comparisons. The largest differences appear in \texttt{DCAppeals}: GPT prompt-only baselines show substantial late-simulation degradation ($\Delta$SAA $=-0.448$ with tags and $-0.383$ without tags), while the corresponding fine-tuned models are much closer to stable ($0.127$ and $-0.034$). Qwen shows a similar pattern on \texttt{DCAppeals}, with prompt-only baselines drifting by $-0.302$ and $-0.353$, compared with $0.160$ and $0.109$ after fine-tuning.

Notably, drift for fine-tuned models is often positive, meaning that later simulated turns are classified as more speaker-attributable than earlier turns. This suggests that fine-tuned agents may become more recognizable as the simulation accumulates context. At the beginning of a simulation, agents have little prior dialogue to condition on, which differs from the setup they were fine-tuned on (\autoref{fig:dataset_format}); training examples were constructed from prior dialogue and truncated to 1{,}024 tokens. As the simulation progresses, the generated context increasingly resembles this training format, giving the model more conversational history, agenda state, and speaker-interaction structure to use when producing later turns  Thus, fine-tuning helps move long-form persona modeling beyond short-horizon mimicry toward more stable behavior over extended simulations.

\subsection{Human Evaluation}
\label{app:simulation_result}

We assess the realism of simulated meetings using a large-scale MTurk study (\autoref{fig:mturk_simulation_main}). Workers were shown short excerpts consisting of \emph{four consecutive utterances} and asked to classify each excerpt as human or simulated. We chose this length to provide enough local context---including speaker transitions, topical flow, and stylistic cues---while keeping the task manageable for crowd workers.

Overall, simulated conversations were difficult for workers to distinguish from real meetings. As shown in \autoref{fig:mturk_simulation_result}, simulated excerpts were correctly identified only 44.5\% of the time, below chance, indicating that the simulations are locally realistic enough to confuse human annotators. Detection accuracy also varies by model (\autoref{fig:mturk_simulation_by_model}). \texttt{GPT} simulations were slightly easier to identify, possibly because the chain-of-thought model produces more structured dialogue.

\begin{figure}[h!]
    \centering

    \begin{subfigure}{0.95\columnwidth}
        \centering
        \includegraphics[width=\textwidth]{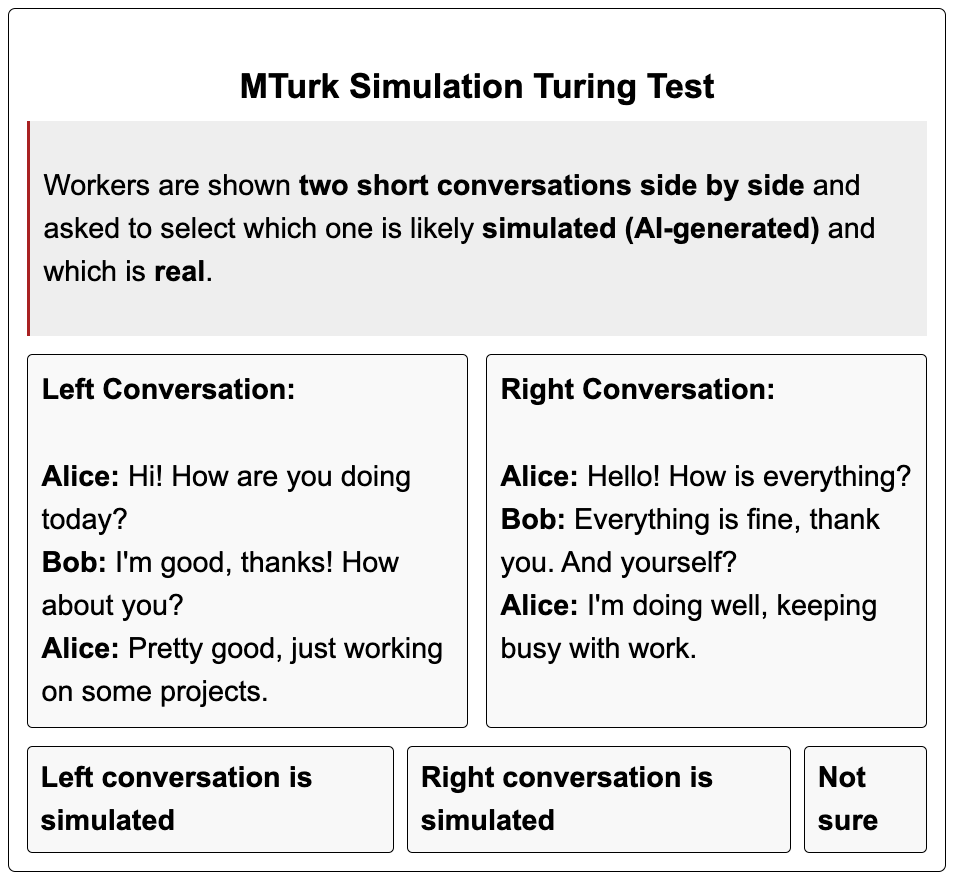}
        \caption{MTurk task interface for classifying excerpts as real or simulated.}
        \label{fig:mturk_simulation_task}
    \end{subfigure}

    \vspace{0.1em}

    \begin{subfigure}{0.95\columnwidth}
        \centering
        \includegraphics[width=\textwidth]{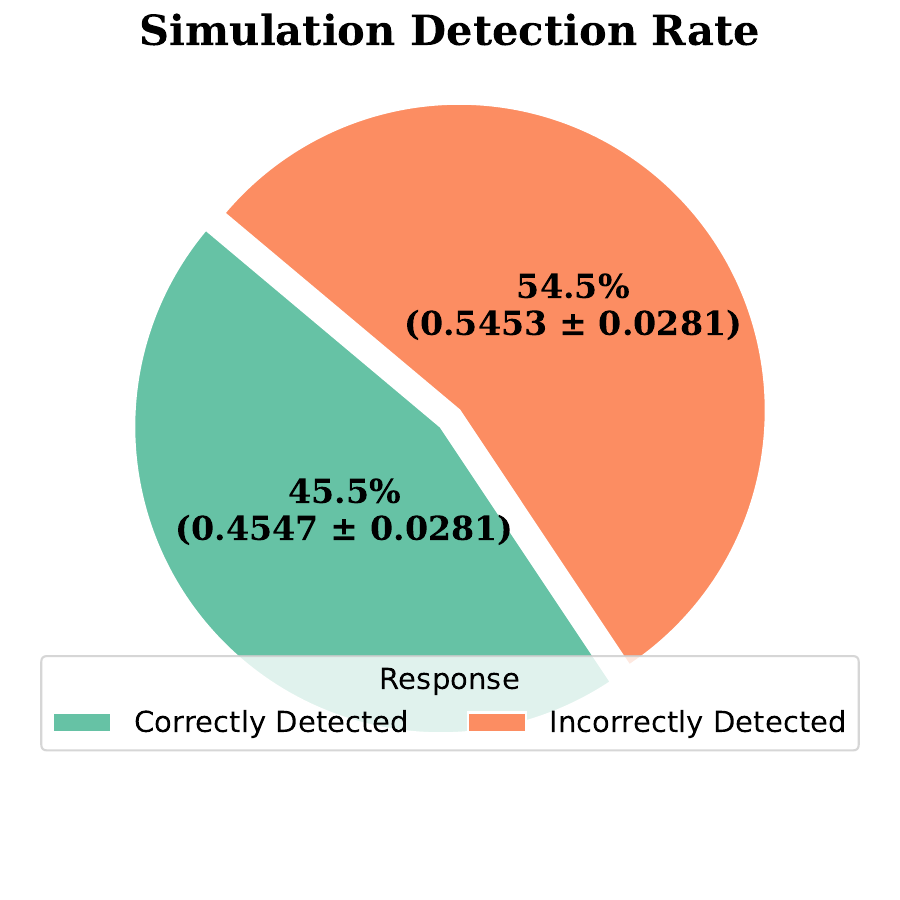}
        \caption{Aggregate human detection accuracy for simulated conversations.}
        \label{fig:mturk_simulation_result}
    \end{subfigure}

    \vspace{0.1em}

    \begin{subfigure}{0.95\columnwidth}
        \centering
        \includegraphics[width=\textwidth]{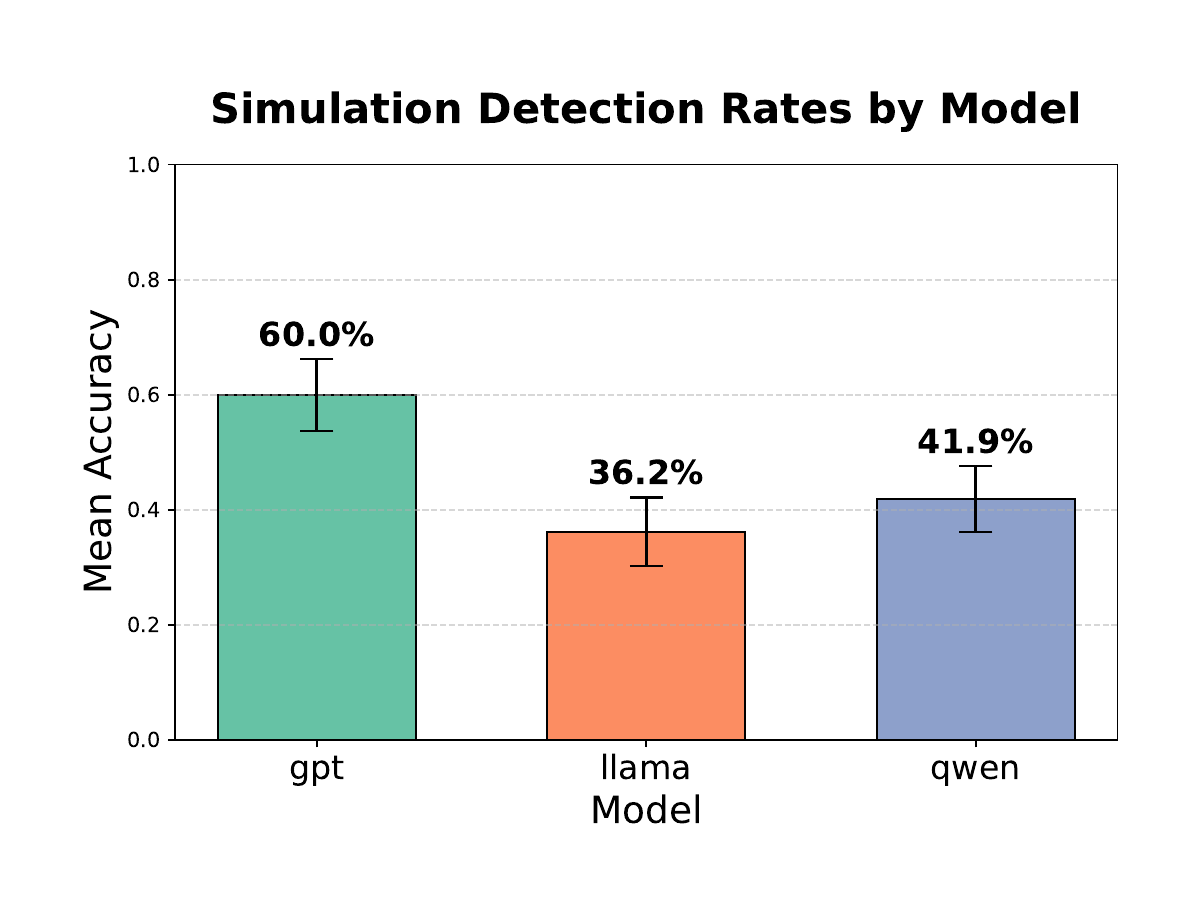}
        \caption{Human detection accuracy broken down by simulation model.}
        \label{fig:mturk_simulation_by_model}
    \end{subfigure}

    \caption{Human evaluation of simulated conversations. Panels show the task interface, aggregate detection accuracy, and detection accuracy by model.}
    \label{fig:mturk_simulation_main}
\end{figure}

\section{System Message Ablation}
\label{app:system_message_ab}

We perform ablations to evaluate how different system prompt components affect \emph{persona fidelity}. Six prompt variations were tested:

\begin{itemize}
\item \textbf{Full System Prompt (\texttt{full}):} Includes persona description, in-context examples, micro-profile, and discussion topic summaries (see \autoref{fig:persona-prompt}).
\item \textbf{No examples (\texttt{no\_examples}):} Full prompt without in-context examples.
\item \textbf{No micro-profile (\texttt{no\_micro}):} Full prompt without the micro-profile.
\item \textbf{No profile (\texttt{no\_profile}):} Prompt without persona description, in-context examples, or micro-profile.
\item \textbf{No context (\texttt{no\_context}):} Full prompt without discussion topic summaries.
\item \textbf{No System Prompt (\texttt{none}):} No system prompt provided.
\end{itemize}

\autoref{tab:system_message_summary} and \autoref{fig:system_message} report PPL, CFR, and SAA across all datasets and models for each system message configuration.

Two clear patterns emerge. For \textbf{non-fine-tuned models}, the system prompt is critical for persona fidelity. Without fine-tuning, models rely heavily on prompt conditioning to identify the target speaker, adopt the appropriate conversational role, and remain grounded in the meeting context. Omitting persona information or removing the system prompt sharply reduces CFR and SAA, as these models often default to generic responses or summaries rather than speaking in character. Quantitatively, \autoref{tab:system_message_summary} shows that removing the system prompt entirely (\texttt{none}) reduces CFR from $0.29 \pm 0.02$ under the full prompt to $0.08 \pm 0.01$, and reduces SAA from $0.23 \pm 0.02$ to $0.09 \pm 0.01$. This confirms that prompt-provided persona information is essential for non-fine-tuned models.

For \textbf{fine-tuned models}, the role of the system prompt shifts. Fine-tuning already encodes much of the task structure and speaker-specific style, so additional conditioning through micro-profiles or in-context examples can become redundant or even mildly harmful. In particular, \autoref{tab:system_message_summary} shows that omitting micro-profiles (\texttt{no\_micro}) or in-context examples (\texttt{no\_examples}) improves persona-fidelity metrics relative to the full prompt: CFR increases from $0.64 \pm 0.03$ to $0.89 \pm 0.03$, and SAA increases from $0.45 \pm 0.04$ to $0.61 \pm 0.04$. This suggests that overly detailed prompts may overcondition fine-tuned models or introduce cues that conflict with the style learned during training. Nevertheless, performance with the full system prompt remains competitive, indicating that fine-tuned models are substantially more robust to prompt variation than non-fine-tuned models.

Overall, the system message ablation shows that prompt conditioning is important for persona fidelity, but its role depends on whether the model has been fine-tuned. Non-fine-tuned models require detailed prompts to maintain speaker identity and context, whereas fine-tuned models can preserve persona fidelity even with simplified prompts.

\begin{table}[h!]
\centering
\scriptsize
\setlength{\tabcolsep}{4pt}
\begin{tabular}{lccc}
\toprule
System Message & PPL $\downarrow$ & Fool Rate $\uparrow$ & SAA $\uparrow$ \\
\midrule
\multicolumn{4}{l}{\textbf{full}}\\
\hspace{1em}Baseline w/o tags & 20.37 ± 1.92 & 0.29 ± 0.02 & 0.23 ± 0.02 \\
\hspace{1em}Baseline w/ tags & 14.37 ± 1.24 & 0.32 ± 0.03 & 0.21 ± 0.02 \\
\hspace{1em}Fine-Tuned w/o tags & 10.31 ± 0.90 & 0.57 ± 0.02 & 0.39 ± 0.04 \\
\hspace{1em}Fine-Tuned w/ tags & 6.64 ± 0.47 & 0.64 ± 0.03 & 0.45 ± 0.04 \\
\midrule
\multicolumn{4}{l}{\textbf{no\_context}}\\
\hspace{1em}Baseline w/o tags & 21.96 ± 1.92 & 0.18 ± 0.01 & 0.13 ± 0.01 \\
\hspace{1em}Baseline w/ tags & 14.95 ± 1.12 & 0.17 ± 0.02 & 0.13 ± 0.01 \\
\hspace{1em}Fine-Tuned w/o tags & 10.78 ± 0.90 & 0.44 ± 0.02 & 0.29 ± 0.03 \\
\hspace{1em}Fine-Tuned w/ tags & 6.83 ± 0.47 & 0.49 ± 0.02 & 0.32 ± 0.03 \\
\midrule
\multicolumn{4}{l}{\textbf{no\_examples}}\\
\hspace{1em}Baseline w/o tags & 20.90 ± 1.80 & 0.32 ± 0.03 & 0.24 ± 0.02 \\
\hspace{1em}Baseline w/ tags & 14.18 ± 1.01 & 0.30 ± 0.04 & 0.21 ± 0.01 \\
\hspace{1em}Fine-Tuned w/o tags & 10.33 ± 0.89 & 0.73 ± 0.03 & 0.51 ± 0.04 \\
\hspace{1em}Fine-Tuned w/ tags & 6.64 ± 0.48 & 0.85 ± 0.03 & 0.55 ± 0.05 \\
\midrule
\multicolumn{4}{l}{\textbf{no\_micro}}\\
\hspace{1em}Baseline w/o tags & 18.99 ± 1.70 & 0.32 ± 0.03 & 0.26 ± 0.02 \\
\hspace{1em}Baseline w/ tags & 13.45 ± 1.11 & 0.32 ± 0.03 & 0.26 ± 0.01 \\
\hspace{1em}Fine-Tuned w/o tags & 10.20 ± 0.87 & 0.80 ± 0.02 & 0.53 ± 0.05 \\
\hspace{1em}Fine-Tuned w/ tags & 6.53 ± 0.45 & 0.89 ± 0.03 & 0.61 ± 0.04 \\
\midrule
\multicolumn{4}{l}{\textbf{no\_profile}}\\
\hspace{1em}Baseline w/o tags & 18.09 ± 1.75 & 0.31 ± 0.03 & 0.24 ± 0.03 \\
\hspace{1em}Baseline w/ tags & 12.49 ± 1.07 & 0.32 ± 0.04 & 0.24 ± 0.02 \\
\hspace{1em}Fine-Tuned w/o tags & 10.08 ± 0.84 & 0.62 ± 0.02 & 0.43 ± 0.03 \\
\hspace{1em}Fine-Tuned w/ tags & 6.58 ± 0.49 & 0.70 ± 0.03 & 0.46 ± 0.04 \\
\midrule
\multicolumn{4}{l}{\textbf{none}}\\
\hspace{1em}Baseline w/o tags & 18.22 ± 2.19 & 0.08 ± 0.01 & 0.09 ± 0.01 \\
\hspace{1em}Baseline w/ tags & 12.63 ± 1.28 & 0.09 ± 0.01 & 0.09 ± 0.01 \\
\hspace{1em}Fine-Tuned w/o tags & 10.48 ± 0.92 & 0.29 ± 0.02 & 0.20 ± 0.02 \\
\hspace{1em}Fine-Tuned w/ tags & 6.69 ± 0.50 & 0.34 ± 0.02 & 0.22 ± 0.03 \\
\midrule
\bottomrule
\end{tabular}
\caption{Persona fidelity across system message ablations. We report perplexity, classifier fool rate, and speaker attribution accuracy for each dataset, model, and system prompt configuration.}
\label{tab:system_message_summary}
\end{table}

\FloatBarrier

\section{Temporal Grounding Ablation}
Across models, temporal grounding substantially improves institutional fidelity (\autoref{fig:simulation_comparison}). The clearest gains appear in agenda coverage (\autoref{fig:agenda_coverage}). Time-aware simulations improve coverage from $0.626 \pm 0.113$ to $0.830 \pm 0.109$ for \texttt{GPT}, from $0.467 \pm 0.209$ to $0.949 \pm 0.051$ for \texttt{LLaMA}, and from $0.815 \pm 0.158$ to $0.907 \pm 0.067$ for \texttt{Qwen}. These results suggest that temporal cues help simulations continue progressing through the full agenda, especially for models that otherwise stall or over-focus on early agenda items.

Temporal grounding also generally improves agenda ordering (\autoref{fig:agenda_order}). \texttt{GPT} improves from $0.639 \pm 0.028$ to $0.806 \pm 0.121$, and \texttt{LLaMA} improves from $0.878 \pm 0.062$ to $0.944 \pm 0.056$. \texttt{Qwen} remains high in both settings, with a slight decrease from $0.917 \pm 0.083$ to $0.874 \pm 0.021$. Thus, temporal cues primarily help models maintain agenda progression, while ordering is already relatively strong for some models even without explicit timing information.

The largest relative improvements occur for vote attempts (\autoref{fig:vote_attempt}). Without temporal grounding, models often discuss agenda items without reliably reaching the procedural endpoint of a vote or decision. Adding temporal cues increases vote attempt rates from $0.169 \pm 0.058$ to $0.330 \pm 0.035$ for \texttt{GPT}, from $0.093 \pm 0.093$ to $0.591 \pm 0.063$ for \texttt{LLaMA}, and from $0.156 \pm 0.034$ to $0.500 \pm 0.000$ for \texttt{Qwen}. This indicates that temporal grounding not only broadens discussion coverage, but also helps simulations complete the procedural actions expected in real meetings.  Overall, time-aware prompting improves institutional fidelity across coverage, ordering, and vote attempts, suggesting that simple temporal cues help keep agents on track in extended simulations.

\begin{figure}[h!]
    \centering

    \begin{subfigure}{0.32\textwidth}
        \centering
        \includegraphics[width=\textwidth]{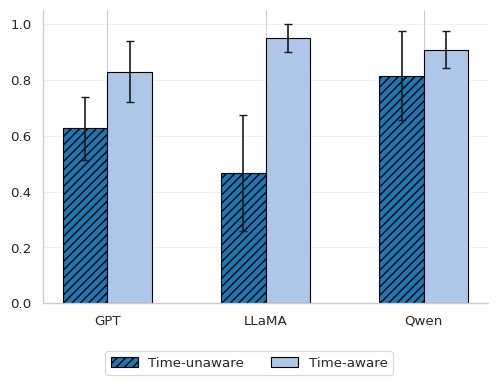}
        \caption{Agenda coverage}
        \label{fig:agenda_coverage}
    \end{subfigure}
    \hfill
    \begin{subfigure}{0.32\textwidth}
        \centering
        \includegraphics[width=\textwidth]{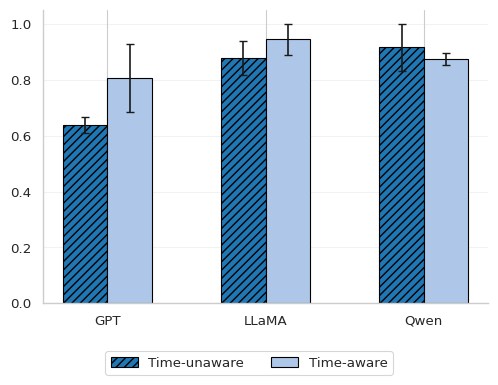}
        \caption{Agenda ordering}
        \label{fig:agenda_order}
    \end{subfigure}
    \hfill
    \begin{subfigure}{0.32\textwidth}
        \centering
        \includegraphics[width=\textwidth]{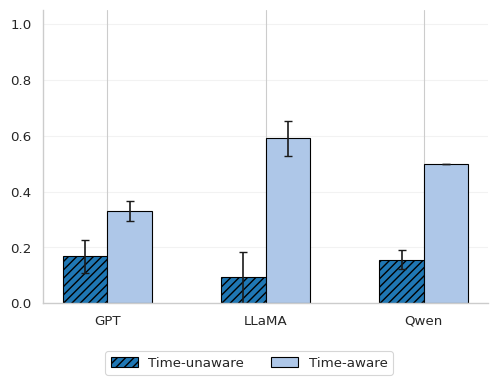}
        \caption{Vote attempt rate}
        \label{fig:vote_attempt}
    \end{subfigure}

    \caption{Temporal grounding improves institutional fidelity in simulated meetings. Time-aware simulations more reliably cover agenda items, preserve agenda order, and attempt votes or decisions when required.}
    \label{fig:simulation_comparison}
\end{figure}

\FloatBarrier
\section{Fine-tuning Method Ablation}
\label{app:fine-tune-ab}

We compared LoRA and QLoRA using the speaker with the most training data per dataset, evaluating PPL, CFR, and SAA. LoRA achieves slightly lower PPL, but differences in CFR and SAA are inconsistent, indicating comparable downstream quality. Overall, QLoRA offers substantial memory savings with minimal impact on generation performance, making it a practical default for fine-tuning.

\begin{figure}[h!]
    \includegraphics[width=1.0\columnwidth]{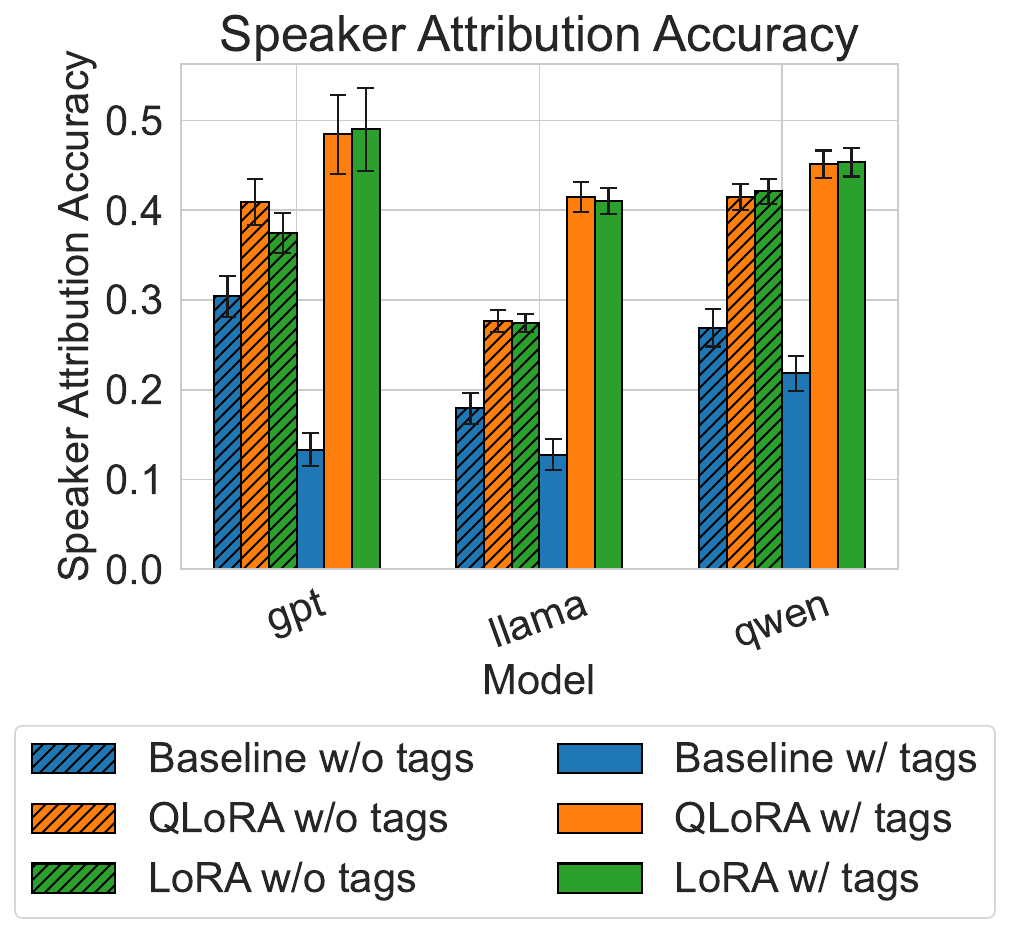}
    \caption{Performance comparison between LoRA and QLoRA fine-tuning. Average Speaker Attribution Accuracy is reported for the three speakers with the largest amount of training data in each dataset.}
        \label{fig:lora_qlora}
\end{figure}

\FloatBarrier

\section{Simulations}
We provide representative simulation excerpts in \autoref{fig:sim_albemarle}, \autoref{fig:sim_dcappeals}, and \autoref{fig:sim_waipa}. Each figure shows the first 10 minutes of a generated meeting, illustrating how simulated participants follow the agenda, adopt speaker-specific styles, and progress through institutional procedures.

\begin{figure}[h!]
\footnotesize
\centering

\begin{tcolorbox}[colback=gray!3,colframe=black!15,
title=\textbf{Albemarle Dataset},
width=\linewidth]

\textbf{Graham Paige}: Good morning, everybody. I am calling this special board meeting for the Albemarle County School Board to order.

\medskip
\textbf{David Oberg}: So, uh, what do you think our main focus needs to be?

\medskip
\textbf{Ellen Osborne}: Well, I mean, shouldn't we assess how well our current safety protocols have worked so far?

\medskip
\textbf{Graham Paige}: And also look at the percentage of kids who've opted out because they're not coming back yet.

\medskip
\textbf{Jonno Alcaro}: What is the latest data showing regarding infections among teachers?

\medskip
\textbf{Judy Le}: Um, yeah, I guess my question would be similar along those lines. Are people getting sick outside of school?

\medskip
\textbf{Kate Acuff}: We’ve had several cases where families were exposed over holidays but didn’t get infected until after they returned home from vacation.

\medskip
\textbf{Katrina Callsen}: Kate, did your statement answer Judy's? Because I missed her exact wording. Did she say something specific about teachers versus students? Or maybe I'm misremembering.

\medskip
\textbf{David Oberg}: It seems to me that the issue here isn't really whether or not the infection occurred within the building walls. The fact remains that if someone gets an active case, then they’re going to miss two weeks of work.

\medskip
\textbf{Ellen Osborne}: That's true - so aren't we essentially trying to balance between keeping everyone safe and figuring out how many absences we could handle without totally disrupting everything?

\medskip
\textbf{Graham Paige}: OK, thank you all very much. Our next section will begin now.

\medskip
\textbf{Jonno Alcaro}: How does the number of substitute teachers available compare to previous years?

\end{tcolorbox}

\caption{Transcript excerpt from Albemarle simulation.}
\label{fig:sim_albemarle}

\end{figure}

\begin{figure}[h!]
\footnotesize
\centering

\begin{tcolorbox}[colback=gray!3,colframe=black!15,
title=\textbf{DCAppeals Dataset},
width=\linewidth]

\textbf{Courtroom Clerk}: All persons having business for the Honorable Chief Judge and Associate Judges are residing with the District of Columbia Court of Appeals. Draw near and give your attention. God save the United States and this Honorable Court. This Honorable Court is now in session. Please come forward.

\medskip
\textbf{Judge Mcleese}: So uh, welcome everyone. And thank you all for joining us today. We're excited to be here and um, look forward to hearing from both sides. The next step will be opening statement by our appellant. That'll take place between nine AM and eight minutes past nine. Our appellant can begin his presentation whenever he'd like. Thank you very much.

\medskip
\textbf{John F. General}: Thank you, Your Honor. This case presents a question of first impression in this court about what it means to be in custody for purposes of Miranda. And the question is whether a person who is questioned in a locked room with a locked door, who is not free to leave, and who is subjected to sharp and pointed questioning about multiple allegations of wrongdoing, is in custody for purposes of Miranda. And we say that that person is in custody.

\medskip
\textbf{Judge Deahl}: Um, counsel, could you tell me where you think the line gets drawn? Is it, uh, once someone says "you know, you don't really have to answer these questions," does that change anything? Or do you think it doesn't matter what words are used?

\medskip
\textbf{John F. General}: Well, I think the line gets drawn, Your Honor, when the person is subjected to the functional equivalent of an arrest. And in this case, the functional equivalent of an arrest was when the door was locked behind me and I was not free to leave. Now, I think it's a little bit misleading to say that the line gets drawn when someone says, "you don't have to answer these questions," because the person may still be in custody even if they're not free to leave. And so, I think it's a bit of a red herring to focus on what words are used. Instead, I think we should focus on the objective circumstances of the interrogation.

\medskip
\textbf{Judge Glickman}: So, uh, yeah, but isn’t that true, though - that the officers told him, you’re not under arrest, right?

\medskip
\textbf{John F. General}: They did say that, Your Honor. But I think that's a bit of a, uh, a distinction without a difference. I mean, the fact that they said I wasn't under arrest didn't change the fact that I was locked in a room and not free to leave. And I think that's the key factor here. I think it's the objective circumstances of the interrogation that matter, not the words that are used.

\medskip
\textbf{Judge Mcleese}: So, one way to frame the issue might be, uh, assuming that the police officer had given a truthful warning saying, "Mr. General, you are currently being detained." Do you think that changes the analysis significantly?

\end{tcolorbox}

\caption{Transcript excerpt from DCAppeals simulation.}
\label{fig:sim_dcappeals}

\end{figure}

\FloatBarrier
\begin{figure}[h!]
\footnotesize
\centering

\begin{tcolorbox}[colback=gray!3,colframe=black!15,
title=\textbf{Waipā Dataset},
width=\linewidth]

\textbf{Jim Mylchreest}: So, we are now live on YouTube. So, yeah, welcome everyone to the Audit and Risk Committee meeting for today.

\medskip
\textbf{Andrew Brown}: Thanks Jimmy, great introduction, uh, thanks also to Susan for her report.

\medskip
\textbf{Ange Holt}: And thank you too, Andrew.

\medskip
\textbf{Claire St Pierre}: Oh, sorry, could I have my audio unmuted? Thank you very much. Yes, thanks Sue. It looks really good. The one thing that caught me is page four where it talks about the premium increase, which is \$400{,}000 or something over two years. That seems quite high compared to what I'm used to seeing. Is this because of the transition between brokers or is it an actual change in premiums generally across all councils at the moment?

\medskip
\textbf{Jim Mylchreest}: Louise can answer those questions better than I can. But look, obviously moving from Marsh to Aon will make a difference. There'll be changes in how they perceive our risk profile and everything else. So, but certainly not happy with increases of that size. We've got to keep pushing back. So, Louisa, did you want to comment on that?

\medskip
\textbf{Lou Brown}: Claire, through you, Madam Chair, um, so I think, firstly, around the broker change, it wasn't necessarily driven by cost savings. Um, although we do expect to see some benefits coming forward. Our existing contract had come up for renewal, and we felt that it would be beneficial to go to tender and test the market again.

\medskip
\textbf{Roger Gordon}: Can you tell us more about why the previous arrangement needed changing? Was it purely based on costs or were there other factors involved?

\medskip
\textbf{Sue Milner}: No, well, Roger, I'd say probably the main reason was that we weren't getting the service level we wanted, particularly after COVID. They didn't seem to be able to deliver anything except via email.

\medskip
\textbf{Andrew Brown}: Yeah, I agree completely. Look, I think we're going to run out of time here soon, so maybe we can discuss further offline.

\medskip
\textbf{Ange Holt}: Yep, yep, okay, cool. Well, I suppose while we're talking about the insurance policy, does someone want to put a motion together that basically supports the recommendation to adopt the risk-based insurance strategy framework and endorse Aon's placement and renewal approach for the 1 November cycle? Or do people feel comfortable enough with the information provided to support that without having to read through pages three and four in detail? Because personally, I haven't done that yet. I haven't gotten down to reading them thoroughly. So I wouldn't mind hearing comments from somebody else before we proceed.

\medskip
\textbf{Claire St Pierre}: Yeah, Ange, I thought it looked really good myself. Like, I liked the way she set it out. She covered off the different areas \textit{[truncated]}

\end{tcolorbox}

\caption{Transcript excerpt from Waipā simulation.}
\label{fig:sim_waipa}

\end{figure}

\section{Artifact Licenses and Terms of Use}
\label{app:artifact_licenses}

This work creates and uses several artifacts, including released datasets, code, public source recordings, pretrained models, software libraries, and human annotations.

\paragraph{Released artifacts.}
We release three speaker-attributed civic deliberation datasets, together with derived persona metadata and pragmatic action tags, as well as the code used to construct and evaluate them. The released datasets are distributed under \texttt{CC BY 4.0}, and the released code is distributed under the \texttt{MIT} license.

\paragraph{Source recordings.}
The datasets are derived from publicly available civic deliberation recordings, including public meetings and hearings. We do not redistribute the original raw videos. The original recordings remain governed by the terms, copyright policies, and public-record policies of their original hosts.

\paragraph{Third-party artifacts.}
Our pipeline uses third-party tools and models, including \texttt{yt-dlp}, Hugging Face libraries, open-weight language models, GPT-based APIs, DeBERTa-family classifiers, and Amazon Mechanical Turk. We cite these artifacts where used and rely on them under their respective licenses and terms of use. We do not redistribute third-party model weights or worker-identifying information.

\paragraph{User responsibility.}
Users of the released artifacts are responsible for complying with the dataset and code licenses, as well as any applicable terms governing the original source recordings and third-party tools or models.

\end{document}